\documentclass[final,3p,times,authoryear]{elsarticle}

\usepackage{amssymb}

\usepackage{url}
\usepackage{xcolor}
\usepackage{booktabs}
\usepackage{mdframed}
\usepackage{lipsum}
\usepackage{amsmath}
\usepackage{multirow}
\usepackage{rotating}
\usepackage[]{linguex}
\usepackage{CJKutf8}
\usepackage{subfig}

\makeatletter
\newcommand\blfootnote[1]{%
  \begingroup
  \renewcommand\thefootnote{}\footnote{#1}%
  \addtocounter{footnote}{-1}%
  \endgroup
}
\makeatother

\journal{Information Fusion}

\begin{document}

\begin{frontmatter}



\title{A Survey on Semantic Processing Techniques}


\author[label1]{Rui Mao\corref{equalcontribution}}
\cortext[equalcontribution]{These authors contributed equally.}
\ead{rui.mao@ntu.edu.sg}

\author[label3]{Kai He\corref{equalcontribution}}
\ead{kai_he@nus.edu.sg}

\author[label2]{Xulang Zhang\corref{equalcontribution}}
\ead{xulang001@e.ntu.edu.sg}

\author[label4,label5,label6]{Guanyi Chen\corref{equalcontribution}}
\ead{g.chen@ccnu.edu.cn}

\author[label2]{Jinjie Ni\corref{equalcontribution}}
\ead{jinjie001@e.ntu.edu.sg}

\author[label1]{Zonglin Yang\corref{equalcontribution}}
\ead{zonglin001@e.ntu.edu.sg}

\author[label1]{Erik Cambria\corref{correspondingauthor}}\cortext[correspondingauthor]{This is to indicate the corresponding author.}
\ead{cambria@ntu.edu.sg}

\affiliation[label1]{organization={Continental-NTU Corporate Lab, Nanyang Technological University, 50 Nanyang Avenue, 639798},
            country={Singapore}}

\affiliation[label2]{organization={School of Computer Science and Engineering, Nanyang Technological University, 50 Nanyang Avenue, 639798},
            country={Singapore}}
            
\affiliation[label3]{organization={Saw Swee Hock School of Public Health, National University of Singapore, 117549},
            country={Singapore}}
            
\affiliation[label4]{organization={Hubei Provincial Key Laboratory of Artificial Intelligence and Smart Learning, Central China Normal University, 382 Xiongchu Avenue, 430079, Wuhan},
            country={China}}

\affiliation[label5]{organization={National Language Resources Monitoring and Research Center for Network Media, Central China Normal University, 382 Xiongchu Avenue, 430079, Wuhan},
            country={China}}

\affiliation[label6]{organization={School of Computer Science, Central China Normal University, 382 Xiongchu Avenue, 430079, Wuhan},
            country={China}}

\begin{abstract}
Semantic processing \blfootnote{\textit{Published at Information Fusion, Volume 101, 2024, 101988, ISSN 1566-2535. The equal contribution mark is missed in the published version due to the publication policies. Please contact Prof. Erik Cambria for details.}} is a fundamental research domain in computational linguistics. In the era of powerful pre-trained language models and large language models, the advancement of research in this domain appears to be decelerating. However, the study of semantics is multi-dimensional in linguistics. The research depth and breadth of computational semantic processing can be largely improved with new technologies. In this survey, we analyzed five semantic processing tasks, e.g., word sense disambiguation, anaphora resolution, named entity recognition, concept extraction, and subjectivity detection. We study relevant theoretical research in these fields, advanced methods, and downstream applications. We connect the surveyed tasks with downstream applications because this may inspire future scholars to fuse these low-level semantic processing tasks with high-level natural language processing tasks. The review of theoretical research may also inspire new tasks and technologies in the semantic processing domain. Finally, we compare the different semantic processing techniques and summarize their technical trends, application trends, and future directions.

\end{abstract}



\begin{keyword}
Semantic Processing \sep Word Sense Disambiguation \sep Anaphora Resolution \sep Named Entity Recognition \sep Concept Extraction \sep Subjectivity Detection


\end{keyword}

\end{frontmatter}


\section{Introduction}
\label{sect:Introduction}

Semantics is a linguistic term, generally referring to the meaning of language. Unlike syntax which studies the structure of sentences~\citep{zhang2023syntactic}, the significance of semantics lies in its ability to aid our comprehension of how meaning is conveyed through words, phrases, and sentences, as well as how language is used to express various ideas, thoughts, and emotions. Language is one of the important carriers of meanings. However, the term ``meaning'' encompasses multiple aspects of language. \citet{palmer1981semantics} argued that there is a lack of consensus regarding the nature of ``meaning'', e.g., which components should be considered part of semantics, and how it should be characterized. Thus, the study of ``semantics'' is also multi-dimensional in academia.

\begin{figure}[t]
\centering
\includegraphics[width=12cm, scale=0.8]{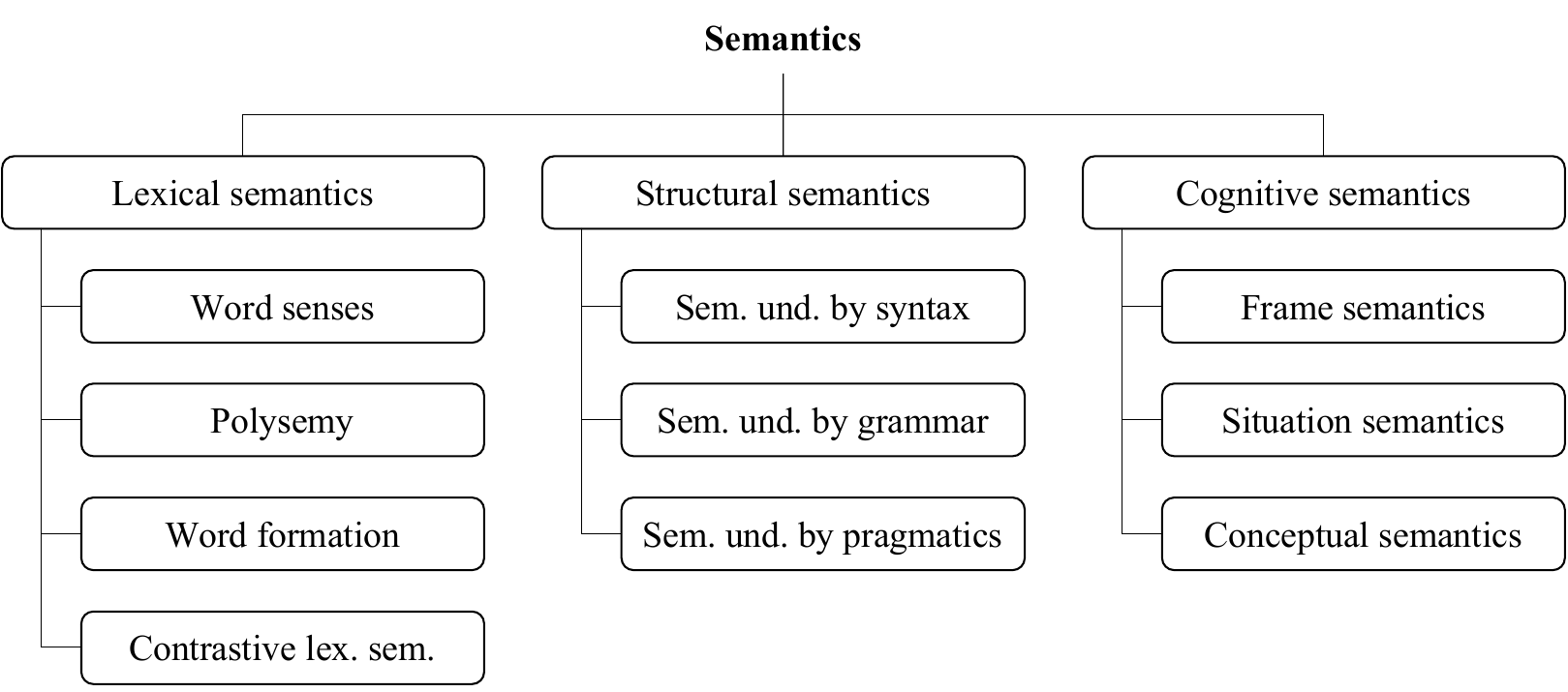}
\caption{Semantic research domains in linguistics. Lex. denotes lexical; sem. denotes semantics; und. denotes understanding.}
\label{fig:semantics}
\end{figure}

The evolution of semantic research reflects the rich connotation of semantics in linguistics. At the early stage, much attention is given to the study of lexical semantics. The first English dictionary, \textit{Robert Cawdrey's Table Alphabeticall}, dates back to 1604~\citep{noyes1943first}. The construction of dictionaries, e.g., \textit{The Oxford English Dictionary}~\citep{Simpson1989oxford} became one of the most significant symbols of lexical semantic research achievements. The research of lexical semantics covers word senses, polysemy, word formation, contrastive lexical semantics, and more. Next, another important research dimension of semantics emerged, termed structural semantics. Structural semantics emphasizes the analysis of sentence structures, including the relationships between words and the ways in which words contribute to the meaning of a sentence. The study of structural semantics includes but is not limited to analyzing the meaning of words by syntax, grammar, and pragmatics. Structural semantics elevates the study of semantics from the word level to the sentence level. The later cognitive semantics further enrich the connotation of semantics. The tenets of cognitive semantics posit that the faculty of language is intricately intertwined with the broader cognitive capacity of human beings~\citep{croft2004cognitive}. In other words, semantics is a reflection of how humans understand and make sense of the world around them. Under cognitive semantics, researchers extend to frame semantics (semantics is the reflection of encyclopedic knowledge), situation semantics (semantics reflects the relationships between situations)~\citep{barwise1981situations}, conceptual semantics (semantics reflects the structural perception of concepts)~\citep{jackendoff1976toward}, and more. Figure~\ref{fig:semantics} summarizes partial semantic research domains in linguistics.

The development of automatic semantic processing techniques has largely facilitated semantic research. Many useful tools and knowledge bases\footnote{A knowledge base normally refers to a collection of organized information that is machine-readable, and supportive for an intelligent system.} were developed for word sense disambiguation, anaphora resolution, named entity recognition, concept extraction, and subjectivity detection. These tools are the embodiment of many theoretical ideas in semantics. For example, word sense disambiguation is an important task in lexical semantics. Anaphora resolution elucidates the relationship between the anaphor, which is the repetition of a reference, and its antecedent, which is the earlier mention of the entity. Anaphora resolution determines the structural semantics of the anaphor. Named entity recognition categorized named entities in texts by conceptually related classes, e.g., names, and locations. Similarly, concept extraction and subjectivity detection tasks also embody the cognitive properties of semantics.  

In addition to improving semantic research, semantic processing techniques can also help other downstream natural language processing (NLP) tasks with more complexity (see Table~\ref{tab:tasks_applications}). For example, subjectivity detection can be an upstream task of sentiment analysis, because subjective expressions can be further categorized by positive, negative, and neutral expressions with different opinionated intensities. The semantic processing techniques that have been reviewed possess a range of potential applications, including the ability to generate features that are effective, as well as to be used as a parser in order to obtain desired categories of text. Additionally, these techniques have the potential to improve the explainability of downstream applications.

\begin{table}[!htbp]
\scriptsize
\centering
\begin{tabular}{lccccc}
\toprule
Downstream tasks & WSD & AR & NER & CE & SD \\
\hline
Sentiment Computing & F, P, E & F & & F, P, E & P \\
Information Retrieval & E & & & F, E & P \\
Machine Translation & F, P, E & F, E & & & \\
Summarization & & F & & & \\
Textual Entailment & & F & & & \\
Knowledge Graph Construction & & & P & & \\
Recommendation Systems & & & F, P, E & & \\
Dialogue Systems & & & P, E & F, P & \\
Commonsense Explanation Generation & & & & F, E & \\
Hate Speech Detection & & & & & F, P \\
Question \& Answering Systems & & & & & F, P \\
\bottomrule
\end{tabular}
\caption{The surveyed semantic processing tasks and their downstream applications. F denotes that the technique yielded features for a downstream task model; P denotes that the technique was used as a parser; E denotes that the technique improved the explainability for a downstream task. WSD denotes word sense disambiguation. AR denotes anaphora resolution. NER denotes named entity recognition. CE denotes concept extraction. SD denotes subjectivity detection.}\label{tab:tasks_applications}
\end{table}

The emergence of pre-trained language models (PLMs) has greatly enhanced the semantic representation capabilities of deep learning models and the ability to fit downstream tasks~\citep{devlin2018bert,liu2019roberta,lewis2020bart}. Some large language models (LLMs), e.g., GPT-4\footnote{\url{https://openai.com/product/gpt-4}} and Bard\footnote{\url{https://bard.google.com/}} even realize the functions of multiple complex NLP tasks by the means of dialogue, such as question answering, translation, and text summarization. Many semantic processing studies have gradually faded out of the field of NLP. Then, in the era of PLMs and LLMs, an intuitive question is what is the motivation for studying semantic processing techniques?

As mentioned before, semantics reflects the multiple aspects of language. Besides understanding word senses, semantics is also the entrance to understanding the mechanism, and perception of language. Language intelligence encompasses more than just achieving a level of accuracy that is equivalent to or surpasses human accuracy for specific tasks. It also entails the capacity to unveil the nature of language and investigate the cognitive processes that underlie language. Much aforementioned semantic research in the context of linguistics has not been explored in computational linguistics to our best knowledge. Thus, we are motivated to propose a survey on semantic processing techniques to encourage future scholars that can expand the depth and breadth of semantic research, leading the public attention from the application value of NLP techniques to the research value of computational linguistics. Nevertheless, we also highlight the fusion of low-level semantic processing techniques and high-level NLP techniques to demonstrate the application value of semantic processing techniques in different domains.

Given the broadness of semantics, our survey scope lies in semantic processing techniques for word sense disambiguation, anaphora resolution, concept extraction, named entity recognition, and subjectivity detection. This is because these low-level semantic processing tasks reflect different aspects of semantics. In addition, there were many research works on these tasks in the field of computational linguistics. We focus on low-level semantic processing tasks, rather than high-level semantic processing tasks, e.g., sentiment analysis and natural language inference, because they provide fundamental building blocks for both high-level semantic processing tasks and higher-level NLP tasks.

\begin{figure}[!th]
\centering
\includegraphics[width=12cm, scale=0.8]{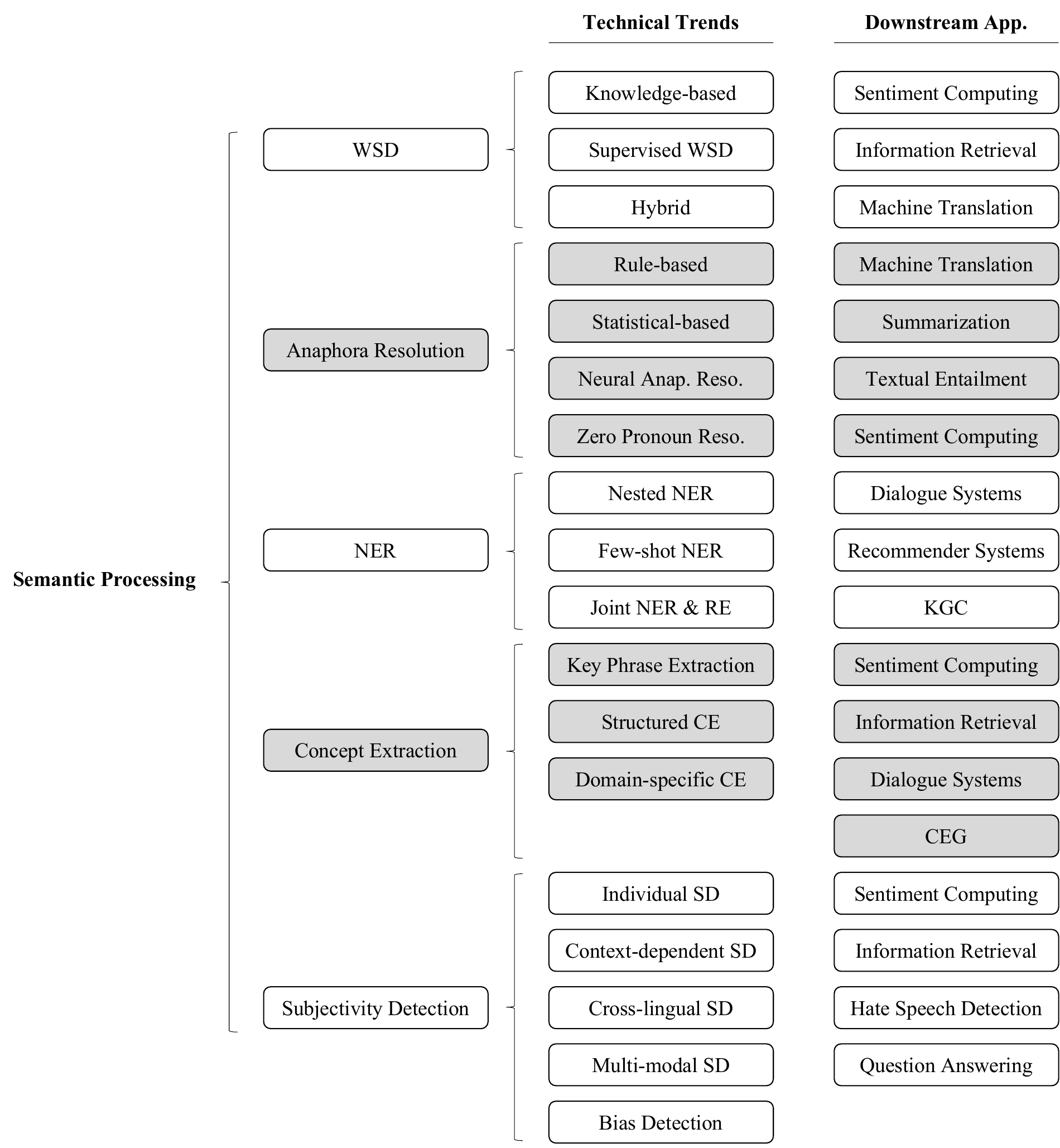}
\caption{The summary of technical trends and downstream applications of surveyed semantic processing tasks. KGC denotes knowledge graph construction. CEG denotes commonsense explanation generation. RE denotes relation extraction.}
\label{fig:taxonomy}
\end{figure}

Multiple semantic processing techniques were rarely surveyed in the same article. \citet{salloum2020survey} surveyed several high-level semantic processing tasks, e.g., latent semantic analysis, explicit semantic analysis, and sentiment analysis. Compare to the work of~\citet{salloum2020survey}, our survey includes the latest research in low-level semantic processing techniques. Compare to the latest semantic processing surveys focusing on specific tasks~\citep{ransing2022survey,poesio2023computational,fu2020clinical,wang2022nested,montoyo2012subjectivity}, we additionally reviewed important theoretical research and downstream task applications in these domains. These contents can help readers better understand the foundation of semantic research in linguistics, as well as potential application scenarios. More importantly, theoretical research shows the big picture of a semantic processing task, which may inspire different research tasks in the computational linguistic community. The collection of multiple semantic processing techniques is helpful for readers to have a comprehensive understanding of a large field, inspiring more fusion research across different domains. Theoretical research of other tasks has the potential to inspire fresh perspectives among researchers who have been concentrating on a specific semantic research task.

The contribution of this survey is threefold:
\begin{itemize}
    \item We survey recent semantic processing techniques, annotation tools, datasets, and knowledge bases for five low-level semantic processing tasks. 
    \item We highlight important theoretical research, and downstream applications to encourage deeper and wider research in the semantic processing domain upon the currently established task setups.
    \item We compare different semantic processing techniques, delineate their technical and application trends, and put forth potential avenues for future research in this domain.
\end{itemize}

In the following sections, we introduce different semantic processing techniques, e.g., word sense disambiguation (Section~\ref{sect:Word Sense Disambiguation}), anaphora resolution (Section~\ref{sect:Anaphora Resolution}), named entity recognition (Section~\ref{sect:Named Entity Recognition}), concept extraction (Section~\ref{sect:Concept Extraction}), and subjectivity detection (Section~\ref{sect:Subjectivity Detection}). We discuss the interactions between the surveyed tasks and the impacts of deep learning and LLMs on semantic processing in Section~\ref{sect:discussion}. Finally, we conclude this survey in Section~\ref{sect:Conclusion}. Each task is structured by theoretical research, annotation schemes, datasets, knowledge bases, evaluation metrics, methods, downstream applications, and a summary. Figure~\ref{fig:taxonomy} demonstrates the taxonomy of methods and downstream applications of each task in this survey.

\section{Word Sense Disambiguation}
\label{sect:Word Sense Disambiguation}

The complexity of human language is difficult for machines to understand it. One of the challenges is the ambiguity of word senses. In natural language, a word may have multiple senses, given different contexts. Consider the following example: 

\ex. He got his shoes wet as he walked along the \textbf{bank}.

According to the Oxford English Dictionary, the major senses of ``bank'' include (a) \textit{an organization that provides various financial services, for example keeping or lending money}; (b) \textit{the side of a river, canal, etc. and the land near it}. With the context, humans can easily know that ``bank'' here refers to the sense (b). However, it is challenging for machines to do so because the interpretation made by humans is contingent upon their comprehension of the fact that the probability of getting one's shoes wet is higher when walking alongside a river bank as compared to a financial institution. Machines rarely take the commonsense into account when inferring the meaning of ``bank''\footnote{Current methods likely disambiguate word senses by word co-occurrences. However, word co-occurrences are not commonsense.}, because they don't have human-like cognition and reasoning abilities by nature.

\begin{figure}[t]
\centering
\includegraphics[scale=0.5]{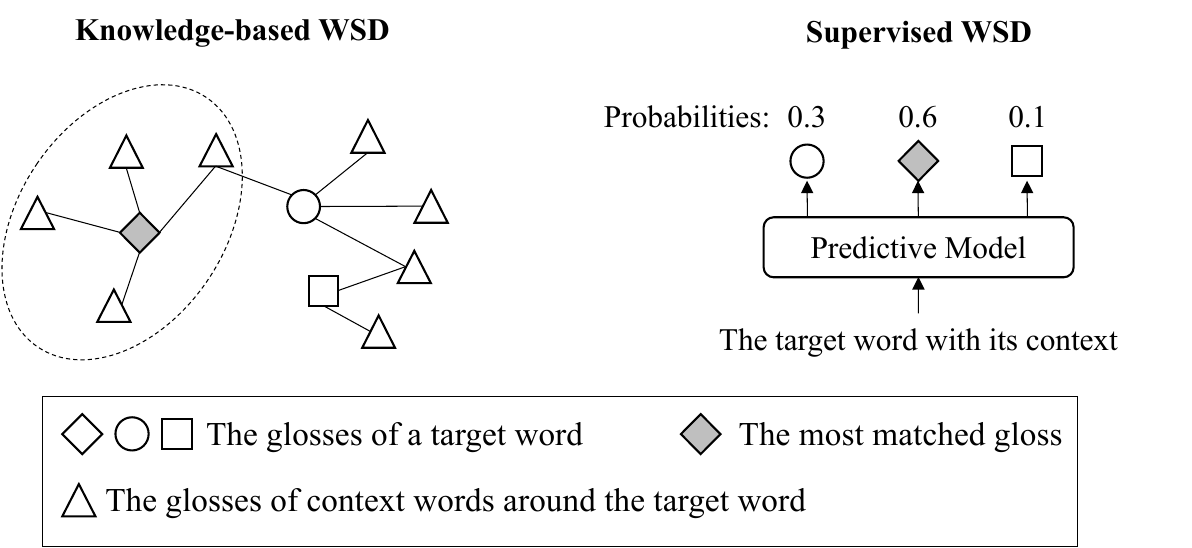}
\caption{Simplified examples of the knowledge-based and supervised WSD.}
\label{fig: knowledge-based and supervised WSD}
\end{figure}

There are two main technical trends in addressing the task of WSD, namely knowledge-based methods and supervised methods. Knowledge-based WSD utilizes the word relations from knowledge graphs, e.g., WordNet and BabelNet~\citep{navigli2012babelnet} to achieve the disambiguation of word senses. In supervised methods, the WSD task is usually defined as a classification task by word senses. A WSD model is trained with annotated data. Two examples of knowledge-based and supervised WSD are illustrated in Figure~\ref{fig: knowledge-based and supervised WSD}. As shown in the figure, a naive strategy of the knowledge-based WSD is that the sense that shares the most relations with the context words is selected as the best-matched one. For supervised WSD systems, the predictive model predicts the potential senses, given the target word and its context words as input. In recent times, the use of knowledge bases has proven advantageous for several modern supervised systems. As a result, there has been a growing trend in integrating knowledge-based and supervised methods to enhance their performance \citep{wang2020synset}.

WSD has been recognized as a crucial module in numerous NLP tasks that heavily rely on word senses, such as sentiment computing, information retrieval, and machine translation. The application of WSD techniques has been demonstrated to be beneficial for these NLP tasks. While prior surveys \citep{DBLP:conf/ijcai/BevilacquaPRN21, navigli2009word} have conducted extensive reviews for WSD, the works discussed in them are outdated. Besides, those works do not link WSD with the linguistic theories and diverse downstream tasks.

\subsection{Theoretical Research}\label{sec:theoretical_research_wsd}

\subsubsection{Distributional Semantics}

The hypothesis from distributional semantics~\citep{firth1957synopsis} argued that word meanings can be inferred from word co-occurrences. Words that appear in similar contexts tend to have similar meanings. Such a hypothesis has been the most significant foundation of developing semantic representations in the computational linguistics community, e.g., vector space representations~\citep{turney2010frequency,mikolov2013distributed,pennington2014glove} and PLMs~\citep{devlin2018bert,liu2019roberta}. Based on such a hypothesis, dense semantic vectorial representation research commonly follows a similar training paradigm, e.g., using context words to predict a target word. Currently, ChatGPT further proves that learning to use words that have appeared before to predict the next possible word can achieve the skills of analogy and reasoning with the help of a very large Transformer~\citep{vaswani2017attention}-based model.

\subsubsection{Selectional Preference}

\citet{wilks1973preference} proposed a concept of selectional preference. It is a procedure for representing the meaning structure of natural language. Compared to the ``derivational paradigm'' of transformational grammar and generative semantics,~\citet{wilks1973preference} believed that selectional preference is a more efficient procedure in natural language understanding. It focuses on determining preferences between various possible interpretations of a text, rather than identifying a solitary and unequivocally correct interpretation. Selection preference theory allows more flexibility and nuance in understanding word senses and language. Besides, the theory is computation-friendly. \citet{wilks1973preference} showed how the procedure could be computed and implemented. The work of~\citet{wilks1973preference} supports that there are multiple possible meanings for a word. The meaning can be defined by the sectional preference of contexts.

\subsubsection{Construction Semantics}

\citet{goldberg2010construction} argued that the meanings of words are frequently derived from larger language units, termed constructions. Constructions consist of a form and a meaning, ranging from single words to full sentences in size. The interpretation of a construction is reliant on both its structure and the situations in which it is employed. \citet{goldberg2010construction} argued that semantic restrictions are better linked with the construction as an entirety rather than with the lexical semantic framework of the verbs. The work of~\citet{goldberg2010construction} highlights that the interpretation of meanings of language units can be extended from individual words to constructions. It shows the necessity of defining language units in WSD.

\subsubsection{Frame Semantics}\label{sect:frame semantics}

\citet{fillmore2006frame} proposed frame semantics that provides a distinct viewpoint on the meanings of words and the principles behind language construction. Frame semantics emphasizes the significance of the surrounding context and encyclopedic knowledge in comprehending word meanings. \citet{petruck1996frame} explained that a ``frame'' refers to a collection of concepts interconnected in a manner that understanding any one concept depends on the understanding of the complete system. In frame semantics, the meaning of ``cooking'' is beyond its dictionary meaning. It also associates with the concept of ``food'', ``cook'', ``container'', and ``heating instrument''. Frame semantics motivates later ontology research, e.g., FrameNet~\citep{ruppenhofer2016framenet} and FrameNet-based WSD systems, significantly.

\subsection{Annotation Schemes}\label{sec:annotation_schemes_wsd}

For knowledge-based WSD, the data are normally presented as ontology, such as WordNet, FrameNet, and BabelNet, where words and concepts are connected by relations. The relations include hyponyms, hypernyms, holonyms, meronyms, attributes, entailment, etc. An explanation (gloss) and a few example sentences are given for each synset. Synsets of the same Part Of Speech (POS) are connected under some relations independently. However, there exist relations when the basic concept of two words is the same but in a different POS (for example, ``propose'' and ``proposal'' were characterized as ``derivationally related synsets'' in WordNet).

For supervised WSD, a particular word in a given sentence is annotated with a sense ID that corresponds to one of the potential senses in a knowledge base, such as WordNet. A sample of annotation is shown in the next section.

\subsection{Datasets}\label{sec:datasets_wsd}

\begin{table}[!tbh]
\scriptsize
\centering
\begin{tabular}{@{}llrl@{}}
\toprule
Dataset & Source & \# Samples & Reference \\ \midrule
SemCor  & WordNet & 200,000  &~\citet{miller1993semantic} \\
MultiSemCor & \begin{tabular}[c]{@{}l@{}}WordNet, \\ bilingual Collins\end{tabular} & 51,847 &~\citet{pianta2002multiwordnet} \\
Line-hard-serve & WSJ, APHB  & 4,000  &~\citet{leacock1993corpus}  \\
Interest & HECTOR  & 2,369  &~\citet{bruce1999decomposable} \\
DSO & Brown, WSJ & 192,800  &~\citet{ng1996integrating} \\
OMWE & Web & 29,165 &~\citet{chklovski2004verbocean} \\ 
OMSTI & UN documents & 1,357,922 &~\citet{taghipour2015one} \\
SensEval-2 & Unknown & 2,282 &~\citet{edmonds2001senseval} \\
SensEval-3 & \begin{tabular}[c]{@{}l@{}}Editorial, news\\story, \& fiction\end{tabular} & 1,850 &~\citet{snyder2004english} \\
SemEval2007 & Brown, WSJ & 455 &~\citet{pradhan2007semeval}\\
SemEval2013 & SMT workshop & 1,644 &~\citet{navigli2013semeval} \\
SemEval2015 & \begin{tabular}[c]{@{}l@{}}EMEA, KDEdoc,\\EUB\end{tabular} & 1,022 &~\citep{moro2015semeval}\\
\bottomrule
\end{tabular}
\caption{WSD datasets and statistics. SMT, EMEA, KDEdoc, and EUB denote statistical machine translation, European Medicines Agency documents, KDE manual corpus, and the EU bookshop corpus, respectively.}\label{tab:wsd dataset statistics}
\end{table}

Our surveyed datasets and their statistics can be viewed in Table~\ref{tab:wsd dataset statistics}. The biggest manually annotated English corpus currently accessible is SemCor\footnote{\url{http://web.eecs.umich.edu/~mihalcea/downloads.html}}~\citep{miller1993semantic}. It has 200K content terms tagged with their related definitions and around 40K phrases. Although SemCor serves as the principal training corpus for WSD, its limited coverage of the English vocabulary for both words and meanings is its most significant drawback. In essence, SemCor merely includes annotations for 22K distinct lexemes in WordNet, the most extensive and commonly employed computerized English dictionary, which corresponds to less than 15\% of all words.

To augment the coverage of words, some studies~\citep{vial2019sense} incorporated the English Princeton WordNet Gloss Corpus (WNG)\footnote{\url{https://wordnetcode.princeton.edu/glosstag.shtml}}, which contains more than 59K WordNet senses, as a complemented data. The WNG is annotated manually or semi-automatically.

SemCor and its variations~\citep{bentivogli2005exploiting, bond2012japanese} lack an acceptable multi-lingual equivalent in the majority of global languages, which limits the scaling capabilities of WSD models beyond English. To address the aforementioned issues, numerous automatic methods for creating multi-lingual sense-annotated data have been developed~\citep{pasini2017train, pasini2018huge, scarlini2019just, pasini2020train}. In an English-Italian parallel corpus known as MultiSemCor~\citep{pianta2002multiwordnet}, senses from the English and Italian versions of WordNet are annotated. 

The Line-hard-serve corpus~\citep{leacock1993corpus} contains 4K samples of the nominal, adjective, and verbal words with sense tags. The data were sourced from Wall Street Journal (WSJ) corpus and the American Printing House for the Blind (APHB) corpus. The Interest corpus~\citep{bruce1999decomposable} contains 2,369 occurrences of the term \textit{interest} that have been sense-labeled. The data were sourced from the HECTOR word sense corpus~\citep{atkins1992tools}. The Defence Science Organisation (DSO), based in Singapore, created the DSO corpus\footnote{\url{https://borealisdata.ca/dataset.xhtml?persistentId=doi:10.5683/SP2/QPOJSI}}~\citep{ng1996integrating}, which contains 192,800 sense-tagged tokens from 191 words from the Brown and WSJ corpora. The Open Mind Word Expert (OMWE) dataset\footnote{\url{http://web.eecs.umich.edu/~mihalcea/downloads/OMWE/OMWE1.0.English.tar.gz}}~\citep{chklovski2004verbocean} is a corpus of sentences with 288 noun occurrences that were jointly annotated by Web users. One Million Sense-Tagged for Word Sense Disambiguation and Induction (OMSTI)\footnote{\url{https://www.comp.nus.edu.sg/~nlp/corpora.html}}~\citep{taghipour2015one} is a semi-automatically annotated WSD dataset with WordNet sense inventory. The data were sourced from MultiUN corpus, which is a collection of United Nation documents. The OMSTI includes 687,871 nouns, 412,482 verbs, and 251,362 adjectives and 6,207 adverbs after including selected samples from SemCor and DSO.

The SensEval and SemEval datasets are created from the SensEval/SemEval evaluation campaigns. Now, these datasets have been the most widely used benchmarking datasets in WSD. \citet{raganato2017word} collected these datasets together\footnote{\url{http://lcl.uniroma1.it/wsdeval/home}} and developed a unified evaluation framework for empirical comparison. The statistics of the following datasets are from the collection of~\citet{raganato2017word}. SensEval-2~\citep{edmonds2001senseval} used WordNet 1.7 sense inventory, including 2,282 sense annotations for nouns, verbs, adverbs and adjectives. SensEval-3~\citep{snyder2004english} employed WordNet 1.7.1 sense inventory, including 1,850 sense annotations.  SemEval-2007 Task 17~\citep{pradhan2007semeval} employed WordNet 2.1 sense inventory, including 455 nominal and verbal sense annotations. SemEval-2013 Task 12~\citep{navigli2013semeval} used WordNet 3.0 sense inventory, including 1,644 nominal sense annotations. SemEval-2015 Task 13~\citep{moro2015semeval} utilized WordNet 3.0 sense inventory, including 1,022 sense annotations. It is worth noting that some of the SemEval tasks are multi-lingual, including SemEval 2013 and 2015, which facilitates multi-lingual WSD. 

All of these corpora are annotated using various WordNet sense inventories, with the exception of the Interest corpus (tagged with LDOCE senses) and the Senseval-1 corpus. The Interest corpus and the Senseval-1 corpus were sense-labeled using the HECTOR sense inventories, a lexicon and corpus from a joint Oxford University Press/Digital project~\citep{atkins1992tools}

Generally, the data and labels in WSD datasets are organized in the following forms. Then, the task is to identify the sense classes, given contexts, and target words.\\

\begin{mdframed}
\noindent\texttt{\scriptsize{{context: "You perform well in the exam, I will reward you.",\\
target word: "perform",\\
pos: "VB",\\
sense: "3"\\}}}

\noindent\texttt{\scriptsize{{context: "She worked in a renowned university for a long time.",\\
"target word": "university",\\
"pos": "NN",\\
"sense": "2"}}}
\end{mdframed}

\subsection{Knowledge Bases}\label{sec:knowledgebases_wsd}

\begin{table}[!htbp]
\centering
\scriptsize
\begin{tabular}{@{}llrl@{}}
\toprule
Name & Knowledge & \# Entities & Structure \\ \midrule
LDOCE 6th ed. & Lexical & 230,000 & Unstructured  \\
ODE 2022  & Lexical & 600,000 & Unstructured \\
CED 12th ed.  & Lexical & 722,000 & Unstructured  \\
OALD 8th ed.  & Lexical & 145,000 & Unstructured  \\
WordNet   & Lexical & 95,600 & Graph \\
FrameNet & Lexical & 13,687 & Graph \\
BabelNet  & Lexical \& Multi-lingual & 26,044,643 & Graph \\ 
SyntagNet  & Lexical & 78,000 & Graph \\ 
\bottomrule
\end{tabular}
\caption{Useful knowledge bases for WSD. LDOCE means Longman Dictionary of Contemporary English. ODE means Oxford Dictionary of English. CED means Collins English Dictionary. OALD means Oxford Advanced Learner’s Dictionary of Current English. Unstructured or structured means the knowledge base contains unstructured or structured lexical knowledge by concepts. }
\end{table}

\textbf{Machine-Readable Dictionaries (MRDs)} have been a useful source for WSD due to their structured knowledge and easy access~\citep{navigli2009word}. Dictionaries frequently contain extensive information about the various meanings of a word, as well as illustrative examples of their usage within context. Therefore, dictionaries can serve as valuable knowledge bases for the task of WSD. Additionally, MRDs may provide further information such as synonyms, antonyms, and related words, which can aid in facilitating a more comprehensive comprehension of a word's meaning. Through the analysis of this information, a system may make more precise determinations about which meaning is most fitting in a given context. There are many electronic dictionaries available for machines to refer to, such as the Longman Dictionary of Contemporary English (LDOCE)~\citep{mayor2009longman}, the Oxford Dictionary of English (ODE)~\citep{dictionary2010oxford}, Collins English Dictionary (CED)~\citep{dictionary1982collins}, and the Oxford Advanced Learner’s Dictionary of Current English (OALD)~\citep{hornby1974oxford}.

\noindent\textbf{WordNet}~\citep{miller1990introduction} is a sizable, manually curated lexicographic database of English. It is arranged as a network with synsets, or collections of contextual synonyms, as nodes. A synset of synonyms each represents one of a word's senses. Through edges that express lexical-semantic links like meronymies (partof) and hypernymies (is-a), synsets and senses are connected to one another. WordNet additionally offers definitions (glosses) and uses examples for each synset as additional lexical information. The most current English WSD works use the 3.0 version, which was published in 2006 and has 117,659 synsets. Following the initial WordNet for English, many WordNets for other languages have been proposed, including languages such as Chinese~\citep{wang-bond-2013-building}, Arabic~\citep{black2006introducing}, Dutch~\citep{postma2016open}, etc\footnote{See \url{http://globalwordnet.org/resources/wordnets-in-the-world/} for a summary.}.

\noindent\textbf{FrameNet}~\citep{ruppenhofer2016framenet} is an English lexical repository that is readable by both humans and machines, established by annotating real-life textual examples that depict the usage of words. It was developed based on the theory of frame semantics, containing 1,224 frames (a frame refers to a diagrammatic representation of a scenario encompassing diverse elements such as participants, props, and other conceptual roles), and 13,687 lexical units (lemmas and their PoS) that evoke frames. In FrameNet, the lexical units of a sentence are associated with frame elements. Frame elements are the semantic role of lexical units. For example, given a sentence ``I ate an apple this afternoon'', ``apple'' would fill the role of ``food'' (a frame element).

\noindent\textbf{BabelNet}~\citep{navigli2012babelnet} is a multi-lingual dictionary that covers both lexicographic and encyclopedic entries from 520 languages. These entries were created by semi-automatically mapping numerous sites, including WordNet, Multi-lingual WordNet, and Wikipedia. The topology of BabelNet is that of a semantic network, where the nodes are multi-lingual synsets (collections of synonyms that have been lexicalized in several languages), and the edges represent the semantic connections between them.

\noindent\textbf{SyntagNet}~\citep{maru2019syntagnet} is a manually developed lexical resource that integrates semantically disambiguated lexical combinations, e.g., noun-verb and noun-noun pairs. The development of SyntagNet involved initially extracting lexical combinations from English Wikipedia and the British National Corpus, which were then subjected to a process of manual disambiguation, based on the WordNet. SyntagNet covers five major languages, e.g., English, German, French, Spanish, and Italian.

\subsection{Evaluation Metrics}\label{sec:evaluation_metrics_wsd}

In the WSD task, given a sentence of \textit{n} words $T = \{x_1, ..., x_n\}$, the model predicts a sense for each word given the dictionary. Normally, the F1 score is adopted, which is a specialization of the F score when $\alpha = 1$:

\begin{equation}
    F = \frac{1}{\alpha \frac{1}{P} + (1-\alpha) \frac{1}{R}}
\end{equation}

Where $P$ denotes precision and $R$ denotes recall:

\begin{equation}
    P = \frac{\text{correct predictions}}{\text{total predictions}}
\end{equation}

\begin{equation}
    R = \frac{\text{correct predictions}}{n}
\end{equation}

The aforementioned metrics do not accurately represent how well systems can produce a level of confidence for a particular sensory choice. \citet{resnik1999distinguishing} developed an evaluation criterion that considers the discrepancies between the accurate and selected senses to weigh misclassification mistakes. Therefore, this error will be penalized less severely than coarser sense distinctions if the chosen sense is a fine-grained distinction of the true sense. There have been evaluation metrics for even more precise measurements, including the Receiver Operation Characteristic (ROC)~\citep{cohn2003performance}. However, compared with traditional metrics such as precision, recall, and F1, these metrics are not frequently utilized.

\subsection{Annotation Tools}\label{sec:annotation_tools_wsd}

\noindent \textbf{LX-SenseAnnotator}\footnote{\url{http://nlx.di.fc.ul.pt/tools.html}}~\citep{neale2015flexible} provides a user interface for manually annotating word senses. The software has the capability to process lexical data in any language, on the condition that the data is compliant with the format of Princeton WordNet. Human annotators can view the pre-processed text in three different modes, including the source text, sense-annotated text, and raw text, which can be switched between by using a tab widget. The source text mode displays the original text along with all tags, while the sense-annotated text mode displays the same text but with newly added sense tags. This allows the annotator to monitor the output file continually. Annotators can view the sense options in real time when annotating the sense for a word. 

\noindent \textbf{LexTag}\footnote{https://babelscape.com/lextag} is another useful tool for WSD. The annotation interface provided is characterized by its user-friendly nature, facilitating users in the annotation of various textual elements such as terms, sentences, and documents. This annotation process involves attributing meanings drawn from pre-existing knowledge graphs and dictionaries, encompassing reputable sources like WordNet, Wiktionary, and WordAtlas. LexTag has been used to create a recent 10-language parallel dataset ELEXIS-WSD 1.0\footnote{https://www.clarin.si/repository/xmlui/handle/11356/1674}.

\subsection{Methods}\label{sec:methods_wsd}

\subsubsection{Knowledge-Based WSD}\label{sec:task1_wsd}
Knowledge-based WSD utilizes knowledge bases to disambiguate word senses. Compared with supervised WSD, this class of WSD methods achieves lower performance but better data efficiency. In knowledge-based WSD, there are essentially two research streams.

\noindent \textbf{A. Semantic Space Matching}

One stream of the knowledge-based WSD is to look for overlaps or similarities between the context of a term whose sense needs to be disambiguated and its sense representation, such as the definition of a potential sense and its associated sense that was retrieved from a knowledge base. The predicted sense is considered to be the sense that is the closest. 

Lesk~\citep{lesk1986automatic} is a naive knowledge-based WSD algorithm that looks for terms that are similar to the target word in the context of each sense. The approach aimed to enumerate the intersections among lexicon definitions of the diverse connotations of every target word contained within a given sentence. \cite{banerjee2003extended} proposed an advanced version of the Lesk, which also includes the definition of related senses, where the standard term frequency-inverse document frequency method is employed for word weighting. Another improved version of Lesk~\citep{basile2014enhanced} includes word embedding for better analysis, which improves the accuracy of determining how close the definition and context of the target word are. SREF$_{KB}$~\citep{wang2020synset} is a state-of-the-art (SOTA) WSD system. It is a vector-based technique that disambiguates word senses by using sense embeddings and contextualized word representations. It applied BERT to represent WordNet instances and definitions, as well as the automatically obtained contexts from the Web.

\noindent \textbf{B. Graph-based Matching}

The other stream of the knowledge-based WSD creates a graph using the given context and connections that have been retrieved from knowledge bases. Here, the synsets and the relationships between them are seen as the nodes and edges, respectively. The senses are then disambiguated based on the constructed graphs. A variety of graph-based techniques, such as Latent Dirichlet Allocation (LDA)~\citep{blei2003latent}, PageRank~\citep{brin1998anatomy}, Random Walks~\citep{agirre2014random}, Clique Approximation~\citep{moro2014entity}, Game Theory~\citep{tripodi2019game}, etc., are used to disambiguate the meaning of a given word using the created graph.

\cite{agirre2009personalizing} presented a graph-based unsupervised WSD system that employs random walk over a WordNet semantic network. They employed a customized version of the Page Rank algorithm~\citep{haveliwala2002topic}. The technique leverages the inherent structural properties of the graph that underlies a specific lexical knowledge base, and shows the capability of the algorithm to identify global optima for WSD, based on the relations among entities. \citet{agirre2014random} evaluated this algorithm with new datasets and variations of the algorithm to prove its effectiveness. \citet{navigli2007graph} also introduced a graph-based unsupervised model for WSD, which analyzed the connectivity of graph structures to identify the most pertinent word senses. A graph is constructed to represent all possible interpretations of the word sequence, where nodes represent word senses and edges represent sense dependencies. The model assessed the graph structure to determine the significance of each node, thus finding the most crucial node for each word. Babelfy~\citep{moro2014entity} is also a graph-based WSD method that uses random walk to identify relationships between synsets. It used BabelNet~\citep{navigli2012babelnet} and performed random walks with Restart~\citep{tong2006fast}. In addition, it incorporated the entire document at the time of disambiguation. The candidate disambiguation is upon automatically developed semantic interpretation graph which used a graph structure to represent various possible interpretations of input text. SyntagRank~\citep{scozzafava2020personalized} is a high-scoring knowledge-based WSD algorithm. It is an entirely graph-based algorithm that uses the Personalized PageRank algorithm to incorporate WordNet (for English), BabelNet (for non-English) and SyntagNet. SyntagRank is generally considered a stronger method than SREF$_{KB}$. BabelNet enabled SyntagRank to improve its ability to scale across a wide range of languages, whereas SREF$_{KB}$ has only been evaluated in English.

\subsubsection{Supervised WSD}\label{sec:task2_wsd}
Currently, supervised approaches, especially deep learning-based supervised learning approaches, have become mainstream in the WSD community. Earlier deep learning-based approaches focused on architectures where WSD was defined as token classification over WordNet senses~\citep{kaageback2016word}. Even though they performed well, these structures showed a lot of flaws, particularly when it came to predicting uncommon and invisible senses. To address these issues, numerous works began to supplement the training data by utilizing various lexical knowledge, such as sense definitions~\citep{kumar2019zero,blevins2020moving}, semantic relations~\citep{bevilacqua2020breaking, conia2021framing}, and data generated via novel generative methods~\citep{barba2021exemplification}. In this section, we review the representative works in supervised WSD.

\noindent \textbf{A. Data-Driven Machine Learning Approaches}

Data-driven machine learning approaches refer to methodologies and techniques in which the design, training, and optimization of traditional machine learning algorithms, heavily rely on large amounts of data. In these approaches, the model's ability to generalize patterns and make predictions is learned directly from the provided data, rather than being explicitly programmed by humans. In the early days, classic machine learning approaches with handcrafted features were frequently used for WSD.

\citet{singh2014decision} employed 5-gram and position features, and a decision tree algorithm to represent classification rules in a tree structure where the training dataset is recursively partitioned. Each leaf node indicates the meaning of a word. They developed a dataset, containing 672 Manipuri sentences to test their method. The sentences were sourced from a local newspaper, termed ``The Sangai Express''. \cite{o2004class} proposed a class-based collocation method that integrates diverse linguistic features in a decision tree algorithm. For the collocation, three distinct word relatedness scores are used: the first is based on WordNet hypernym relations; the second is based on cluster-based word similarity classes; and the third is based on dictionary definition analysis. The authors also utilized PoS and word form features. The It Makes Sense (IMS) WSD system~\citep{zhong2010makes} used a Support Vector Machine (SVM) classifier. Different positional and linguistic features were considered, including nearby words, nearby words' PoS tags, and nearby collocations. Later, word embeddings became important features in WSD. \citet{taghipour2015one, rothe2015autoextend, iacobacci2016embeddings} used IMS as the base model to examine word embeddings. \cite{iacobacci2016embeddings} offered many approaches where different word embeddings were applied as features to test how many parameters impact the effectiveness of a WSD system. The authors found that word2vec~\citep{mikolov2013distributed} which was trained with OMSTI can yield the strongest results on the three examined all-word WSD tasks.

\noindent \textbf{B. Data-Driven Neural Approaches}

More recently, neural approaches started to be used. Data-driven neural approaches refer to methodologies and techniques that utilize neural networks and supervised learning to learn patterns and representations directly from data.

\cite{popov2017word} proposed to use BiLSTM~\citep{graves2005framewise}, GloVe word embeddings, and word2vec lemma embeddings. \cite{yuan2016semi} suggested another LSTM-based word sense disambiguation approach that was trained in a semi-supervised fashion. The semi-supervised learning was achieved by employing label propagation~\citep{talukdar2009new} to assign labels to unannotated sentences by assessing their similarity to labeled ones. The best performance on the SensEval-2 dataset can be observed from the model that was semi-supervision-trained with OMSTI and 1,000 additional unlabeled sentences. Additionally,~\cite{le2018deep} looked more closely at how many elements affect its performance, and several intriguing conclusions were drawn. The initial point to highlight is that achieving strong WSD performance does not necessitate an exceedingly large unannotated dataset. Furthermore, this method provides a more evenly-distributed sense assignment in comparison to prior approaches, as evidenced by its relatively strong performance on infrequent cases. Additionally, it is worth noting that the limited sense coverage of the annotated dataset may serve as an upper limit on overall performance. 

With the development of self-attention-based neural architectures and their capacity to extract sophisticated language information~\citep{vaswani2017attention}, the use of transformer-based architectures in fully supervised WSD systems is becoming more and more popular. The WSD task is usually fine-tuned on a pre-trained transformer model, which is a popular strategy. The task-specific inputs are given to the pre-trained model, which is then further trained across a number of epochs with the task-specific objective. Likewise, in recent token classification models for WSD, the contextualized representations are usually generated by a pre-trained model and then fed to either a feedforward network~\citep{hadiwinoto2019improved} or a stack of Transformer layers~\citep{bevilacqua2019quasi}. These methods outperform earlier randomly initialized models~\citep{raganato2017neural}. \citet{hadiwinoto2019improved} tested different pooling strategies of BERT, e.g., last layer projection, weighted sum of hidden layers, and Gated Linear Unit~\citep{dauphin2017language}. The best performance on SensEval-2 is given by the strategy of the weighted sum of hidden layers, accounting for 76.4\% F1. \citet{bevilacqua2019quasi} proposed a bi-directional Transformer that explicitly attends to past and future information. This model achieved 75.7\% F1 on SensEval-2 by training with the combination of SemCor and WordNet’s Tagged Glosses\footnote{\url{https://wordnetcode.princeton.edu/glosstag.shtml}}. It is worth noting that, the categorical cross-entropy, which is frequently utilized for training, limits the performances. In reality, it has been demonstrated that the binary cross-entropy loss performs better~\citep{conia2021framing} because it enables the consideration of many annotations for a single instance in the training set as opposed to the use of a single ground-truth sense alone. In the above-mentioned approaches, each sense is assumed to be a unique class, and the classification architecture is limited to the information provided by the training corpus.

\subsubsection{Knowledge-augmented Supervised WSD}\label{sec:task3_wsd}

The edges that connect the senses and synsets are a valuable source of knowledge that augments the annotated data. Traditionally, graph knowledge-based systems, such as those based on Personalized PageRank~\citep{scozzafava2020personalized}, have taken advantage of this information. Moreover, utilizing WordNet as a graph has benefited many modern supervised systems. Thus, formally, knowledge-augmented supervised WSD is defined as a methodology that combines traditional supervised machine learning techniques with external knowledge resources to improve the accuracy and performance of word sense disambiguation.

\cite{wang2020synset} used WordNet hypernymy and hyponymy relations to devise a try-again mechanism that refines the prediction of the WSD model. The SemCor corpus was utilized to acquire a supervised sense embedding for every annotated sense in their supervised method (SREF$_{Sup}$).  \cite{vial2019sense} reduced the number of output classes by mapping each sense to an ancestor in the WordNet taxonomy, then yielding a smaller but robust sense vocabulary. The authors used BERT contextualized embeddings. By training with SemCor and WordNet gloss corpora, the model achieved 79.7\% F1 on SensEval-2. Different variations also achieve outstanding performance on diverse WSD datasets.

\cite{loureiro2019language} created representations for those senses not appearing in SemCor by using the averaged neighbor embeddings in the WordNet. The token-tagger models EWISE~\citep{kumar2019zero} and EWISER~\citep{bevilacqua2020breaking} both leveraged the WordNet graph structure to train the gloss embedding offline, where EWISER demonstrated how the WordNet entire graph feature can be directly extracted. EWISE used ConvE~\citep{dettmers2018convolutional} to obtain graph embeddings. \citet{conia2021framing} provided a new technique to use the same edge information by replacing the adjacency matrix multiplication with a binary cross-entropy loss where other senses connected to the gold sense are also taken into account. The edge information was obtained from WordNet. In general, edge information is increasingly used in supervised WSD, gradually blending with knowledge-based techniques. However, it can only be conveniently utilized by token classification procedures, whereas its incorporation into sequence classification techniques has not yet been researched.

It has also been extensively studied how to use sense definitions as an additional source for supervised WSD apart from the traditional data annotations. It considerably increased the scalability of a model on the senses that are underrepresented in the training corpus. \cite{huang2019glossbert} argued that WSD has traditionally been approached as a binary classification task, whereby a model must accurately decide if the sense of a given word in context aligns with one of its potential meanings in a sense inventory, based on the provided definition. define the WSD task as a sentence-pair classification task, where the WordNet gloss of a target word is concatenated after an input sentence. \citet{blevins2020moving} used a bi-encoder to project both words in context and WordNet glosses in a common vector space. Disambiguation is then carried out by determining the gloss that is most similar to the target word. Glosses are employed similarly by more advanced techniques like SensEmBERT~\citep{scarlini2020sensembert}, ARES~\citep{scarlini2020more}, and SREF~\citep{wang2020synset}. They used quite different approaches to find new contexts automatically in order to develop the supervised portion of the sense embedding. ARES achieved 78.0\% F1 on the SensEval-2 dataset by utilizing collocational relations between senses to get novel example sentences from websites. SensEmBERT leveraged BabelNet and Wikipedia explanations, achieving significant improvements on nominal WSD tasks over 5 major datasets. \citet{barba2021esc} proposed to solve WSD as a text extraction problem where, given a word in context and all of its potential glosses, models extract the definition that best matches the term under consideration. The authors demonstrated the advantages of their approach in that it does not require huge output vocabularies and enables models to take into account both the input context and all meanings of the target word simultaneously. By using sparse coding,~\cite{berend2020sparsity} has demonstrated that it is also possible to make existing sense embeddings sparse. All of these methods handle each word independently of the others when disambiguating multiple words that co-occur in the same context. Thus, a word's explicit meaning is neither taken into account during word disambiguation nor does it have an impact on the disambiguation of surrounding words.

\subsection{Downstream Applications}\label{sec:downstream_applications_wsd}

\subsubsection{Sentiment Computing}\label{sec:Sentiment Analysis}
WSD has been applied in many Sentiment Analysis (SA) works to improve accuracy and explainability. \cite{farooq2015word} proposed a WSD framework to enhance the performance of sentiment analysis. To determine the orientation of opinions related to product attributes in a particular field, a lexical dictionary comprising various word senses is developed. The process involves extracting relevant features from product reviews and identifying opinion-bearing texts, followed by the extraction of words used to describe the features and their contexts to form seed words. These seed words, which consist of adjectives, nouns, verbs, and adverbs, are manually annotated with their respective polarities, and their coverage is extended by retrieving their synonyms and antonyms. WSD was utilized to identify the sentiment-orientated senses, such as the positive, negative, or neutral senses of a word in a sentence, because a word may have different sentiment polarities by taking different senses in different contexts. 

\cite{nassirtoussi2015text} offered a novel approach to forecast intra-day directional movements of the EUR/USD exchange rates based on news headline text mining in an effort to address semantic and sentiment components of text-mining. They evaluated news headlines semantically and emotionally using the lexicons, e.g., WordNet and SentiWordNet~\citep{baccianella2010sentiwordnet}. SentiWordNet is a publicly accessible lexical resource designed for sentiment analysis that allocates a positivity score, negativity score, and objectivity score to each synset within WordNet. \cite{nassirtoussi2015text} found that both positive and negative emotions may influence the market in the same way. WSD worked as a technique to abstract semantic information in their framework. Thus, it enhances the feature representations and explainability in their downstream task modeling. SentiWordNet has served as a basis for various sentiment analysis models. In the work of~\citet{ohana2009sentiment}, the feasibility of using the emotional scores of SentiWordNet to automatically classify the sentiment of movie reviews was examined. Other applications, e.g., business opinion mining~\citep{saggionalpha2010interpreting}, article emotion classification~\citep{devitt2007sentiment}, word-of-mouth sentiment classification~\citep{hung2013using,hung2016word} also showed that SentiWordNet as a semantic feature enhancement knowledge base can deliver accuracy gains and model insights in sentiment analysis tasks.

\subsubsection{Information Retrieval}\label{sec:Information Retrieval}
The impacts of using WSD for information retrieval have been examined in many works. \citet{krovetz1992lexical} disambiguated word senses for terms in queries and documents to examine how ambiguous word senses impact information retrieval performance. The researchers arrived at the conclusion that the advantages of WSD in information retrieval are marginal. This is due to the fact that query words have uneven sense distributions. The impact of collocation from other query terms already plays a role in disambiguation. WSD was used as a parser to study this task. However, the findings from~\citet{gonzalo1998indexing} are different. They examined the impact of improper disambiguation using SemCor. By accurately modeling documents and queries together with synsets, they achieved notable gains (synonym sets). Additionally, their study demonstrated that WSD with an error rate of 40\%–50\% may still enhance IR performance when used with the synset representation, which incorporated synonym information. \citet{gonzalo1999lexical,stokoe2003word} further confirmed the significance of WSD to information retrieval. \citet{gonzalo1999lexical} also found that PoS information has a lower utility for information retrieval. Based on artificially creating word ambiguity,~\cite{sanderson1994word} employed pseudo words to explore the effects of sense ambiguity on information retrieval. They came to the conclusion that the high accuracy of WSD is a crucial condition to accomplish progress. \citet{blloshmi2021ir} introduced an innovative approach to multi-lingual query expansion by integrating WSD, which augments the query with sense definitions as supplementary semantic information in multi-lingual neural ranking-based IR. The results demonstrated the advantages of WSD in improving contextualized queries, resulting in a more accurate document-matching process and retrieving more relevant documents.

\citet{kim2004information} labeled words with 25 root meanings of nouns rather than utilizing fine-grained sense inventories of WordNet. Their retrieval technique preserved the stem-based index and changed the word weight in a document in accordance with the degree to which it matched the query's sense. They credited their coarse-grained, reliable, and adaptable sense tagging system with the improvement on TREC collections. The detrimental effects of disambiguation mistakes are somewhat mitigated by the addition of senses to the conventional stem-based index.

\subsubsection{Machine Translation}\label{sec:Machine Translation}
The challenge of ambiguous word senses poses a significant barrier to the development of an efficient machine translator. As a result, a number of researchers have turned their attention to exploring WSD for machine translation. Some works tried to establish datasets to quantify the WSD capacity of machine translation systems. \cite{rios-gonzales-etal-2017-improving} proposed a test set of 6,700 lexical ambiguities for German-French and 7,200 for German-English. They discovered that WSD remains a difficult challenge for neural machine translation, especially for uncommon word senses, even with 70\% of lexical ambiguities properly resolved. \citet{campolungo2022dibimt} proposed a benchmark dataset that aims at measuring WSD biases in Machine Translation in five language combinations. They also agreed that SOTA systems still exhibited notable constraints when confronted with less common word senses. Incorporating sense labels and lexical chains leads to enhanced performance of Neural Machine Translation (NMT) models, particularly with regard to infrequent word senses. \cite{raganato-etal-2019-mucow} proposed MUCOW, a multi-lingual contrastive test set automatically created from word-aligned parallel corpora and the comprehensive multi-lingual sense inventory of BabelNet. MUCOW spans 16 language pairs and contains more than 200,000 contrastive sentence pairs. The researchers thoroughly evaluated the effectiveness of the ambiguous lexicons and the resulting test suite by utilizing pre-trained NMT models and analyzing all submissions across nine language pairs from the WMT19 news shared translation task.

Some works analyzed the internal representations to understand the disambiguation process in machine translation systems. \cite{marvin-koehn-2018-exploring} examined the extent to which ambiguous word senses could be decoded through the use of word embeddings in relation to deeper layers of the NMT encoder, which were believed to represent words with contextual information. In line with prior research, they discovered that the NMT system frequently mistranslated ambiguous terms. \cite{tang2019encoders} trained a classifier to determine if a translation is accurate given the representation of an ambiguous noun. The fact that encoder hidden states performed much better than word embeddings suggests that encoders are able to appropriately encode important data for disambiguation into hidden states. \citet{liu-etal-2018-handling} discovered that an increase in the number of senses associated with each word results in a decline in the performance of word-level translation. The root of the issue may be the mapping of each word to similar word vectors, regardless of its context. They proposed to integrate techniques from neural WSD systems into an NMT system to address this issue. 

\subsection{Summary}\label{sec:summary_wsd}

WSD as a computational linguistics task most closely related to lexical semantics research, has won extensive discussions among researchers from different fields. Linguists came up with important hypotheses to guide the modeling of word senses. We have observed that some hypotheses have been well grounded in NLP, e.g., learning and representing word meanings with their contexts and word co-occurrences. However, we also observe some important linguistic arguments were rarely studied in the computational linguistic domain, e.g., defining the scope of linguistic units for WSD and integrating relevant concepts (frames) for word sense representations. The development of WSD datasets has greatly ignited the research enthusiasm of scholars in WSD. However, we also observed that the computational research on WSD is also limited by these well-defined datasets because WSD datasets generally follow a very similar labeling paradigm. Relevant linguistic studies have shown broader possibilities in WSD. Finally, we find that many of WSD modeling techniques do not link well with downstream applications. The research of WSD methods has intersections with downstream applications, whereas they cannot well cover the needs of downstream tasks. This also shows that the research opportunities in WSD can be largely extended besides word sense classification.

\subsubsection{Technical Trends}\label{sec:summary_technical_wsd}

\begin{sidewaystable}[!htbp]
\centering
\scriptsize
\begin{tabular}{llllllll}
\toprule
Task & Reference & Tech  & Feature and KB.  & Framework  & Dataset & Score & Metric \\ \midrule
\multirow{8}{*}{Knwl} &~\citet{lesk1986automatic} & Prob. & Statistics, OALD & Count def. overlaps & - & - & - \\
 &~\citet{banerjee2003extended} & ML  & Emb., WN  & Score function & SensEval-2 & 34.60\% & F1  \\
 &~\citet{navigli2007graph} & Graph & Sense graph, WN & Connectivity measures & SemCor & 31.80\% & F1 \\
 &~\citet{basile2014enhanced} & Prob. & Emb., BN &  DSM & SE2013-EN & 71.50\% & F1 \\
 &~\citet{wang2020synset}$_{KB}$ & DL & BERT, WN & Vector represent. & SensEval-2 & 72.70\% & F1 \\
 &~\citet{agirre2009personalizing} & Graph & WN & PageRank & SensEval-2 & 58.60\% & Recall \\
 &~\citet{moro2014entity} & Graph & Sem. graph, BN & PageRank & SE2013-EN & 69.20\% & F1 \\
 &~\citet{scozzafava2020personalized} & Graph & WN, SN  & PageRank & SensEval-2 & 71.60\% & F1  \\ \hline
\multirow{9}{*}{Sup.}  &~\citet{singh2014decision} & ML  & 5-gram, position & Decision Tree  & Manipuri & 71.75\% & Acc \\
&~\cite{o2004class} & ML & Relatedness scores & Decision Tree & SensEval-3 & 65.90\% & F1 \\
 &~\citet{zhong2010makes} & ML & Position, PoS & SVM  & SensEval-2 & 68.20\% & F1  \\
 &~\citet{iacobacci2016embeddings} & ML & Emb., position, PoS & SVM & SensEval-2 & 68.30\% & F1 \\
 &~\citet{popov2017word} & DL  & Emb.  & BiLSTM & SensEval-2 & 70.11\% & Acc \\
 &~\citet{yuan2016semi} & DL & Emb., label propag. & LSTM & SensEval-2 & 74.40\% & F1 \\
 &~\citet{le2018deep} & DL & Emb. & LSTM & SensEval-2 & 72.00\% & F1 \\
 &~\citet{hadiwinoto2019improved} & DL & BERT & Transformer & SensEval-2 & 76.40\% & F1 \\
 &~\cite{bevilacqua2019quasi} & DL & Emb. & BiTransformer & SensEval-2 & 75.70\% & F1 \\
 \hline
\multirow{20}{*}{\begin{tabular}[c]{@{}l@{}}Knwl\\+\\Sup.\end{tabular}} &~\citet{wang2020synset}$_{Sup}$ & DL & BERT, WN & Vector represent. & SensEval-2 & 78.60\% & F1 \\
&~\citet{vial2019sense} & DL & BERT, WN & Transformer & SensEval-2 & 79.70\% & F1 \\
 &~\citet{loureiro2019language} & DL & BERT, WN & Transformer & SensEval-2 & 76.30\% & F1 \\ 
&~\citet{kumar2019zero} & DL & \begin{tabular}[c]{@{}l@{}}Graph emb., \\emb., WN\end{tabular} & \begin{tabular}[c]{@{}l@{}}BiLSTM, Att. \\ConvE\end{tabular} & SensEval-2 & 73.80\% & F1 \\
 &~\citet{bevilacqua2020breaking} & DL & BERT, WN & Trans., Struct. logit & 5 datasets & 80.80\% & F1 \\
 &~\citet{conia2021framing} & DL & BERT, WN & Transformer & SensEval-2 & 78.40\% & F1 \\
 &~\citet{huang2019glossbert} & DL & BERT, WN & \begin{tabular}[c]{@{}l@{}}Transformer, sentence-\\pair classification\end{tabular} & SensEval-2 & 77.70\% & F1 \\
&~\citet{blevins2020moving} & DL & BERT, WN & Trasformer, Score Func. & SensEval-2 & 79.40\% & F1 \\
&~\citet{scarlini2020sensembert} & DL & BERT, BN, Wiki & \begin{tabular}[c]{@{}l@{}}Transformer, Context\\Retrieval\end{tabular} & \begin{tabular}[c]{@{}l@{}}Nouns of\\5 datasets\end{tabular} & 80.40\% & F1\\
&~\citet{scarlini2020more} & DL & BERT, WN, SN & \begin{tabular}[c]{@{}l@{}}Transformer, Context\\Retrieval\end{tabular} & SensEval-2 & 78.00\% & F1 \\
&~\citet{barba2021esc} & DL & BERT, WN & \begin{tabular}[c]{@{}l@{}}Transformer, Extractive\\Sense Learning\end{tabular} & SensEval-2 & 81.70\% & F1 \\
&~\citet{berend2020sparsity} & DL & BERT, WN & \begin{tabular}[c]{@{}l@{}}Transformer, sparse\\coding, PMI\end{tabular} & SensEval-2 & 79.60\% & F1 \\
 \bottomrule
\end{tabular}
\caption{A summary of representative WSD techniques. Knwl denotes knowledge-based methods. Sup. denotes supervised methods. KB denotes knowledge bases. WN denotes WordNet. BN denotes BabelNet. DSM denotes Distributional Semantics Models. Prob. denotes probability. SE2013-EN denotes the SemEval2013 English WSD task. PMI denotes Pointwise Mutual Information.}\label{tab:technical trend_wsd}
\end{sidewaystable}

Table~\ref{tab:technical trend_wsd} shows the technical trends of WSD methods. As seen in the table, earlier approaches likely used knowledge-based and supervised approaches. WordNet and BabelNet are useful knowledge bases that were frequently used by knowledge-based methods. Word embeddings, pre-trained language models, and linguistic features, e.g., PoS tags and semantic relatedness were frequently used by supervised methods. For old pure knowledge-based methods, the PageRank framework was likely used, because many knowledge bases are represented as graphs. PageRank is an algorithm used in graph computation to measure the importance of nodes in a graph. Classical machine learning techniques, e.g., Decision Tree, SVM, LSTM, and Transformers were commonly used by supervised WSD methods. Supervised learning algorithms demonstrate superior performance in comparison to knowledge-based approaches. Nevertheless, it is not always reasonable to assume the availability of substantial training datasets for different areas, languages, and activities. \citet{ng1997getting} predicted that a corpus of around 3.2 million sense-tagged words would be necessary in order to produce a high-accuracy, wide-coverage disambiguation system. The creation of such a training corpus requires an estimated 27 person-years of labor. The accuracy of supervised systems might be greatly improved above the SOTA methods with such a resource. However, the success of this hypothesis is at the cost of huge resource consumption.

We observe more hybrid approaches that leverage knowledge bases in a supervised learning fashion in recent years. This is because researchers have observed the limitations of typical supervised WSD in processing rare or unseen cases. Knowledge bases provide additional information to support the learning of unseen cases. Knowledge bases provide additional knowledge for the languages whose annotated data are scarce. In this case, multi-lingual knowledge bases can enhance the representations of word senses in a new domain. As a result, we can observe the accuracy of the hybrid approaches surpasses the pure knowledge-based or supervised approaches.

Most existing WSD datasets define the task as a word sense classification task. Then, the following methodology research upon the datasets focused on improving the accuracy of mapping the sense of a word to its dictionary sense class. However, should the research on WSD be limited to word sense classification? We have observed that many knowledge-based systems used existing knowledge bases to conduct word sense classification tasks. They have realized the importance of developing an effective knowledge base for WSD. However, it is rare to see that WSD research tries to improve the construction of knowledge bases according to the effectiveness of word sense classification. On the other hand, the meaning of WSD is much larger than detecting the definition of words in a dictionary. Mapping a word to a sense in a dictionary is just an aspect of WSD. Previous works rarely studied what is an appropriate linguistic unit for WSD; what concepts are associated with a word sense in a context. These are very interesting research topics from linguistic and cognitive aspects. However, these topics were not well studied in the computational WSD community.

\subsubsection{Application Trends}\label{sec:summary_application_wsd}

\begin{table}[t]
\centering
\scriptsize
\begin{tabular}{@{}lcccc@{}}
\toprule
Reference & Downstream Task & Feature & Parser  & Explainability \\ \midrule
\citet{farooq2015word}  & Sentiment Computing  & \checkmark & &  \\
\citet{nassirtoussi2015text}  & Sentiment Computing  & \checkmark & & \checkmark  \\
\citet{ohana2009sentiment} & Sentiment Computing  & \checkmark & \checkmark & \\
\citet{saggionalpha2010interpreting}  & Sentiment Computing & \checkmark & \checkmark & \checkmark  \\
\citet{devitt2007sentiment} & Sentiment Computing & \checkmark & & \checkmark  \\
\citet{hung2013using}  & Sentiment Computing  & \checkmark & \checkmark &  \\
\citet{hung2016word}  & Sentiment Computing  & \checkmark & \checkmark & \checkmark \\
\citet{krovetz1992lexical}  & Information Retrieval & & \checkmark &  \\
\citet{gonzalo1998indexing}  & Information Retrieval & & & \checkmark \\
\citet{gonzalo1999lexical}  & Information Retrieval & & & \checkmark  \\
\citet{sanderson1994word}  & Information Retrieval & & & \checkmark  \\
\citet{stokoe2003word} & Information Retrieval & & & \checkmark  \\
\citet{kim2004information} & Information Retrieval & \checkmark & \checkmark & \\
\citet{blloshmi2021ir} & Information Retrieval & \checkmark & & \checkmark\\
\citet{rios-gonzales-etal-2017-improving}  & Machine Translation & \checkmark & &  \\
\citet{raganato-etal-2019-mucow}  & Machine Translation & \checkmark & &  \\
\citet{marvin-koehn-2018-exploring}  & Machine Translation & & \checkmark & \checkmark  \\
\citet{tang2019encoders} & Machine Translation & \checkmark & & \checkmark \\
\citet{liu-etal-2018-handling} & Machine Translation & \checkmark & & \\
 \bottomrule
\end{tabular}
\caption{A summary of the representative applications of WSD in downstream tasks. \checkmark denotes the role of WSD in a downstream task.}
\end{table}

The WSD task was commonly defined as a word sense classification task. However, we observe that classifying words by sense classes is not the only need for downstream NLP tasks. 

There are three main tasks that are strongly related to WSD, e.g., sentiment computing, information retrieval, and machine translation in our survey. One of the roles of WSD on the three tasks is to deliver or enhance features to gain improvements on the three tasks. On the other hand, we also observe many downstream works used WSD techniques as a parser to obtain words with different levels of word sense ambiguity or used WSD to gain insights into their model behaviors to improve the explainability of a study. In these cases, defining WSD as a sense classification task may be sub-optimal for downstream applications. 

WSD has a huge potential in NLP research. For example, disambiguating word senses in a large corpus can lead to a deeper understanding of language usage patterns and the semantic relationships between words. WSD is also a significant component in semantic explainable AI, because it helps researchers better understand the decision-making process of a model on the semantic level. Researchers can develop a more transparent and trustworthy model by explaining word senses in contexts. As a feature generator, a WSD may be more effective if it can generate contextualized word meanings in natural language, rather than predict a sense class that maps to a predefined gloss in a dictionary. However, research in these fields is rare in the WSD community.

Finally, according to~\citet{navigli2009word}, the lack of end-to-end applications that utilize WSD can be attributed to the insufficient accuracy of current WSD systems. This suggests that in the future, more precise WSD systems may be developed, which could potentially enable the use of more semantics-dependent applications.

\subsubsection{Future Works}\label{sec:summary_future_wsd}

As argued before, the task of WSD can be broader than the current word sense classification task setup from either the theoretical research side or the downstream application side. Besides, the improvements in WSD accuracy can also attract more downstream applications. Thus, we come up with the following future work suggestions.

\noindent\textbf{Extending the form of WSD.} WSD can have different learning forms, besides word sense classification, e.g., paraphrasing an ambiguous word into a less ambiguous one~\citep{mao2018word,mao2022metapro}, generating contextualized word senses in natural language. Such an extension may have significance in downstream applications. From the perspective of linguistic and cognitive research, studying how to define a language unit to better disambiguate word senses, or studying how to link a word to its associated concepts in a context can also improve the significance of WSD in the era of LLM-based NLP. Future works may study how to define the task of WSD to better support the research in different disciplines.

\noindent\textbf{Rethinking existing knowledge bases by WSD.} Most of the existing knowledge bases were developed according to human-defined ontologies and word senses. These knowledge bases have been considered as an important resource for many knowledge-based systems. Although the knowledge bases have been used on different tasks, few works analyzed the weakness of the ontologies. Future WSD-related research may try to improve the knowledge bases by rethinking the sense definition, concept node connections, and coverage, rather than simply developing models to enhance the learning ability on a specific task.

\noindent\textbf{Multi-lingual WSD.} Most of the semantic representations are learned from monolingual corpora. As a result, the semantic representations are different between different languages. However, the disambiguation of meanings is not characterized by languages~\citep{boroditsky2011language}. It will significantly improve multi-lingual semantic research if WSD  research can break down language barriers from a cognitive perspective. As argued by frame semantics~\citep{fillmore2006frame}, the meaning of a word is beyond its dictionary definitions. It also associates with the concepts, interconnected with the word. Representing word senses by concepts may achieve a more robust multi-lingual WSD.

\noindent\textbf{Learning WSD as a pre-training task.} Recent years witness great success of PLMs in various domains. The existing PLMs followed the same hypothesis that the sense of a word can be learned from its associated context. However, there has not been a PLM that explicitly disambiguates word senses to enhance the learning of semantic representations. Naively learning the semantic representation of a target word by its associated context words cannot learn the conceptual association of the target word. For example, many words can associate with the word ``apple''. How can we know an apple as fruit is red or green, sweet, tree-growing, nutritious, etc? As an electronic device, Apple is associated with an operating system, a circuit board, a brand, etc. Disambiguating word senses before pre-training may build such connections between concepts.

\noindent\textbf{Fusing WSD with other tasks.}
As~\cite{bevilacqua2021recent} argued, WSD can also be integrated with an entity linking task~\citep{moro2014entity}, where the model predicts associated entities to help WSD systems explore the related glosses and relations. Related fusion works also include fusing WSD for Sentiment Analysis~\citep{farooq2015word}, Information Retrieval~\citep{blloshmi2021ir} and Machine Translation~\citep{campolungo2022dibimt}. The future study of WSD can be grounded on an end task so that the end task can more effectively benefit from the fusion of a WSD model.

\section{Anaphora Resolution}
\label{sect:Anaphora Resolution}

In computational linguistics, Ruslan Mitkov defined anaphora as a \emph{phenomena of pointing back a previously mentioned item in the text}~\citep{mitkov2022oxford}. The pointing back phrase is called an \emph{anaphor} while the previously mentioned item is called an \emph{antecedent}. 

The concept of anaphora should not be confused with co-reference. On the one hand, either anaphora or cataphora (e.g., the phenomena of pointing ahead to a subsequently mentioned item) could be a kind of co-reference. On the other hand, an anaphor and its antecedent are not always co-referential. By definition, the difference between anaphora and co-reference is that anaphora does not require \emph{identify-of-reference} while co-reference requires. In other words, anaphora may describe a relation between expressions that do not have the same referent. For example, in sentence~\ref{ex:ar_sense}, the anaphor ``one'' has the same sense as its antecedent ``a dog'', but they do not refer to the same dog. 
\ex. Jack has \underline{a dog} and Mary also has \textbf{one}. \label{ex:ar_sense}

Building on this, in relation to anaphora, both anaphor and its antecedent are not necessarily referring expressions. For instance, an anaphor can be a verb (henceforth, verb anaphora). In the following example from~\citet{mitkov2014anaphora},
\ex. When Manchester United swooped to lure Ron Atkinson away from the Albion, it was inevitable that his midfield prodigy would \textbf{follow}, and in 1981 he \textbf{did}.

the anaphor ``did'' is a verb, having an antecedent ``follow''. Another example is the \emph{bound anaphora} where the antecedent is a quantified expression~\citep{reinhart1983coreference}:
\ex. \underline{Each manager} exploits the secretary who works for ``him''.

The anaphor ``him'' refers to the quantified expression ``each manager''. Since antecedents in both above two examples are not referring expressions, neither of them is a co-reference. 

Given the definition of anaphora, the task of anaphora resolution is to identify the antecedent of an anaphor. In this survey, we decided to merely focus on anaphora resolution (rather than co-reference resolution) because, on the one hand, most semantic processing tasks only require identifying antecedents. On the other hand, we are not only interested in referring to noun phrases but also other phrases that an anaphor can refer to (e.g., verb phrases and quantified expressions; see the discussion above).

It is worth noting that there have been reviews in the past 20 years about AR/CR from either computer scientists~\citep{Sukthanker2018AnaphoraAC,liu2023brief} or linguists~\citep{mitkov2022oxford,poesio2023computational}. In this survey, our objective is to establish a connection between AR techniques across theoretical research and practical applications.

\subsection{Theoretical Research}\label{sec:theoretical_research_ar}

\subsubsection{Constraints} \label{sec:theoretical_research_ar_constraints}

When human beings resolute co-reference, there are semantic and syntactic constraints. As for the semantic constraints, agreements such as gender and number agreements are the strongest type~\citep{garnham2001mental}. However, most recently, agreement mismatch problems (especially for gender agreements) have been becoming more frequent since more people have started to use plural pronouns to avoid gender bias. 

As for syntactic constraints, according to the binding theory~\citep{buring2005binding}, in the sentence (a) of the following example, ``John'' cannot co-refer with ``him'' while in the sentence (b) ``John'' can.
\ex. \a. John likes him.
     \b. John likes him in the mirror.

\subsubsection{Centering Theory} \label{sec:theoretical_research_ar_center}

Centering Theory~\citep{joshi1979centered,grosz-etal-1983-providing,grosz-etal-1995-centering} was introduced as a model of \emph{local coherence}\footnote{Instead of focusing on the whole discourse, centering theory focuses only on the \emph{discourse segment}.} based on the idea of \emph{center of attention}. The theory assumes that, during the production or comprehension of a discourse, the discourse participant's attention is often centered on a set of entities (a subset of all entities in the discourse) and such an \emph{attentional state} evolves dynamically. It models transitions of the attentional state and defines three types of transitions: \texttt{CONTINUE}, \texttt{RETAIN}, and \texttt{SHIFT}. For each utterance, the transition is decided by its backward-looking center (defined as the most salient entity in the previous utterance that is also realized in the current utterance and denoted as $C_b$) as well as forward-looking center (defined as the most salient entity in the current utterance and denoted as $C_f$).
Consider the following discourse adopted from~\citet{kehler1997current}:
\ex. \a. Terry really gets angry sometimes. \label{ex:ar_center_1a}
\b. Yesterday was a beautiful day and he was excited about trying out his new sailboat. [$C_b=$ Terry, $C_f=$ Terry] \label{ex:ar_center_1b}
\c. He wanted Tony to join him on a sailing expedition, and left him a
message on his answering machine. [$C_b=$ Terry, $C_f=$ Terry] \label{ex:ar_center_1c}
\d. Tony called him at 6AM the next morning. [$C_b=$ Terry, $C_f=$ Tony] \label{ex:ar_center_1d}
\e. Tony was furious with him for being woken up so early. [$C_b=$ Tony, $C_f=$ Tony] \label{ex:ar_center_1e}

\noindent where we annotate each utterance with its backward-looking and forward-looking centers. The transition from utterance~\ref{ex:ar_center_1a} to~\ref{ex:ar_center_1b} is a \texttt{CONTINUE} as both backward-looking and forward-looking centers are unchanged. The next one is a \texttt{RETAIN} transition since although the most salient entity changes (i.e., $C_f$), the forward-looking center stays the same, whereas the transition from utterance~\ref{ex:ar_center_1d} to~\ref{ex:ar_center_1e} is a \texttt{SHIFT} transition because of the change of backward-looking transition. Intuitively, a discourse with more \texttt{CONTINUE} transitions is more coherent than the one with more \texttt{SHIFT} transitions.

Though Centering Theory is not a theory of Anaphora Resolution, Anaphora Resolution can directly benefit from modeling transitions, which provides certain information about the preference for the referents of pronouns (e.g., in a coherent segment, centers co-refer; see~\citet{joshi2006anaphora} for more discussion about the relation between Centering Theory and Anaphora Resolution).

\subsubsection{Discourse Salience} \label{sec:theoretical_research_ar_salience}

A prominent strand of work in psycholinguistics investigates how human beings use anaphora. A referent is more likely to be realized as a pronoun if it is salient in a given discourse~\citep{givon1983topic} (aka. \emph{discourse salience}). Discourse salience is thought to be influenced by various factors, including givenness~\citep{chafe1976givenness, gundel1993cognitive}, grammatical role~\citep{brennan1995centering, Stevenson1994Thematic}, recency~\citep{givon1983topic, arnold1998}, syntactic parallelism~\citep{Chambers1998Structural, arnold1998}, and many other factors. Similar to Centering Theory, most research on discourse salience is about the production of anaphora~\citep{mccoy1999generating, orita2014quantifying, orita2015discourse, chen2018modelling}, but it also provides insights about an antecedent’s relative likelihood for a given anaphor in a given discourse. In this sense, it is plausible to use the aforementioned factors as features to rank candidate antecedents of an anaphor~\citep{lappin1994algorithm, bos2003implementing}.

\subsubsection{Coolness} \label{sec:theoretical_research_ar_cool}

\begin{CJK}{UTF8}{gbsn}

\citet{huang1984distribution} classified human languages into cool languages and hot languages. If a language is ``cooler'' than another language, then understanding a sentence in that language relies more on context (see~\citet{chen2022computational,chen-van-deemter-2022-understanding,chen2023neural} for computational investigations of the theory of Coolness). The evidence that~\citet{huang1984distribution} identified is about the differences between the use of anaphora. Specifically, cool languages (e.g., Mandarin) make liberal use of zero pronouns. Take the following conversation as an example:
\ex. \a. 你今天看见比尔了吗？(Did you see Bill today?)
	 \b. *pro*看见*pro*了。(*I* saw *him*.)

where a *pro* represents a zero pronoun\footnote{In linguistics, a zero pronoun is a pronoun that is implied but not explicitly expressed in a sentence.} (ZP). The first ZP refers to one of the speakers while the second ZP refers to Bill. ZPs of this kind are called Anaphoric ZPs (AZPs). In addition to Mandarin, a number of other languages (i.e., cool languages) also allow ZPs, including examples like Japanese, Arabic, and Korean. The current theory suggests that the anaphora resolution of cool languages should also take AZPs into consideration, namely AZP resolution~\citep{chen-ng-2013-chinese}. 

\end{CJK}

\subsection{Annotation Schemes}\label{sec:annotation_schemes_ar}

In this subsection, we introduce two commonly used annotation schemes for anaphora resolution: MUC and MATE. There are also other schemes, for example, the Lancaster scheme~\citep{fligelstone1992developing} and the DRAMA scheme~\citep{passonneau1997instructions}.

\subsubsection{MUC}

MUC~\citep{hirschman1997automating, hirschman-chinchor-1998-appendix} is one of the very first schemes, which is used for annotating the MUC~\citep{chinchor1995message} and the ACE~\citep{doddington2004automatic} corpora and is still widely used these years. It is primary goal is to annotate co-reference chains in discourse, in which MUC defines and proposes to annotate the IDENTITY (IDENT) relation. Relations as such are symmetrical (i.e., if A IDENT B, then B IDENT A) and transitive (i.e., if A IDENT B and B IDENT C, then A IDENT C). Annotation is done using SGML, for example:
\ex. \label{ex:muc} $\langle$COREF ID=``100''$\rangle$Lawson Mardon Group Ltd.$\langle$/COREF$\rangle$ said $\langle$COREF ID=``101" TYPE=``IDENT" REF=``100"$\rangle$it$\langle$/COREF$\rangle$ ...

The annotation above construct a link between the pronoun ``it'' and the noun phrase ``Lawson Mardon Group Ltd.''. 

MUC proposes to annotate co-reference chains following a paradigm analogous to anaphora resolution. Annotators are first asked to annotate markable phrases (e.g., nouns, noun phrases, and pronouns) and partition the phrases into sets of co-referring elements. This helps the annotation task achieve good inter-annotator agreement (i.e., larger than 95\%).

Nevertheless, it has been pointed out by~\citet{deemter2000coreferring} that MUC has certain flaws: MUC does not guarantee that the annotated relations are all co-referential. It includes either relation that does not follow the principle of identity-of-reference or bound anaphora. Therefore, the resulting corpus would often be a mixture of co-reference and anaphora.

\subsubsection{MATE}

Instead of annotating a single device INDENT, MATE~\citep{poesio1999mate,poesio2004mate} was proposed to do so-called ``anaphoric annotation'' which is explicitly based on the discourse model assumption~\citep{heim1982semantics,gundel1993cognitive,webber2016formal,kamp2013discourse}. The scheme was first proposed to annotate anaphora in dialogues but was then extended to relations in discourse (see~\citet{pradhan2012conll} for more details). Such a good extensibility is a result of the fact that MATE is a \emph{meta-scheme}: It consists of a core scheme and multiple extensions. The core scheme can be used to conduct the same annotation task as MUC and can be extended with respect to different tasks. The annotation normally uses XML, but many of its extensions use other their own formats.

\subsubsection{Zero Pronoun, Bridging Reference, and Deictic Reference}

In addition to the ``co-referential'' relation discussed above, many are also interested in ``hard'' cases, each kind of which is often annotated as following an extension of MATE. These include the following three: (1) zero pronoun:~\citet{pradhan2012conll} annotated (both anaphoric and non-anaphoric) ZPs in Chinese and Arabic (see Section~\ref{sec:theoretical_research_ar_cool}); (2) bridging reference: bridging anaphora is a kind of indirect referent, where the antecedent of an anaphor is not explicitly mentioned but ``associated'' information is mentioned~\citep{clark1975bridging}. Identifying such a relation needs commonsense inference. Consider the following example from~\citet{clark1975bridging}:
\ex. I looked into the room. The ceiling was very high.

``the room'' is an antecedent of ``the ceiling'' because the room has a ceiling; (3) deictic reference: deixis~\citep{webber-1988-discourse} is a phrase that refers to the ``speaker's position'' (e.g., time, place, and situation), which is always abstracted. For example, in
\ex. I went to school yesterday. \label{ex:deixis}

the first person pronoun ``I'' and the word ``yesterday'' are deictic references, which refer to the speaker and the day before the date when~\ref{ex:deixis} was uttered, respectively. Schemes like ARRAU~\citep{poesio2008anaphoric} extended MATE and is able to annotate bridging and deictic references.

\subsection{Datasets} \label{sec:datasets_ar}

\begin{table}[t]
\centering
\scriptsize
\begin{tabular}{llrl} 
\toprule
Dataset & Source & \#Samples & Reference \\ \midrule
MUC & WSJ &	200	&~\citet{chinchor1995message} \\
ACE & News & 1,800 &~\citet{doddington2004automatic} \\
GNOME & Multi-domain & 505 &~\citet{poesio2000annotating} \\
OntoNotes & Multi-domain & 4,560 &~\citet{hovy2006ontonotes} \\
WSC & Manually Written & 285 &~\citet{levesque2012winograd} \\
DPR & Manually Written & 1,880 &~\citet{rahman-ng-2012-resolving} \\
GAP & Wikipedia & 4,454 &~\citet{webster-etal-2018-mind} \\
NP4E & Reuters & 104 &~\citet{hasler2006nps} \\
ECB+ & News & 982 &~\citet{cybulska-vossen-2014-using} \\
ARRAU & Multi-domain & 552 &~\citet{poesio-artstein-2008-anaphoric} \\
\bottomrule
\end{tabular}
\caption{Anaphora Resolution datasets and statistics.}\label{tab:datasets_ar}
\end{table}

As we discussed when we introduced annotation schemes in Section~\ref{sec:annotation_schemes_ar}, there is no clear cut between co-reference and anaphora in computational linguistics research. We hereby review either mainstream corpora utilized in Anaphora Resolution or co-reference resolution, while being mindful of the scope of each of them. The datasets and their statistics are summarized in Table~\ref{tab:datasets_ar}.

The 6th version of MUC~\citep[MUC-6,][]{chinchor1995message} is the first corpus that enables the co-reference resolution, where the task of co-reference resolution and the MUC annotation scheme was first defined. Its texts are inherited from the prevision MUCs and are English news. An example of MUC-6 is shown in Example~\ref{ex:muc}. \citet{chinchor-1998-overview} updated MUC-6 in 2001 and construct the MUC-7/MET-2 corpus. MUC-7 was designed to be multi-lingual (NB: data in Chinese and Japanese are included in MET-2, which has been considered as a part of MUC-7) and to be more carefully annotated than MUC-6 by providing annotators with a clearer task definition and finer annotation guidelines. 

ACE is a multi-lingual (i.e., English, Chinese, and Arabic) multi-domain co-reference resolution corpus~\citep{doddington2004automatic}. In terms of co-reference resolution, it was built with the same purpose as MUC\footnote{Though, in terms of entity recognition, they don't have the same purpose.} and they same problems pointed by~\citet{deemter2000coreferring} (see Section~\ref{sec:annotation_schemes_ar} for more discussion). In addition to MUC and AEC, there are works following the MUC scheme, while targeting domains other than news, which include GENIA~\citep{kim2003genia}, GUM~\citep{Zeldes2017}, and PRECO~\citep{chen2018preco}.

The GNOME corpus was first proposed to investigate the effect of salience on language production (see Section~\ref{sec:theoretical_research_ar_salience} and~\citet{poesio2000annotating,pearson2001effects}) and then be used to develop and evaluate anaphora resolution algorithms~\citep{poesio-2003-associative, poesio2004general} targeting especially the bridging reference resolution, in the course of which the MATE scheme was introduced (see Section~\ref{sec:annotation_schemes_ar}). GNOME is an English multi-domain corpus. The initial GNOME corpus~\citep{poesio1999towards} consists of data from the museum domain (building on the SOLE project~\citep{hitzeman1998use}) and patient information leaflets (building on the ICONOCLAST project), which is then expended to include tutorial dialogues~\citep{poesio2000annotating}. GNOME followed the MATE scheme. Each noun phrase is marked by an $\langle ne \rangle$ and its anaphoric relations (marked by) are annotated separately, for example:\\

\begin{mdframed}
\noindent\texttt{\scriptsize{{$\langle$ne ID="ne07" ... $\rangle$ \\
Scottish-born, Canadian-based jeweller, Alison Bailey-Smith$\langle$/ne$\rangle$\\
...\\
$\langle$ne ID="ne08"$\rangle$ $\langle$ne ID="ne09"$\rangle$Her$\langle$/ne$\rangle$ materials$\langle$/ne$\rangle$\\
\\
$\langle$ante current="ne09"$\rangle$\\
$\langle$anchor ID="ne07" rel="ident" ... $\rangle$ \\
$\langle$/ante$\rangle$
}}}
\end{mdframed}

OntoNotes~\citep{hovy2006ontonotes} is a multi-lingual (i.e., English, Chinese, and Arabic) multi-domain dataset. It is one of the most commonly used anaphora/co-reference resolution and was used in the CoNLL 2012 shared task~\citep{pradhan2012conll}. It was annotated following an adapted version of the MATE (named M/O scheme by~\citet{poesio2023computational}). Though it has been widely used in co-reference resolution tasks, many of its relations are not co-reference. For example, bound anaphora frequently appear (see the start of this section for more discussion). Additionally, OntoNotes annotates ZPs in its Chinese and Arabic portions (see Section~\ref{sec:theoretical_research_ar_cool}). There are other corpora following M/O, but targeting different domains, including the biomedical (e.g., CRAFT~\citep{cohen2017coreference}), Wikipedia (e.g., GAP~\citep{webster-etal-2018-mind} and WikiCoref~\citep{ghaddar2016wikicoref}), and literary text (e.g., LitBank~\citep{bamman-etal-2020-annotated}); and different anaphorical phenomena, including bridging anaphora (e.g., ISNOTE~\citep{hou-etal-2018-unrestricted}), style variation (e.g., WikiCoref~\citep{ghaddar2016wikicoref}), and ambiguity (e.g., GAP~\citep{webster-etal-2018-mind}).

ARRAU is an English multi-domain (i.e., dialogue, narrative, and news) anaphora resolution dataset, annotated following the MATE scheme~\citep{poesio-artstein-2008-anaphoric,uryupina2020annotating}. However, different from other corpora that also follow MATE, ARRAU extended MATE to annotate anaphoric ambiguity explicitly (recall that MATE is a meta-scheme). \citet{poesio-artstein-2008-anaphoric} introduced the \emph{Quasi-identity} relation, which is used for the situation when co-refer is possible but not certain by annotators and allowed each anaphor to have two distinct interpretations. In the example sample below, the footnote ``1,2'' of the anaphor ``it'' means ambiguity exists and it can either refer to `engine E2' or ``the boxcar at Elmira''. \\

\begin{mdframed}
\noindent\texttt{\scriptsize{{(u1) M: can we .. kindly hook up ... uh ... [engine E2]$_1$ to [the boxcar at Elmira]$_2$ \\
(u2) M: +and+ send [it]$_{1,2}$ to Corning as soon as possible please
}}}
\end{mdframed}

The Winograd Scheme Challenge~\citep[WSC,][]{levesque2012winograd} focuses on the ``hard'' cases of CR, which often require lexical and commonsense knowledge. It can be traced back to Terry Winograd's minimal pair~\citep{winograd1972understanding}:
\ex. \a. \textbf{The city council} refused the demonstrators a permit because \textbf{they} feared violence. \label{ex:winograd_a}
\b. The city council refused \textbf{the demonstrators} a permit because \textbf{they} advocated violence. \label{ex:winograd_b}

The antecedent of ``they'' changes from ``the city council'' to ``the demonstrators'' from~\ref{ex:winograd_a} to~\ref{ex:winograd_b}. \citet{levesque2012winograd} introduced the WSC benchmark consisting of hundreds of such minimal pairs. Since then, many larger-scale WSC-like corpora have been constructed. This includes the DPR corpus~\citep{rahman-ng-2012-resolving}, the PDP corpus~\citep{davis2017first}, and the Winogrande corpus~\citep{sakaguchi2021winogrande}. Following a similar paradigm, GAP~\citep{webster-etal-2018-mind}, Winogender~\citep{rudinger-etal-2018-gender} and Winobias~\citep{zhao-etal-2018-gender} were proposed for ``hard'' cases that link to gender bias. 

NP4E~\citep{hasler2006nps} and ECB+~\citep{cybulska-vossen-2014-using} are corpora for investigating cross-document co-reference. They annotated both entities and events co-reference and both within and cross-document co-reference. These corpora were built by starting from a set of clusters of documents, the documents of each of which describe the same fundamental events. 

The corpora mentioned above are all in English, some of which have Chinese and Arabic portions. There are anaphora/co-reference resolution corpora that focus on languages other than them. These include ANCOR~\citep[in French,][]{muzerelle-etal-2013-ancor}, ANCORA~\citep[in Catalan and Spanish][]{taule2008ancora}, COREA~\citep[in Dutch][]{hendrickx-etal-2008-coreference}, NAIST~\citep[in Japanese][]{iida-etal-2007-annotating}, PCC~\citep[in Polish][]{ogrodniczuk2013polish}, PCEDT~\citep[in Czech][]{nedoluzhko2014annotation}, and TUBA-DZ~\citep[in German][]{telljohann2004tuba}.

\subsection{Knowledge Bases}\label{sec:knowledgebases_ar}

Both lexical and world knowledge are useful for anaphor interpretation. See the following examples from~\citet{martin_2015}:
\ex. \a. There was a lot of \textbf{Tour de France riders} staying at our hotel. Several of \textbf{the athletes} even ate in the hotel restaurant.
\b. She was staying at \textbf{the Ritz}, but even that \textbf{hotel} didn’t offer dog walking service.

We need the lexical knowledge that indicates ``riders'' are ``athletes'' while need the world knowledge of the fact that ``Ritz'' is a ``hotel''. 

\begin{table}[t]
\centering
\scriptsize
\begin{tabular}{llrl} 
\toprule
Name & Knowledge & \#Entities & Structure  \\ \midrule
WordNet & Lexical & 155,327 &  Graph \\
COW & Lexical & 157,112 & Graph \\
ODW & Lexical & 92,295 & Graph \\
AWN & Lexical & $\approx$10,000 & Graph \\
Wikipedia & World & 13,489,694 & Unstructured \\
Wikidata & World & 100,905,254 & Graph \\
DBpedia & World & $\approx$4,580,000 & Graph \\
Freebase & World & $\approx$2.4 B & Graph \\
YAGO & World & 4,595,906 & Graph \\
WikiNet & World & 3,347,712 & Graph \\
OMCS & World & 62,730 & Graph \\
Medical-KG & World & 22,234 & Graph \\
\bottomrule
\end{tabular}
\caption{Useful knowledge bases for anaphora resolution.}
\end{table}

\noindent \textbf{WordNet} provides lexical knowledge of English~\citep{miller1998wordnet}, including lexical entries (e.g., meaning, part-of-speech, etc.) and relations (e.g., synonyms, hyponyms, and meronyms, etc.) among them.

\noindent \textbf{Wikipedia} has been an important world knowledge source for many anaphora/co-reference resolution systems. These knowledge bases consist of documents from Wikipedia as well as related meta-data. Typical examples include bases from those directly dumped from raw Wikipedia documents\footnote{\url{https://dumps.wikimedia.org/}} to better-structured ones, such as Wikidata~\citep{vrandevcic2014wikidata}, DBpedia~\citep{auer2007dbpedia}, and Freebase~\citep{bollacker2008freebase}.

\noindent \textbf{Knowledge Graphs} have become popular in anaphora/co-reference resolution tasks because bases that build on raw Wikipedia are needed to be further processed (e.g., entity and relation extraction) before use. Popular knowledge graphs include those that build on Wikipedia (e.g., YAGO~\citep{suchanek2008yago} and WikiNet~\citep{nastase-etal-2010-wikinet}), that are about Commonsense (e.g., OMCS~\citep{singh2002open}), and that are about expert knowledge (e.g., Medical-KG~\citep{uzuner2012evaluating}).

\noindent \textbf{Search Engines}, e.g., Bing and Google were also used by a few works (e.g.,~\citep{emami-etal-2018-generalized}) to ``hunt'' knowledge for the target entities in order to resolve hard anaphora like those in WSC (see Section~\ref{sec:datasets_ar}), in addition to the above knowledge bases in the strict sense.

\subsection{Evaluation Metrics} \label{sec:evaluation_metrics_ar}

\noindent \textbf{Vanilla Precision, Recall and F1.} A plausible way to assess anaphora resolution systems is by viewing both mention detection and mention linking tasks as simple classification tasks and measuring the performance using vanilla precision, recall, and F1 scores. A good evaluation metric needs to be both interpretable and discriminative. However, unfortunately, these measures cannot meet any of these criteria~\citep{moosavi-strube-2016-coreference}, especially for the mention linking task as they overlook the structure of these relations (most of which are chain-structured).

\noindent \textbf{MUC and Beyond.} Along with MUC-6 (see Section~\ref{sec:datasets_ar}),~\citet{vilain1995model} proposed the MUC score. It computes the recall and precision of anaphora/co-reference resolution outputs by considering co-reference chains in a document as a graph. \citet{vilain1995model} first defined two sets: a set of key entities $\mathcal{K}$, in which there are gold standard reference chains (NB: a chain is sometimes named as a class or a cluster), and a set of response entities $\mathcal{R}$, in which there are system generated chained. MUC score computes the recall based on the number of missing links in $\mathcal{R}$ compared to $\mathcal{K}$, formally:
\begin{equation}
    \mbox{Recall} = \frac{\sum_{k_i \in \mathcal{K}}\left( \lvert k_i \rvert - \lvert p(k_i, \mathcal{R}) \rvert \right)}{\sum_{k_i \in \mathcal{K}}\left( \lvert k_i \rvert - 1 \right)}
\end{equation}
where $\lvert k_i \rvert$ is the number of mentions in the chain $k_i$ and $p(k_i, \mathcal{R})$ is the set of partitions that is constructed by intersecting $k_i$ with $\mathcal{R}$. The computation of MUC precision is done by switching $\mathcal{K}$ and $\mathcal{R}$. However, it has been pointed out that MUC has certain flaws: on the one hand, since MUC is merely building on mismatches of links between the two sets, it is not discriminative enough~\citep{bagga1998algorithms,luo2005coreference}. For example, it does not tell the difference between an extra link between two singletons or two prominent entities. On the other hand,~\citet{luo2005coreference, kubler-zhekova-2011-singletons} argued that MUC prefers singletons. For instance, if we merge all mentions in OntoNotes into singletons, the resulting MUC will be higher than that of the SOTA~\citep{moosavi-strube-2016-coreference}. 

Many metrics beyond MUC have been proposed by measuring recall and precision using mentions instead of links. \citet{bagga1998algorithms} proposed $B^3$, which considers the fractions of the correctly identified mentions in $\mathcal{R}$:
\begin{equation}
    \mbox{Recall} = \frac{\sum_{k_i \in \mathcal{K}} \sum_{r_j \in \mathcal{R}} \frac{\lvert k_i \cap r_j \rvert^2}{\lvert k_i \rvert}}{\sum_{k_i \in \mathcal{K}} \lvert k_i \rvert}
\end{equation}
The precision is also computed by switching $\mathcal{K}$ and $\mathcal{R}$. As pointed by~\citet{luo2005coreference} and~\citet{luo2016evaluation}, $B^3$ still cannot fully properly handle singletons and, additionally, repeated mentions. To solve this,~\citet{luo2005coreference} proposed CEAF to incorporate measures of similarities between entities:
\begin{equation}
    \mbox{Recall} = \frac{\sum_{k_i \in \mathcal{K^*}} \phi(k_i, g(k_i))}{\sum_{k_i \in \mathcal{K}} \phi(k_i, k_i)}
\end{equation}
where $\mathcal{K}^*$ is the set of key entities that have the optimal mapping with $\mathcal{R}$, which is found by the Kuhn-Munkres algorithm, and $\phi(\cdot)$ is a similarity measure. Nevertheless, CEAF has two shortcomings: it overlooks all unaligned response entities~\citep{denis2009global} and weights entities equally~\citep{stoyanov2009conundrums}. 

In addition to above mentioned based metrics, to handle singletons,~\citet{recasens2011blanc} proposed BLANC to also consider non-coreference/non-anaphoric links. It measures the fiction of both correctly identified co-reference links and non-coreference entities, and averages them to obtain the final score. 

\citet{moosavi-strube-2016-coreference} conducted controlled experiments and proved that all the aforementioned computations of precision and recall are neither interpretable nor reliable as they suffer from the so-called \emph{mention identification effect}. They proposed the LEA metric, which was claimed to be able to solve the above issues from two perspectives: (1) it considers both links and mentions; (2) it weights entities with respect to their importance.

\subsection{Annotation Tools}\label{sec:annotation_tools_ar}

\noindent \textbf{Text Editors}. In the early years, anaphora/co-reference were annotated using text editors or manipulation tools. For example, MUC-6 and ACE were annotated using plain text editors while GNOME was annotated using the XML manipulation tool developed by the University of Edinburgh\footnote{http://www.ltg.ed.ac.uk/software/}.

\noindent \textbf{Co-reference Annotation Tools}. Later, linguists and computer scientists developed software that enables multi-layer annotation. The software that is designed for annotating co-reference or allows the annotations of relations between phrases can be used for anaphora/co-reference annotation tasks. For example, ARRAU and PCC used MMAX2, which is a free, extensible, general-purpose, and desktop-based annotation tool. It allows users to annotate relations using fields in a form, and the form is customizable. The NP4E project used PALinkA and ECB+ used CAT~\citep{bartalesi-lenzi-etal-2012-cat}. Both of them were designed for the event and reference annotation. More recently, co-reference annotation tools that provide better visualization, allow drag-and-drop annotation, and offer post-annotation analysis have been built. Typical examples include CorefAnnotator~\citep{Reiter2018ag}, which is open-sourced and desktop-based, SCAR~\citep{OBERLE18.178}, which is open-sourced and web-based, and LightTag, which is not fully free but provides good online teamwork services.

\noindent \textbf{Annotation Tools with Advanced Functionalities}. Some annotation tools provide extra services that help to make sure the annotation procedure is fast and reliable. We classify these services into three categories: (1) External Knowledge: BRAT~\citep{stenetorp2012brat} and INCEpTION~\citep{klie-etal-2018-inception} integrate external knowledge bases, e.g., Freebase and Wikidata (see Section~\ref{sec:knowledgebases_ar}). Once an annotator identifies an entity, these tools would search the linked base and return related entry; (2) Pre-trained Models: Tools such as TagEditor, Togtag, INCEpTION, and MyMiner~\citep{salgado2012myminer} can call embedded pre-trained entity recognition models so that they can suggest positions of possible name entities during annotation, in which MyMiner was designed specifically for the medical domain (see~\citet{neves2021extensive} for an overview of annotation tools for medical NLP). Additionally, beyond name entities, TagEditor and INCEpTION can also suggest potential reference chains based on their integrated pre-trained co-reference resolvers, enabling active learning for anaphora/co-reference resolution; (3) Cross-document Annotation: using CROMER~\citep{girardi2014cromer} and CoRefi~\citep{bornstein-etal-2020-corefi}, annotators can tag, link, or update entities across multiple documents. This is done by allowing annotators to cluster documents based on topics and annotate documents in a cluster together.

\subsection{Methods}\label{sec:methods_ar}

\subsubsection{Rule-based Methods}\label{sec:task1_ar}

\noindent \textbf{A. Linguistically-inspired Approaches} 

Like many other tasks in NLP, early works on anaphora resolution built on rules that are rooted cognitively and linguistically. Here, the term ``early'' represents the age when systematic evaluations of anaphora resolution, e.g., MUC, had not been introduced. The very first algorithm is the naive algorithm proposed by~\citet{hobbs1978resolving}. It first does a breadth-first search from the parse tree of the sentence to search for identifying mentions and links mentions based on constraints introduced in Section~\ref{sec:theoretical_research_ar_constraints}.

Later on, a series of anaphora resolution systems were proposed together with computational investigations of the effect of salience (see Section~\ref{sec:theoretical_research_ar_salience}). Based on a set of factors that proved to influence salience,~\citet{sidner1979towards} introduced rules that are used to compute the expected focus of discourse and rules that are used to interpret anaphora. As a matter of fact, this work was built on the ``centering view'' rooted from~\citet{grosz1977representation}, which suggests that, during anaphora resolution, the searching of antecedents should be restricted to the set of centered entities. It could be seen as a prototype of the idea of ``center of salience'' of the centering theory (see Section~\ref{sec:theoretical_research_ar_center}), but the rules proposed by~\citet{sidner1979towards} are extremely complex.

Starting from~\citet{sidner1979towards},~\citet{carter1987interpreting} focused on the rules about salience and developed a system coined Shallow Processing Anaphor Resolver (SPAR). SPAR maintains linguistically-inspired rules as domain knowledge and does commonsense inference over them. As pointed out by~\citet{carter1987interpreting}, since maintaining domain knowledge and reasoning rules is expensive, SPAR made them as simple as possible. That is why it was called ``shallow processing''. Carter assessed SPAR on a set of 322 test samples and found that SPAR could successfully resolve 93\% pronominal anaphors and 87\% non-pronominal anaphora. \citet{hobbs-etal-1988-interpretation} formalized commonsense inference in anaphora resolution as abduction and introduced TACITUS. To do abduction, in TACITUS, knowledge (i.e., rules) is maintained in formal logic (first-order predicate logic in this case). Focusing on salience,~\citet{lappin-leass-1994-algorithm} proposed the Resolution of Anaphora Procedure (RAP) algorithm. After selecting a set of candidate antecedents based on semantic and syntactic constraints, RAP contains a rule-based procedure for assigning values to several salience parameters, which are then used for resolute anaphors. An assessment on 360 hand-crafted texts containing pronouns showed RAP defeated the naive algorithm by 2\%.

Also starting from~\citet{sidner1979towards}, there were subsequent works that extended the idea of ``focus'' on the basis of the introduction of the concept of ``centering''. \citet{brennan-etal-1987-centering} introduced the BFP algorithm for anaphora resolution, which roughly has three stages: (1) construct a set of candidate antecedents with accordance to the rules of the semantic constraint; (2) filter and classify the candidates based on which action a candidate belongs to in centering theory (see Section~\ref{sec:theoretical_research_ar_center}); and (3) select the best candidate in according to a pre-defined preference over the actions. One limitation of the BFP algorithm is that its final choice is merely based on a linear preference order. To optimize this selection process,~\citet{beaver2004optimization} marries BFP with the optimality theory. Another limitation is that, by only considering the center theory, BFP overlooked a key pattern of how human resolute pronouns, namely, incremental resolution~\citep{kehler1997current}. In response to this problem,~\citet{tetreault-2001-corpus} proposed the Left-to-Right Centering (LRC) algorithm, which is an incremental resolution algorithm that adheres to centering constraints. An evaluation on the New York Time corpus~\citep{ge1998statistical} suggests that LRC outperformed both BFP and the naive algorithm.

\noindent \textbf{B. Knowledge-poor Approaches} 

After the introduction of the MUC-6 shared task, anaphora resolution systems are able to be evaluated on a large scale. However, the trade-off is that the anaphora resolution systems can no longer access inputs that are annotated with gold-standard semantic and syntactic knowledge. Building on this setting, ``knowledge-poor'' approaches were proposed and most systems of this kind prefer rules that have high precision but do not rely on knowledge. The most influential work is CogNIAC~\citep{baldwin-1997-cogniac}, which is a heuristic precision-first anaphora resolver that relies on rules that are almost always true. For example, CogNIAC contains a rule saying \emph{if there is just one possible antecedent in entire the prior discourse, then that entity is the antecedent}. 
Its rules were selected based on the precision tested on a set of test sentences.
It is worth noting that rules in CogNIAC are still used in many SOTA practical anaphora resolution systems (e.g., the Stanford Deterministic Coreference Resolver~\citep{lee-etal-2013-deterministic}).

\noindent \textbf{C. Approaches with Approximate Knowledge}

As pointed out by~\citet{poesio2023computational}, this encourages two major changes in anaphora resolution: one this that instead of relying on perfect knowledge and doing reasoning on it, anaphora resolution systems started to syntactic parsers and approximate knowledge like WordNet. The other is that the focus of anaphora resolution models moved from being aware of only pronouns to all kinds of nominal phrases (that function as referring). 

\citet{kameyama-1997-recognizing} proposed to resolve anaphors that are proper names, descriptions, and pronouns. It relies on syntactic and semantic constraints, but the related information came from a syntactic parser and morphological filter based on person, number, and gender features. Later on, approaches that marry rules with WordNet were introduced~\citep{harabagiu-maiorano-1999-knowledge,liang-wu-2003-automatic}. They made use of heuristic rules (as in CogNIAC), some of which consider lexical information from WordNet.

The most famous rule-based anaphora resolution system is the one proposed by~\citet{haghighi-klein-2009-simple}, which is still frequently used as a strong baseline in today's research on anaphora resolution. In addition to aforesaid syntactic and semantic constraints,~\citet{haghighi-klein-2009-simple} makes full use of the parse trees. For example, it contains rules that rely on the distance between mentions, which is obtained from computing the shortest path between two mentions in the parse tree. It also uses Wikipedia as a resource for acquiring semantic knowledge of each entity.

One limitation of heuristic-based systems is that lower precision features often overwhelm higher precision features. In response to this, more recently rule-based systems~\citep{raghunathan-etal-2010-multi,lee-etal-2013-deterministic} categorized rules into sieves and made decisions with an ordered set of rules. These works are often called multi-sieve approaches.

\subsubsection{Statistical-based Methods}\label{sec:task2_ar}

The introduction of large-scale benchmarks also encourages the trend of using machine learning techniques in anaphora resolution. Basically, these learning-based models treat anaphora resolution as a series of classification problems. We categorize them on the basis of how they define the classification task.

\noindent \textbf{A. Mention-pair Models} 

Mention-pair models train a classifier to determine whether two mentions co-refer or not. It was first introduced by~\citet{aone1995automated} and then perfected by~\citet{soon-etal-2001-machine}. To build a mention-pair model, there are five steps:
\begin{enumerate}
    \item Identifying Mentions: As a practical anaphora resolution model, the first step of this framework is to identify mentions. \citet{soon-etal-2001-machine} break down the mention identification into two stages: they first used three statistical sequence taggers (which is a Hidden Markov Model~\citep{church1989stochastic}) to do part-of-speech tagging, noun phrase identification, and name entity recognition, respectively. The outputs of them are noun phrases as well as name entities. Then, they designed rules to recognize nested noun phrases based on the identified noun phrases. For each discourse, the resulting set of mentions is the union of noun phrases, name entities, and nested noun phrases. In later works, this module was replaced by more advanced sequence taggers, e.g., conditional random field. See~\citep{lata2022mention} for a survey.
    \item Feature Engineering: Akin to many statistical models, feature engineering is always needed. \citet{soon-etal-2001-machine} made use of not only syntactic and semantic features as usual but also lexical features with the help of WordNet. In addition to~\citet{soon-etal-2001-machine}, many works used knowledge bases for feature engineering (e.g.,~\citet{vieira-poesio-2000-empirically,ponzetto-strube-2006-exploiting}).
    In 2008,~\citet{bengtson-roth-2008-understanding} found that a simple model with good feature engineering can defect the SOTA model at that moment. 
    \item Generating Training Examples: They used a heuristic-based method to generate training pairs (i.e., a pair of positive and negative examples). More specifically, a positive instance consists of an anaphor $A_1$ and its closest preceding antecedent $A_2$ while a negative instance consists of the same anaphor $A_1$ and the mention that intervenes $A_1$ and $A_2$. There has been a number of modifications to this strategy. For example,~\citet{ng-cardie-2002-improving} forced that $A_1$ can only be a non-pronominal once $A_2$ is also a non-pronominal. \citet{harabagiu-etal-2001-text,ng-cardie-2002-combining,strube-etal-2002-influence,yang-etal-2003-coreference} further enhanced this process by applying rule-based or learning-based filters. 
    \item Building a Classifier: In this step, statistical machine learning techniques have been used. These include decision trees~\citep{soon-etal-2001-machine,mccarthy1995using}, random forests~\citep{lee2017scaffolding}, Max Entropy classifier~\citep{berger-etal-1996-maximum,ge1998statistical}, and memory-based learning~\citep{daelemans2004timbl}. 
    \item Generating Co-reference Chains: The last step is to partition these anaphora into co-reference chains. Normally, clustering techniques are used in this step. These include closest-first clustering~\citep{soon-etal-2001-machine}, best-first clustering~\citep{ng-cardie-2002-improving}, correlational clustering~\citep{mccallum2004conditional}, and graph partitioning algorithms~\citep{mccallum2003object,nicolae-nicolae-2006-bestcut}.
\end{enumerate}

\noindent \textbf{B. Entity-Mention Models} 

As a matter of fact, the task mention-pair anaphora resolution is counter-intuitive from the perspective of linguists and cognitive scientists. Additionally,~\citet{poesio2023computational} pointed out that mention-pair models also overlook features of entities~\citep{ng-2010-supervised}. In response to this, entity-mention models were proposed. They directly link mentions to entities by clustering. Specifically,~\citet{cardie-wagstaff-1999-noun} trained a model to classify whether a mention belongs to a partially constructed cluster. However, according to the evaluation by~\citet{luo2005coreference}, the performance of the models of this kind is not comparable to mention-pair models. 

\noindent \textbf{C. Mention-Ranking Models} 

Another problem of mention-pair models is that they only do binary classification without comparing different potential antecedents. To remedy this,~\citet{denis-baldridge-2008-specialized} proposed an entity-ranking model, replacing the binary classification loss with a ranking loss. \citet{rahman2011narrowing} combined entity-ranking strategy with the entity-mention model, yielding SOTA performance at that moment.

\subsubsection{Neural Anaphora Resolution}\label{sec:task3_ar}

\noindent \textbf{A. Conventional Deep Learning Models} 

\citet{wiseman-etal-2015-learning} was the first to use deep neural networks in anaphora resolution. It is a non-linear mention-ranking model. Instead of conjunction features (as in statistical models), the model of~\citeauthor{wiseman-etal-2015-learning} uses a neural network to learn feature representations as an extension to the mention-ranking model. They defined two feature vectors, each of which is obtained from pre-training the model on any of the sub-tasks of anaphora resolution, namely, mention identification and mention linking. The final decision is made through a non-linear classification, based on these features. Both~\citet{wiseman-etal-2016-learning} and~\citet{clark-manning-2016-improving} augmented the work of~\citet{wiseman-etal-2015-learning} by inducing global features, but they followed different schemes. \citet{wiseman-etal-2016-learning} ran a recurrent neural network (RNN) to encode the representation of each sequence of mentions corresponding to an entity (i.e., a cluster) in the history. Whereas,~\citet{clark-manning-2016-improving} first used a feed-forward neural network to encode each mention-pair of an entity and computed the entity representation by pooling over all mention-pairs. Later on,~\citet{clark-manning-2016-deep} extended their previous work~\citep{clark-manning-2015-entity}, which built up co-reference chains with agglomerative clustering. Each mention starts in its own cluster and then pairs of clusters are merged using imitation learning (a type of reinforcement learning technique) by assuming merging clusters are actions. \citet{clark-manning-2016-deep} replaced imitation learning with deep reinforcement learning. \citet{liu2023multitask} proposed a multi-task learning framework for mention detection and mention linking tasks, because they found that the learning of mention detection task can enhance the learning of dependent information of input tokens, which is complimentary for mention linking detection. Such an approach achieved comparable performance to~\citet{kocijan-etal-2019-wikicrem} with only 0.05\% WIKICREM training samples.

\noindent \textbf{B. End-to-End Models} 

A significant benefit of employing deep learning models lies in their capacity to operate without the requirement of handcrafted features, thus enabling the creation of end-to-end (End2End) systems. \cite{lee-etal-2017-end} proposed the first End2End anaphora resolution system. It needs no human-craft feature or parser and, more importantly, it learns to process mention identification and linking tasks jointly. To this end, the fundamental idea is to first view all spans in the previous discourse as candidate antecedents and do mention ranking (NB: it was called span ranking in~\citet{lee-etal-2017-end} as the spans it sent for rank are not always mentions). The inputs pass through an RNN and each span is represented by the concatenation of the RNN hidden states of the first token and the last token as well as the weighted sum of all tokens in the span using the attention mechanism~\citep{bahdanau2014neural}. The final decision of each pair is made using a feed-forward neural network. One limitation of this method is that since it searches over all possible spans, the search space would be extremely large. To remedy this, candidate spans are pruned by limiting the maximum span width, the number of spans per word, the maximum number of antecedents, and the length of input documents. This End2End model was tested on the OntoNotes dataset and outperformed all previous works.

Akin to mention-pair anaphora resolution systems, End2End anaphora resolution is problematic because it ranks every span-anaphor pair separately. In response to this problem,~\citet{lee-etal-2018-higher} introduced a higher-order coarse-to-fine inference strategy for End2End anaphora resolution models (henceforth, C2F-AR), which, in short, does cluster ranking. It infers in an iterative manner. The antecedent distributions are used to update the span representations before doing inference, enabling later decisions conditioned on previous decisions. C2F-AR uses a coarse factor that can further prune candidate span during this higher-order inference,

More recent works focused on either improving span representations or selecting candidate spans. For example,~\citet{luo-glass-2018-learning} used a two-layer bi-directional RNN and combined the representations of adjacent sentences in order to improve span representation with cross-sentence dependency information. \citet{zhang-etal-2018-neural-coreference} proposed to enrich the span representations by training a mention identification model jointly assigning each candidate span an antecedent score. For each pair of spans,~\citet{kirstain-etal-2021-coreference} replaced span representations with a combination of lightweight bilinear functions between pairs of endpoint token representations. \citet{wu-etal-2020-corefqa} formalized the End2End anaphora resolution as a question-answering task. A query is produced for each entity and predicts the positions of all spans in the co-reference chain. 

\noindent \textbf{C. Knowledge-based Models}

Analog to classical rule-based and statistical-based approaches, works on neural anaphora resolution models also seek to integrate knowledge. In terms of the use of open knowledge bases,~\citet{aralikatte-etal-2019-rewarding} used world knowledge to compute rewards for reinforcement learning-based anaphora resolution models. More specifically, they submitted the predictions to an OpenIE system and compared the predicted anaphora with the knowledge to compute the reward. \citet{zhang-etal-2019-knowledge} extracted knowledge triples related to each entity from knowledge graphs and used them to enrich span representations using a knowledge attention module. 

It has been pointed out that pre-trained language models are knowledge bases~\citep{petroni-etal-2019-language}. Many recent anaphora resolution models have incorporated pre-trained language models, including BERT~\citep{devlin2018bert}, SpanBERT~\citep{joshi-etal-2020-spanbert}, and CorefBERT~\citep{ye-etal-2020-coreferential}.

There has been a line of work focusing on addressing mention linking in WSC-like corpora (see Section~\ref{sec:datasets_ar}). As aforementioned, resolving these ``hard'' cases needs reasoning with world knowledge. Works of this line incorporate either external knowledge bases~\citep{emami-etal-2018-generalized} or pre-trained language models~\citep{kocijan-etal-2019-wikicrem,attree-2019-gendered}.

\subsubsection{Anaphoric Zero Pronoun Resolution}\label{sec:task4_ar}

As mentioned in Section~\ref{sec:theoretical_research_ar_center}, ``cool'' languages (e.g., Chinese, Japanese, Korean, and Arabic) contain anaphoric zero pronouns (AZPs), and many works have focused on resolving AZPs. As with other anaphora resolution tasks, early works on AZP resolution (AZPR) used rule-based approaches and statistical approaches. Theoretically, these works are built on the fact that speakers process zero pronouns (ZPs) in the same way as pronouns~\citep{yang1999comprehension}. Early on, most of the works are for Japanese because of the NAIST corpus~\citep{iida-etal-2007-annotating}, in which AZPs are annotated. \citet{kameyama1985zero, okumura-tamura-1996-zero} used center theory-based approaches for AZPR in Japanese. Statistical-based approaches were proposed with a focus on exploring useful features, including syntactic pattern features~\citep{iida2007zero}, heuristic rules~\citep{isozaki-hirao-2003-japanese}, and features that had been considered in anaphora resolution systems~\citep{nakaiwa1995extrasentential,nakaiwa-shirai-1996-anaphora,seki2001probabilistic,seki2002probabilistic,sasano-etal-2008-fully,sasano-kurohashi-2011-discriminative}. Meanwhile, there were also a number of Korean AZPR systems building on the Korean portion of Penn Treebank~\citep{byron2006resolving,han2006korean}.

Later on, the development of systems for Chinese~\citep{zhao-ng-2007-identification,kong-zhou-2010-tree,chen-ng-2013-chinese,chen-ng-2014-chinese,chen-ng-2015-chinese-zero} and Arabic AZPs became active after the introduction of OntoNotes~\citep{aloraini-poesio-2020-cross}. 

From~\citet{chen-ng-2016-chinese}, AZPR systems also went into the age of deep learning. Most of the works were for Chinese AZPR, including approaches that use deep feedforward neural networks~\citep{chen-ng-2016-chinese}, RNNs~\citep{yin2017deep,yin2019chinese}, attention network~\citep{yin-etal-2018-zero}, memory network~\citep{yin-etal-2017-chinese}, deep reinforcement learning~\citep{yin-etal-2018-deep} and BERT~\citep{song-etal-2020-zpr2}.

The training of AZPR systems shares the problem of lacking annotated training data. For example, the AZPR largest corpus, i.e., the Chinese portion of OntoNotes, contains only 12,111 AZPs. To incorporate more data into training, there have been three paradigms: (1) Joint modeling:~\citet{chen-etal-2021-tackling} and~\citet{aloraini2022joint} proposed to train a model that resolves either AZPs and non-zero pronouns jointly; (2) Multi-linguality:~\citet{iida2011cross} and~\citet{aloraini-poesio-2020-cross} trained multi-lingual AZPR systems which were trained on AZPR data in multiple languages; (3) Data augmentation:~\citet{liu-etal-2017-generating} made use of large-scale reading comprehension dataset in Chinese to generate pseudo training data for Chinese AZPR. \citet{aloraini2021data} augmented Arabic AZPR data by a number of augmentation strategies, e.g., back translation, masking candidate mentions, etc.

\subsection{Downstream Applications}\label{sec:downstream_applications_ar}

\subsubsection{Machine Translation}\label{sec:application1_ar}

\citet{stojanovski-fraser-2018-coreference} provided the following example to illustrate how oracle anaphora singles can help machine translation systems. 
\ex. \a. Let me summarize the novel for you.
     \b. It presents a problem.
     \c. er!@\#\$XPRONOUN It presents a problem.
     \d. Er prasentiert ein Problem.

Given the context (a) and the course sentence (b), based on the oracle anaphora information,~\citet{stojanovski-fraser-2018-coreference} pre-pend the input sentence of machine translation with pronoun translation as shown in (c) and ask the system to translation with a target (d) in German. In this case, the pronoun ``it'' which refers to ``novel'' (in German ``Roman'') is translated to ``er'' (the German masculine pronoun agreeing with ``Roman''). Without this information, they argued that machine translation will be hard to produce ``er''. The experiment on a number of Neural machine translation models suggested that would improve the BLEU scores by 4-5 points. This argumentation was strengthened by the experiments conducted by~\citet{saunders-etal-2020-neural}, who concluded that NMT does not translate gender co-reference.
Despite these theoretical studies, many works~\citep{le-nagard-koehn-2010-aiding,hardmeier2010modelling,guillou-2012-improving} focused on improving machine translation with anaphora resolution outputs. The solution is often using anaphora resolution outcomes to obtain features of each pronoun (including, gender, number, and animacy) in order to enhance the pronoun translation performance. Beyond these works,~\citet{miculicich-werlen-popescu-belis-2017-using} proposed to use clustering scores which are used for generating co-reference chains in anaphora resolution (see Section~\ref{sec:task2_ar}) as features for re-ranking machine translation results. 

There has been a long tradition of studying the impact of AZPs on machine translation systems, especially when translating from a pro-drop language to a non-pro-drop language. For example, the Japanese-English machine translation in the 1990s had already been deployed an AZPR systems~\citep{nakaiwa1992zero}. Later systems followed a slightly different strategy. Instead of doing a full anaphora resolution, these systems only detect AZPs in the source language and directly translate them into the target language without further resolute them~\citep{tan2019detecting,wang-etal-2019-one}.

\subsubsection{Summarization}\label{sec:application2_ar}

There are two major uses of anaphora resolution in text summarization~\citep{steinberger2007two}. One is to help with finding the important terms while the other is to help with evaluating the coherence of the summarization. Many works have demonstrated that incorporating the information of co-reference chains contributes to both the faithfulness and the coverage of summarization systems~\citep{bergler2003using,witte2003fuzzy,sonawane2016role,liu-etal-2021-coreference}. Nevertheless, it is also worth noting that there are also some studies that showed that anaphora resolution had negative effects~\citep{orasan2007influence,mitkov2007anaphora}. One possible explanation is that the effect highly depends on the task the summarization system is addressing and the performance of the anaphora resolution systems (NB: these studies have been 15 years old).

\subsubsection{Textual Entailment}\label{sec:application3_ar}

For textual entailment, to understand the impact of anaphora resolution,~\citet{mirkin-etal-2010-assessing} manually analyzed 120 samples in the RTE-5 development set~\citep{bentivogli2009fifth}. They found that for 44\% samples anaphora relations are mandatory for inference and for 28\% sample anaphora optionally support the inference. Based on this fact, many systems that got involved in the RTE challenge made use of anaphora resolution. Nevertheless, since anaphora resolution systems at that moment were not strong enough, errors they made would propagate to downstream textual entailment systems~\citep{adams-etal-2007-textual,agichtein2008combining}. As a consequence, the contribution of anaphora resolution was negative or not significant~\citep{bar2008efficient,chambers-etal-2007-learning}.

\subsubsection{Sentiment Computing}\label{sec:application4_ar}

For sentiment computing,~\citet{Sukthanker2018AnaphoraAC} listed two situations when anaphora resolution can help. One is when doing sentiment analysis on online reviews, a characteristic of them is that online reviews often focus on a particular entity and, therefore, the mentions often in less elaborated forms (e.g., pronouns). Resolution of these mentions can chain them into a global entity and, hence, improve the sentiment analysis performance. The other is that anaphora resolution can also be used in fine-grained aspect-based sentiment analysis. Anaphora resolution plays a pivotal role in this task by facilitating the clustering of entities into distinct aspects. This, in turn, aids in the extraction of sentiments and opinions associated with each aspect.

The contribution of anaphora resolution in sentiment computing tasks can be summarized as follows: it enables discourse-level sentiment analysis by linking mentions from different sentences. Many efforts have been carried out to demonstrate such an ability for anaphora resolution. \citet{Nicolov2008SentimentA} conducted systematic experiments to understand the impacts of anaphora resolution on sentiment analysis. Specifically, they tried to incorporate anaphora information into a number of sentiment analysis models and assessed them on varieties of datasets. They concluded that, on average, anaphora resolution can boost sentiment analysis performance by 10\%. Based on this finding, sentiment analysis systems that are assembled with anaphora resolution have been proposed~\citep{Jakob2010UsingAR,Ding2010ResolvingOA,Le2016SentimentAU}.

\subsection{Summary}\label{sec:summary_ar}

Anaphora resolution has been explored extensively by theoretical linguists, psycholinguists as well as computational linguistics. It is the manifest of structural semantics because the meaning of an anaphor elucidates the syntactic relationship between the anaphor and its antecedent. Early anaphora resolution models were inspired by theories and findings in linguistics, such as the theory of syntactic and semantic constraints from theoretical linguistics and the findings about factors that influence the choice of referential form from psycholinguists. Later on, by marrying these theories with computational models, linguists also gained insights regarding the comprehension and production of anaphora from anaphora resolution systems. For instance, we could understand better how each salience factor contributes to the use of anaphora through the importance analysis of a computational model that considers the factor. Most recently, though most computational works focus on building End2End anaphora resolution systems based on deep learning techniques, linguistic theories about anaphora are still proven to play vital roles~\citep{chai-strube-2022-incorporating}. Dataset is core for either practical or theoretical anaphora resolution research. Though many annotation schemes and datasets have been introduced, we found that they share two limitations: one is that due to the fact that anaphora is a complex concept, annotations of anaphora resolution datasets are always imperfect~\citep{deemter2000coreferring}. The other is the lack of wide-coverage datasets that covers all kinds of anaphora. Finally, we found that anaphora resolution is useful in many downstream tasks, including major tasks of both natural language understanding and natural language generation. It is always utilized as a producer of additional features for downstream tasks. Different from other tasks in this survey, we rarely see how anaphora resolution techniques help boost the explainability of downstream models, apart from the work of~\citet{saunders-etal-2020-neural}. We also have not observed that anaphora resolution techniques are used for constructing datasets for downstream tasks.

\subsubsection{Technical Trends}\label{sec:summary_technical_ar}

\begin{sidewaystable}[!htbp]
\centering
\scriptsize
\begin{tabular}{lllllll}
\toprule
Task & Reference & Feature & Framework & Dataset & Score & Metric \\ 
\midrule
\multirow{8}{*}{Rule-based} &~\citet{carter1987interpreting} & Salience & Logic rules & self-collected dataset & 93.00\% & Acc \\
 &~\citet{lappin-leass-1994-algorithm} & Salience & Logic rules & self-collected dataset & 85.00\% & Acc \\
 &~\citet{brennan-etal-1987-centering} & Semantic constraints & Centering theory & New York Times & 59.40\% & Acc \\
 &~\citet{tetreault-2001-corpus} & Semantic constraints & Centering theory & New York Times & 80.40\% & Acc \\
 &~\citet{baldwin-1997-cogniac} & \begin{tabular}[c]{@{}l@{}}Syntactic, Semantic,\\Discourse\end{tabular} & Logic rules & self-collected dataset & 77.90\% & Acc\\
 &~\citet{liang-wu-2003-automatic} & WordNet & Logic rules & Brown Corpus & 77.00\% & Acc \\
 &~\citet{haghighi-klein-2009-simple} & Syntactic, Semantic & Logic rules & ACE & 79.60\% & MUC-F \\ \hline
 \multirow{7}{*}{Stat.-based} &~\citet{soon-etal-2001-machine} & \begin{tabular}[c]{@{}l@{}}Syntactic, Semantic,\\WordNet\end{tabular} & Mention-pair & MUC-6 & 62.60\% & MUC-F \\
 &~\citet{cardie-wagstaff-1999-noun} & \begin{tabular}[c]{@{}l@{}}Lexical, Syntactic,\\Semantic\end{tabular} & Entity-Mention & MUC-6 & 64.90\% & MUC-F \\
 &~\citet{denis-baldridge-2008-specialized} & Linguistic \& Positional & Mention-ranking & ACE & 67.00\% & CEAF-F \\
 &~\citet{rahman2011narrowing} & \begin{tabular}[c]{@{}l@{}}Lexical, Syntactic,\\Semantic\end{tabular} & Mention-ranking & ACE & 60.80\% & CEAF-F \\ \hline
 \multirow{13}{*}{DL-based} &~\citet{wiseman-etal-2015-learning} & Syntactic, Semantic & Mention-rank., DNN & OntoNotes & 82.86\% & Acc \\
 &~\citet{wiseman-etal-2016-learning} & \begin{tabular}[c]{@{}l@{}}Syntactic, Semantic,\\Global Feature\end{tabular} & Mention-rank., RNN & OntoNotes & 64.21\% & CoNLL-F \\
 &~\citet{clark-manning-2016-deep} & Syntactic, Semantic & DRL & OntoNotes & 65.73\% & CoNLL-F \\
 &~\citet{clark-manning-2016-improving} & \begin{tabular}[c]{@{}l@{}}Syntactic, Semantic,\\Global Feature\end{tabular} & Mention-ranking, DNN & OntoNotes & 65.52\% & CoNLL-F \\
 &~\citet{lee-etal-2017-end} & Word \& Cha. Emb. & End2End, LSTM, DNN & OntoNotes & 68.80\% & CoNLL-F \\
 &~\citet{lee-etal-2018-higher} & ELMo & End2End, LSTM, DNN & OntoNotes & 73.00\% & CoNLL-F \\
 &~\citet{zhang-etal-2018-neural-coreference} & Glove \& Cha. Emb. & BiLSTM, Joint Learning & OntoNotes & 69.20\% & CoNLL-F \\
 &~\citet{joshi-etal-2019-bert} & BERT &~\citet{lee-etal-2018-higher} & OntoNotes & 76.90\% & CoNLL-F \\
 &~\citet{joshi-etal-2020-spanbert} & SpanBERT &~\citet{lee-etal-2018-higher} & OntoNotes & 79.60\% & CoNLL-F \\
 &~\citet{wu-etal-2020-corefqa} & SpanBERT & QA & OntoNotes & 83.10\% & CoNLL-F \\
 &~\citet{kocijan-etal-2019-wikicrem} & BERT\_WikiCREM & DNN & DPR & 84.80\% & Acc \\ 
 &~\citet{liu2023multitask} & BERT & Transformer, MTL & DPR & 84.58\% & Acc \\ \hline
 \multirow{6}{*}{AZPR} &~\citet{okumura-tamura-1996-zero} & Salience & Center Theory & self-collected dataset & 78.30\% & Acc \\
&~\citet{sasano-etal-2008-fully} & Salience & Probalistic & self-collected dataset & 39.10\% & F1 \\
&~\citet{chen-ng-2016-chinese} & Syntactic, Lexical & DNN & OntoNotes & 52.20\% & F1 \\
&~\citet{yin2017deep} & Word2Vec, Global & RNN & OntoNotes & 53.60\% & F1 \\
&~\citet{yin-etal-2018-deep} & Word Embedding & DRL & OntoNotes & 57.20\% & F1 \\
&~\citet{song-etal-2020-zpr2} & BERT & DNN, MTL & OntoNotes & 58.49\% & F1 \\
\bottomrule
\end{tabular}
\caption{A summary of representative anaphora resolution techniques. Note that~\citet{rahman2011narrowing} reported that the performance of~\citet{denis-baldridge-2008-specialized} was 57.7\% CEAF-F and that CoNLL-F is the average of MUC, B3, and CEAF scores. Stat. denotes statistics. DL denotes deep learning. AZPR denotes Anaphoric Zero Pronoun Resolution. Cha. Emb. denotes character embedding. ACE denotes automatic content extraction. MTL denotes multi-task learning. DRL denotes deep reinforcement learning}\label{tab:ar_technical_summary}
\end{sidewaystable}

As seen in Table~\ref{tab:ar_technical_summary}, there are two clear technical trends. One is that the research interest in the realm of anaphora resolution has shifted from machine learning-based or rule-based anaphora resolution to neural approaches, especially the End2End neural anaphora resolution, which does mention identification and linking simultaneously. Another one is that, as previously elucidated in Section~\ref{sec:methods_ar}, there exist distinct shortcomings associated with each of the task formulations such as mention pair, entity mention, and mention ranking. Consequently, a recent tendency is to employ higher-order inferences~\citep{lee-etal-2018-higher} to directly rank clusters or entities, which allows for the incorporation of benefits from all the formulations. To sum up, the SOTA anaphora resolution models are often \emph{End2End cluster ranking models}.

Most recent advances tended to further improve this paradigm from two angles, namely reducing the search space as an End2End anaphora resolution searches across all possible spans in its inputs for antecedents~\citep{wu-etal-2020-corefqa}; and equipping anaphora resolution systems with knowledge (which, recently, often large-scale pre-trained language models) to boost their ability of reasoning~\citep{joshi-etal-2019-bert,joshi-etal-2020-spanbert}. Furthermore, recent investigations on anaphora resolution have also led to advancements in various deep learning paradigms. Deep reinforcement learning and multi-task learning were employed for obviating the need for language-orientated hyperparameter tuning~\citep{clark-manning-2016-deep}, investigating the enduring impact of pronoun-candidate antecedent pairs~\citet{yin-etal-2018-deep}, and enhancing the dependency learning of mention pairs~\citep{liu2023multitask}.

Meanwhile, there were also certain efforts that concentrated on resolving ``hard'' cases and multi-linguality in anaphora resolution. As for the former one, people were aware of the models' capacity to resolve ambiguous pronouns and biases (especially, gender bias) learned by anaphora resolution models~\citep{levesque2012winograd,rudinger-etal-2018-gender}. The SOTA models of this line of work are often assembled with knowledge bases~\citep{emami-etal-2018-generalized} or pre-trained language models~\citep{kocijan-etal-2019-wikicrem}. As for the latter one, multi-lingual anaphora resolution systems were developed in order to either, theoretically, unify the theory of reference for different languages~\citep{nedoluzhko2022corefud}, or, practically, enrich the datasets for low-resource anaphora resolution languages or tasks (e.g., AZPR;~\citet{aloraini-poesio-2020-cross}).

In addition to these two trends for developing practical anaphora resolution systems, there is also a long tradition of studying how human beings understand and use anaphors with the algorithms introduced in this section from the age of rule-based methods~\citep{sidner1979towards, carter1987interpreting} to the most recent deep learning based methods~\citep{chai-strube-2022-incorporating,same-etal-2022-non}.

\subsubsection{Application Trends}\label{sec:summary_application_ar}

\begin{table}[!htbp]
\scriptsize
\centering
\begin{tabular}{llcc} 
\toprule
Reference & Downstream Task & Feature & Explain.  \\ \midrule
\citet{le-nagard-koehn-2010-aiding} & Machine Translation & \checkmark&  \\ 
\citet{hardmeier2010modelling} & Machine Translation & \checkmark & \\ 
\citet{miculicich-werlen-popescu-belis-2017-using} & Machine Translation & \checkmark & \\ 
\citet{saunders-etal-2020-neural} & Machine Translation & \checkmark & \checkmark  \\ 
\citet{steinberger2007two} & Summarization Evaluation & \checkmark & \\ 
\citet{bergler2003using} & Summarization & \checkmark & \\ 
\citet{liu-etal-2021-coreference} & Summarization  &\checkmark & \\ 
\citet{agichtein2008combining} & Textual Entailment & \checkmark & \\ 
\citet{Jakob2010UsingAR} & Sentiment Computing & \checkmark & \\ 
\citet{Ding2010ResolvingOA} & Sentiment Computing & \checkmark &\\ 
\bottomrule
\end{tabular}
\caption{A summary of the representative applications of anaphora resolution in downstream tasks. \checkmark denotes the role of anaphora resolution in a downstream task.}\label{tab:ar_downstream_application}
\end{table}

Many demonstrations were carried out approximately 15 years ago to validate the necessity of anaphora resolution for both language generation and understanding downstream tasks~\citep{steinberger2007two,mirkin-etal-2010-assessing,Nicolov2008SentimentA,li2021knowledge,he2022knowledge}. Nevertheless, practically, at that moment, anaphora resolution often had negative effects~\citep{bar2008efficient,chambers-etal-2007-learning,orasan2007influence,mitkov2007anaphora}. This is mainly because anaphora resolution systems were not powerful enough and errors they made may propagate to their downstream tasks. 

Recently, with significant advancements in the capabilities of anaphora resolution systems, more and more anaphora resolution systems have been used for providing anaphora information for downstream tasks (see Table~\ref{tab:ar_downstream_application}). In short, anaphora resolution helps its downstream applications mainly in two ways. It links noun phrases in different sentences. As a consequence, these applications have better performance in comprehending discourse-level information. On the other hand, linking noun phrases helps downstream applications to do higher-level reasoning, e.g., extracting global entities~\citep{Sukthanker2018AnaphoraAC} and recovering the ellipses~\citep{aralikatte-etal-2021-ellipsis}.

Most downstream task models utilize anaphora resolution as an additional feature to improve task performance. However, we did not see how anaphora resolution techniques help to explain how and why anaphora is used in a certain context. 

\subsubsection{Future Works}\label{sec:summary_future_ar}

\noindent \textbf{Developing robust annotation schemes.} Current annotation schemes for anaphora practically work fine but are theoretically problematic as there is no unified rule of what is remarkable, and no clear cut between co-reference and anaphora (though there is a clear boundary between them in linguistic theory). Annotation schemes so far are imperfect to improve the practicality so that large anaphora/co-reference resolution datasets (that can be used for training and assessing data-driven anaphora resolution systems) could be constructed. In exchange, the resulting corpora are imperfect in terms of both quality (i.e., some annotated relations might not be anaphoras) and coverage (i.e., some kinds of anaphora are not covered). On a different note, anaphora resolution, which can also be seen as a pragmatics task, disagreement on how an anaphora is interpreted happens across different readers~\citep{uma2022scaling}. Nonetheless, many datasets resolve disagreements through majority voting, while only a few works explicitly annotated ambiguities, which are the causes of the disagreements (e.g.,~\citet{poesio-artstein-2008-anaphoric}). In aggregate, it is plausible to design a scheme (probably by extending MATE) that not only handles disagreements but also balances quality, practicality, and coverage. Furthermore, it is important to empirically investigate how the errors and limitations inherent in the annotation scheme can impact the performance of anaphora resolution systems.

\noindent \textbf{Anaphora resolution evaluation.} Analogue to the disagreements in the anaphora annotation, one can expect that, for a single mismatch between an output and a reference answer, it might be an error for some readers but not an error for the rest. For different mismatches, they might have different severity. The impact of severity of errors has been studied for the production of reference (see~\citet{van-miltenburg-etal-2020-gradations}; e.g., saying ``a woman is a man'' is more serious than saying ``a red coat is pink''), but it has never been explored in the realm of anaphora resolution. This said, roughly computing the overlaps between model outputs and reference outputs might be problematic. On the one hand, due to discrepancies and varying degrees of errors in anaphora resolution, human evaluation~\citep{martschat-strube-2014-recall} is necessary to improve the analysis and evaluation of anaphora resolution models, as well as to establish benchmarks for developing more accurate evaluation metrics. On the other hand, when designing new evaluation metrics, disagreements, and error severity should be considered by data-driven methods.

\noindent \textbf{Model development.} Regarding future advancements in anaphora resolution models, a significant area of focus should be on computational studies of anaphora resolution tasks that are firmly grounded in theory but have yet to be extensively explored. Examples of such tasks include but are not restricted to (1) bridging, deictic, and plural references, which are crucial aspects of referential language, yet their computational treatment has been limited, possibly due to a shortage of relevant annotated datasets; and (2) disagreement resolution, which involves learning from discrepancies in human interpretations of anaphoric expressions to better capture the pragmatic nuances of such references, and should be incorporated into future models~\citep{uma2021learning}; and (3) cross-document anaphora resolution, which is critical for downstream applications such as knowledge graph construction and cross-document information extraction, yet has received insufficient attention in terms of data, methods, and evaluation metrics, particularly in relation to event resolution.

\section{Named Entity Recognition}
\label{sect:Named Entity Recognition}

Name Entity Recognition (NER) is a critical component of Information Extraction, which involves identifying entity mentions in text, defining their boundaries, and assigning them entity types. The most commonly recognized entity types by NER systems are Location, Person, and Organization, and tokens referring to these entities are classified as entity mentions. In the following example:
\ex. Steve Jobs is the founder of Apple.

an NER system would recognize the entities that ``Steve Jobs'' is Person; ``Apple'' is Organization. NER systems use pre-defined entity types, which may vary across different implementations. For example, Stanford's widely used NER software~\citep{finkel2005incorporating} provides three versions that recognize three classes (Location, Person, Organization), four classes (Location, Person, Organization, Misc), and seven classes (Location, Person, Organization, Money, Percent, Date, Time), respectively. NER is a critical component in the field of NLP~\citep{NJJ1,gao2022graph,he2021construction} and is often combined with other tasks, such as Relation Extraction (RE), to serve as a foundation for various NLP applications. Besides, NER is also used in various data mining tasks to recognize keywords, topics, and attributes~\citep{he2019understanding,li2021knowledge, li2019implementation}.

NER can be traced back to the third Message Understanding Conference (MUC-3)~\citep{chinchor1993evaluating}. The task for MUC-3 was designed to extract relevant information from the text and convert it into a structured format based on a predefined template, e.g., incident, the targets, perpetrators, date, location, and effects. Early NER systems that participated in MUC-3 primarily relied on rule-based approaches, which involved the manual creation of rules to identify named entities based on their linguistic and contextual features. However, with the dominance of deep learning in the NLP community, most NER tasks are now performed using neural networks. One of the first neural networks for NER was proposed by~\cite{collobert2008unified}, which used a single convolutional neural network with manually constructed feature vectors. Later, this approach was replaced with high-dimensional continuous vectors, which were learned from large amounts of unlabeled data in an unsupervised manner~\citep{collobert2011natural}. With stronger models, now, the research in NER has been largely extended to nested NER~\citep{su2022global}, few-shot NER~\citep{huang-etal-2022-copner}, joint entity and relation extraction (JERE)~\citep{zhong-chen-2021-frustratingly, mao2022uncertainty}.

Compared to standard NER whose entity relationship is absent, entities in nested NER have a hierarchical or nested structure, where one entity is embedded within another entity. For example, given 
\ex. The Ontario Supreme Court said ...

``Ontario'' is a state entity that is embedded under the government entity of ``Ontario Supreme Court''~\citep{ringland2019nne}. Given the very expensive annotation costs, few-shot NER is also a very important research trend. It learns NER with a limited amount of labeled data. JERE tasks are established based on the needs of downstream applications. In many cases, people not only need to know what an entity is but also need to know the relationship between entities. Thus, JERE needs to identify named entities in text as well as extract the relationships that exist between them. In the following example
\ex. Greg Christie has been one of the greatest engineers at Apple.

For standard NER, ``Greg Christie'' should be identified as Person; ``Apple'' should be identified as Company. However, for JERE, besides the above entity recognition, an additional relationship label, ``work\_at'' should also be predicted. Compared to identifying entities that are hierarchically structured within each other in nested NER tasks, the outcomes of JERE deliver another relationship dimension to connect entities. Both tasks are helpful in developing a comprehensive knowledge graph.

Due to the wide range of applications of NER, there have been several surveys conducted on this typical NLP task~\citep{li2020survey, yadav2018survey}. One recent study~\citep{song2021deep} focused specifically on NER in the biomedical field, also known as Bio-NER. In this domain, the presence of meaningless characters in biomedical data presents a significant challenge, particularly with regards to inconsistent word distribution. Similarly, \citet{liu2022chinese} summarized and discussed the challenges specific to Chinese NER, rather than the more general English NER tasks. Meanwhile, \citet{nasar2021named} explored both NER and RE tasks, as they are closely linked and are typically composed of pipeline tasks. The aforementioned surveys focus on the technical perspective of NER, based on deep learning technology, while this section broadens the horizon of NER from theoretical foundations to applications.

\subsection{Theoretical Research}\label{sec:theoretical_research_ner}

\subsubsection{Prototype Theory}\label{sect:Prototype Theory}
\citet{rosch1973natural} argued that our classification system, which includes the classification of named entities, is based on a central or prototype example. A prototype is a typical example of a category that represents the most common features or characteristics associated with the category. For example, the prototype of ``bird'' must associate the features, such as wings, feathers, and the ability to fly. Birds such as ostriches or penguins, which do not perfectly possess these characteristics, may be viewed as less typical examples. \citet{rosch1975family} discovered that individuals can identify typical category examples faster and with greater precision than atypical examples. Thus, learning from prototypes can help to quickly grasp the important features of a named entity with a few examples.

\subsubsection{Graded Membership}~\label{sect:Graded Membership}
\citet{rosch1976basic} argued that the classification of categories is frequently determined not by strict boundaries, but by various degrees of membership. We can use this theory for NER because the NER task also categorizes entities by predefined classes. The idea of Graded Membership implies how humans perceive and categorize the world around us. Some categories, e.g., ``vegetable'',  may be viewed as less distinct and vaguer. The theory suggests that the borders between categories may not be well-defined in some cases, leading to ambiguities when attempting to classify certain items, such as tomatoes or mushrooms. The ambiguity can be further compounded by cultural or regional differences in how categories are defined or classified. 

\subsubsection{Conceptual Blending}~\label{sect:Conceptual Blending}
According to~\citet{fauconnier2008way}, the act of blending different elements and their corresponding relationships is an unconscious process that is believed to be ubiquitous in everyday thought and language. This process involves the combination of various mental spaces or cognitive domains that are drawn from different scenarios and experiences. These scenarios may be derived from personal experiences, cultural practices, or societal norms, among others. Concept blending allows us to create a new concept by combining existing ones in novel ways. For example ``SpaceX'' may be mapped to mental spaces related to ``aerospace'' and ``technology''; ``Tesla'' may be mapped to mental spaces related to ``car'' and ``clean energy''. Conceptual blending provides an explanation for the recognition and comprehension of newly named entities by mapping them onto existing mental spaces or concepts.

\subsubsection{Grammatical Category}\label{sect:Grammatical Category}
From the aspect of computational linguistics, the core issue of NER is how to define a named entity.~\cite{marrero2013named} group the criteria of a named entity as grammatical category, rigid designation, unique identification, and the domain of applications. However, many of the entity definitions in the NER domain are imperfect. From the view of grammatical category, a named entity is traditionally defined as a proper noun or a common name for a proper noun. Previous work has described NER as the recognition of proper nouns in general. However, as pointed out by~\citet{borrega2007we}, the classic grammatical approach to proper noun analysis is insufficient to deal with the challenges posed by NER applications. For instance, in a toy question-answering task such as 
\ex. Do crocodiles live in the sea or on land?

``crocodiles", ``sea", and ``land" are not proper nouns, while they are commonly recognized as the essential entities for a proper understanding of the question. Consequently, a proper noun is no longer considered a criterion for identifying named entities in current NER research.

\subsubsection{Rigid Designation}
The rigid designation is a concept in the philosophy of language which suggests that certain names or labels are inherently linked to the things they represent, e.g., ``Barack Obama'' rigidly designates the person who is the 44th President of the US, and it cannot be used to refer to any other person or entity. NER can be viewed as a form of rigid designation as it assigns labels to entities based on their intrinsic identity~\citep{kripke1972naming}, rather than on their usage in the text. However,~\citet{laporte2006rigid} noted that not all expressions that appear to designate rigidly can be analyzed as directly referring to an object in every possible world. This highlights the difficulty of defining entities with complex concepts in real-world applications. As a result, annotators likely make subjective judgments when labeling complex entities, which may be affected by entity descriptions and annotators' understanding.

\subsubsection{Unique Identification}
From the view of unique identification, the MUC conferences require that NER tasks annotate the ``unique identification" of entities for all expressions~\citep{grishman1996message}. However, determining what is unique depends on contextual elements, and can be a subjective process. While this ``unique identification" is typically considered to be the reference being referred to, the definition itself poses a challenge in terms of defining what is truly unique.

\subsubsection{Domain of Applications}
The definition of named entities was frequently grounded in the domain of applications. Entity definitions can be different between different NER tasks. For instance, in drug-drug interaction tasks~\citep{DDI_501}, diseases may not be considered entities, whereas they are entities in adverse drug events~\citep{Demner-Fushman2019}. Inconsistent entity definitions create challenges for machine learning. Because inconsistent entity definitions mean that for the same semantic unit, the machine has to summarize different entity representations to distinguish their labels under different tasks. This is also not conducive to training an all-around NER classifier on different application domains.

\subsection{Annotation Schemes}\label{sec:annotation_schemes_ner}

\begin{table}[t]
\centering
\scriptsize
\begin{tabular}{lcccccccccc}
\toprule
Tokens:  & West & African & Crocodile & are & semiaquatic & reptiles & that & live & in & Africa\\ \midrule
IO  & I  & I & I & O & I & I  & O  & O  & O  & I\\
BIO & B  & I & I & O & B & I  & O  & O  & O  & B\\
BIOES & B  & I & E & O & B & E  & O  & O  & O  & S\\
\bottomrule
\end{tabular}
\caption{The three common annotation schemes for NER.}
\label{tag_scheme}
\end{table}

NER is typically approached as a sequence labeling task, where each token in a sentence is assigned a label. Three common annotation schemes are shown in Table~\ref{tag_scheme}. The IO scheme is a classification task that distinguishes between two classes, namely ``Inner'' and ``Other'', to determine whether a token belongs to an entity or not. On the other hand, the BIO scheme employs three labels, namely ``Beginning'', ``Inner'', and ``Other'', to identify tokens that represent the start of an entity, tokens that belong to an entity, and tokens that do not belong to any entity. The BIOES scheme expands on the BIO scheme by incorporating two additional labels, namely ``Single'' and ``End'', to more precisely define the boundaries of entities.

By employing the IO scheme, the binary classification of tokens is simplified, as each token is labeled as either belonging to an entity or not. This straightforward labeling system makes it easier to identify entities in a text, but it fails to specify the position of the entities within the text. In contrast, the BIO scheme provides more precise annotations by identifying the beginning and continuation of an entity in the text. This labeling system allows for more accurate recognition of entities in a text and better classification of individual tokens. The BIOES scheme further extends the BIO scheme by providing more precise boundaries for entities, thereby allowing for better recognition of entity boundaries in a text. The ``Single'' label is used to denote an entity that consists of a single token, whereas the ``End'' label is used to indicate the final token of an entity. By incorporating these additional labels, the BIOES scheme provides a more nuanced approach to entity recognition and annotation.

\subsection{Datasets}\label{sec:datasets_ner}

\begin{table}[!tbh]
\centering
\scriptsize
\begin{tabular}{@{}llll@{}}
\toprule
Dataset & Source & \# Sample & Reference \\ \midrule
MUC-6 & Newswire & 318 articles &~\cite{grishman1996message} \\
ACE-05 & Social media & 12,548 sentences &~\cite{walker2006ace} \\
TACRED & Newswire & 106,264 instances &~\cite{zhang2017tacred} \\
CoNLL-2003 & Reuters\footnote{\url{www.reuters.com/researchandstandards/}} & 1,499 articles &~\cite{sang2003introduction} \\
I2B2 & ECI Corpus\footnote{\url{http://www.ldc.upenn.edu/}} & 1,600 patient records &~\cite{stubbs2015annotating} \\
ADE & MEDLINE \footnote{ http://www.nlm.nih.gov/bsd/indexing/training/PUB\_050.htm } & 2,972 document &~\cite{Gurulingappa2012a} \\
DDI & DrugBank\footnote{\url{https://go.drugbank.com/}} & 1,025 document &~\cite{Herrero-Zazo2013} \\
WNUT-17 & Social media & 2,295 documents &~\cite{derczynski2017results} \\
OntoNote 5.0 & Social media & - &~\cite{weischedel2013ontonotes} \\
CPR & MEDLINE & - &~\cite{Krallinger2017} \\ 
MultiNERD & Wikipedia & 10 languages &~\cite{tedeschi-navigli-2022-multinerd} \\ 
HIPE-2020 & Newspapers & 17,553 mentions &~\cite{ehrmann_extended_2022} \\ 
NNE & Newswire & 49,208 sentences &~\cite{ringland2019nne} \\ 
GENIA & MEDLINE & 18,546 sentences &~\cite{kim2003genia} \\ 
\bottomrule
\end{tabular}
\caption{NER datasets and statistics.}\label{tab:ner dataset statistics}
\end{table}

The surveyed popular NER datasets and their statistics can be viewed in Table~\ref{tab:ner dataset statistics}. The first NER-focused dataset was published in the 6th MUC Conference~\citep{grishman1996message}. This task consists of three sub-tasks, including entity names, temporal expressions, and number expressions. The defined entities include organizations, persons, and locations; The defined time expressions include dates and times; The defined quantities include monetary values and percentages. More details can be seen in the office website\footnote{https://cs.nyu.edu/\~grishman/NEtask20.book\_2.html}. The example of this dataset is shown as follows.~\\ 

\begin{mdframed}
\noindent\texttt{\scriptsize{{text: "Taga Co.",\\
type: "ORGANIZATION".
}}}
\end{mdframed}~

The MUC conference was replaced by Automatic Content Extraction (ACE) after 1997. ACE05~\citep{walker2006ace} is another popular NER dataset published at ACE Conference. ACE05 is a multi-lingual dataset, which contains English, Arabic, and Chinese data. The corpus consists of data of various types annotated for entities, relations, and events. Its data source includes broadcast conversation, broadcast news, newsgroups, telephone conversations, and weblogs. More details can be seen on the office website\footnote{https://catalog.ldc.upenn.edu/LDC2006T06}. The example of this dataset is shown as follows.

\begin{mdframed}
\noindent\texttt{\scriptsize{{entity id: "NN\_ENG\_20030630\_085848.18-E1",\\
type: "GPE",\\
subtype: "State-or-Province",\\
class: "SPC",\\
start: "82",\\
end: "91",\\
name: "california".
}}}
\end{mdframed}

After MUC, the Text Analysis Conference (TAC) published the Knowledge Base Population challenge. In this challenge, the Stanford NLP Group developed TAC Relation Extraction Dataset (TACRED)~\citep{zhang2017tacred}, which contains 106,264 instances with annotated entities, relations and some other NLP tasks. More details can be seen on the office website\footnote{https://nlp.stanford.edu/projects/tacred/\#intro}. The example of this dataset is shown as follows.~\\ 

\begin{mdframed}
\noindent\texttt{\scriptsize{{id: "e7798fb926b9403cfcd2",\\
docid: "APW\_ENG\_20101103.0539",\\
relation: "per:title",\\
token: "[`At', `the', `same', `time', `,', `Chief', ...]",\\
subj\_start: "8",\\
subj\_end: "9",\\
obj\_start: "12",\\
obj\_end: "12",\\
subj\_type: "PERSON",\\
obj\_type: "TITLE",\\
stanford\_pos: "[`IN', `DT', `JJ', `NN', `,', `NNP', `NNP',  ...]",\\
stanford\_ner: "[`O', `O', `O', `O', `O', `O', `O', `O', ...]"\\
stanford\_head: "[4, 4, 4, 12, 12, 10, 10, 10, 10, 12, ...]",\\
stanford\_deprel: "[`case', `det', `amod', `nmod', `punct', ....]".
}}}
\end{mdframed}

CoNLL-2003~\citep{sang2003introduction} is another widely used NER dataset. This task concerned language-independent named entity recognition, which concentrates on four kinds of named entities: locations, persons, organizations, and names of miscellaneous entities that do not belong to the previous three kinds. The related data files are available in English and German. More details can be seen on the office website\footnote{https://www.clips.uantwerpen.be/conll2003/ner/}. The example of this dataset is shown as follows.~\\ 

\begin{mdframed}
\noindent\texttt{\scriptsize{{text: "[`U.N.', `official', `Ekeus', `heads', ...], ",\\
pos: "[`NNP', `NN', `NNP', `VBZ', ...], ",\\
syntactic chunk: "[`I-NP', `I-NP', `I-NP', `I-VP', ...], ",\\
named entity tag: "[`I-ORG', `O', `I-PER', `O', ...]".
}}}
\end{mdframed}

Besides the above famous datasets, MultiNERD~\citep{tedeschi-navigli-2022-multinerd}, HIPE-2020~\citep{ehrmann_extended_2022}, and NNE~\citep{Ringland-acl-2019} are also popular NER datasets in general domain. NER tasks have garnered considerable attention in numerous specialized domains. Informatics for Integrating Biology and the Bedside (I2B2)~\citep{stubbs2015annotating} is a national biomedical computing project sponsored by the National Institutes of Health (NIH) from 2004 to 2014. I2B2 actively advocates mining medical value from clinical data and has organized a series of evaluation tasks and workshops for unstructured medical record data, and these evaluation tasks and open datasets have gained wide influence in the medical NLP community. I2B2 is maintained in the Department of Biomedical Information at Harvard Medical School and continues to conduct assessment tasks and workshops, and the project has been renamed National NLP Clinical Challenges (N2C2). More details can be seen on the office website\footnote{\url{https://www.i2b2.org/}}. Besides, there also exist many other biomedical datasets for specific medical NER tasks, including Adverse Drug Events (ADE)~\cite{Gurulingappa2012a,Alvaro2017},  Drug-Drug Interaction~\cite{Herrero-Zazo2013}, and Chemical Protein Reaction (CPR)~\cite{Krallinger2017}, and GENIA~\citep{shibuya-hovy-2020-nested}.

\subsection{Knowledge Bases}\label{sec:knowledgebases_ner}

\begin{table}[!tbh]
\centering
\scriptsize
\begin{tabular}{@{}llll@{}}
\toprule
Name & Knowledge & \# Entities & structure \\ \midrule
Wikipedia & World & 13,489,694 & unstructured \\
Wikidata & World & 100,905,254 & graph \\
DrugBank & Medical & over 500,000 & structured \\
UMLS & Medical & 16,857,345 & structured \\
BioModels & Medical & unclear & structured \\
SNOMED CT & Medical & over 350,000 & structured \\
ICD-10 & Medical & unclear & structured   \\
MIMIC-III & Medical & unclear & structured \\
MeSH & Medical & over 28,000 & structured \\
GeoNames & Geographical & over 25,000,000 & structured \\
EDGAR & Financial & unclear & structured \\
EduKG & Educational & 5,452 & structured \\ \bottomrule
\end{tabular}
\caption{Useful knowledge bases for NER.}\label{tab:knowledge_base_NER}
\end{table}

Table~\ref{tab:knowledge_base_NER} illustrates useful knowledge bases for NER. The biggest ones are Wikidata\footnote{\url{https://www.wikidata.org/}} and  Wikipedia\footnote{\url{https://en.wikipedia.org/}}, which are multi-lingual free online encyclopedias maintained by worldwide volunteers.

There are also knowledge bases in a specific field. SNOMED CT (Systematized Nomenclature of Medicine - Clinical Terms)~\citep{donnelly2006snomed} is a systematically organized collection of medical terms that provides a standardized representation of clinical information, which is often used in NER tasks involving clinical data. MeSH (Medical Subject Headings)~\citep{lipscomb2000medical} is another controlled vocabulary, developed by the U.S. National Library of Medicine. It is used for indexing and organizing biomedical literature. Other medical knowledge bases include UMLS (Unified Medical Language System)~\citep{wheeler2007database,bodenreider2004unified}, ICD-10~\citep{hirsch2016icd}, MIMIC-III~\citep{johnson2016mimic}, DrugBank~\citep{wishart2018drugbank}, and bioinformatics knowledge base BioModels~\citep{li2010biomodels}. GeoNames~\citep{ahlers2013assessment} is a comprehensive geographic knowledge repository that encompasses over 25 million geographical names and comprises over 11 million distinctive features, including cities, countries, and landmarks. EDGAR (Electronic Data Gathering, Analysis, and Retrieval)~\citep{branahl1998edgar} is a database maintained by the U.S. Securities and Exchange Commission (SEC), containing financial filings and reports from publicly traded companies.
EduKG~\citep{hu2016approach} is an educational knowledge base. 

\subsection{Evaluation Metrics}\label{sec:evaluation_metrics_ner}

In the process of named entity recognition task evaluation, the main evaluation metrics are also Precision, Recall, and F-value.

\subsection{Annotation Tools}\label{sec:annotation_tools_ner}

\noindent \textbf{One AI}\footnote{\url{https://docs.oneai.com/docs}} is an online platform that offers NLP-as-a-service. The utilization of APIs enables developers to effectively analyze, manipulate, and transform natural language inputs within their programming code without requiring any specialized knowledge of NLP. One AI facilitates the interpretation of both the meaning and information conveyed in textual data, and can produce structured data in context via language processing.

\noindent \textbf{GATE Teamware}\footnote{\url{https://gate.ac.uk/teamware/}}~\citep{bontcheva2013gate} is an integrated annotation tool for comprehensive language processing tasks, especially for Information Extraction systems.
The University of Sheffield developed GATE Teamware that enables collaborative semantic annotation projects through a shared annotation environment. The software comprises several beneficial attributes such as the ability to load document collections, create project templates that can be used multiple times, initiate projects based on templates, assign project roles to individual users, monitor progress and obtain various project statistics in real-time, report project status, annotator activity, and statistics, and apply automatic annotations or post-annotation processing via GATE-based processing routines.

\noindent \textbf{MAE}\footnote{\url{https://keighrim.github.io/mae-annotation/}}~\citep{rim2016mae2} (Multi-document Annotation Environment) is a general-purpose and lightweight natural language annotation tool. The tool enables users to specify and create their customized annotation tasks, annotate any text spans of their choice, utilize non-consuming tags, effortlessly establish links between annotations, and produce annotations in stand-off XML format. It also provides a simple adjudication process with a visualization feature that displays the extent tags, link tags, and non-consuming tags of any XML standoff annotated documents.

\noindent \textbf{UIMA}\footnote{\url{https://uima.apache.org/sandbox.html}}~\citep{ferrucci2004uima} (Unstructured Information Management Applications) is a framework that falls under the purview of the Apache Software Foundation. It serves as a comprehensive platform for managing language processing projects and is licensed under Apache's open-source license. With its versatile capabilities, UIMA can effectively handle a diverse array of language processing tasks and extract various types of information. The UIMA's Regular Expression Annotator is capable of identifying entities such as email addresses, phone numbers, URLs, zip codes, or any other entities based on the utilization of regular expressions and concepts. The tool can generate an annotation for each detected entity or update an existing annotation with relevant feature values.

\noindent \textbf{Brat}\footnote{\url{https://brat.nlplab.org/}} (Browser-based Rapid Annotation Tool) is a free data labeling tool that offers a seamless browser-based interface for annotating text. It streamlines numerous annotation tasks related to natural language processing. With a thriving support community, Brat is a well-known and widely used tool in NER. It also offers the option of integrating with external resources, such as Wikipedia. Moreover, Brat enables organizations to establish servers that allow multiple users to collaborate on annotation tasks. However, implementing this feature does necessitate some technical proficiency and server management skills.

\subsection{Methods}\label{sec:methods_ner}

\subsubsection{Nested NER}\label{sec:task1_ner}

\noindent \textbf{A. Multi-label Method}

Due to the fact that nested named entities can have multiple labels for a single token, traditional sequence labeling methods are not directly applicable to the recognition of nested named entities. To address this issue, researchers have attempted to convert the multi-label problem into a single-label problem or adjust the decoder to assign multiple labels to the same entity.

\citet{katiyar2018nested} proposed a method to address nested named entity recognition by modifying the label representation in the training set. Instead of using one-hot encoding, they used a uniform distribution over the specified classes as the label. During inference, a hard threshold is set and any class with probability above this threshold is predicted for the token. However, this approach has two limitations: it is difficult to determine the objective for model learning; the method is sensitive to the manually chosen threshold value.

\citet{strakova-etal-2019-neural} changed nested NER from multi-label to single-label tasks by modifying the annotation schema.
They combined any two categories that may co-occur to produce a new label (e.g., combine B-Location with B-Organization to construct a new label $B\_Loc\_Org$). One benefit of this approach is that the final classification task is still a single category because all possible classification targets had been covered in the schema. Nonetheless, this method brought about a proliferation of label categories in an exponential manner, leading to sparsely annotated labels that proved difficult to learn, particularly in the context of entities nested across multiple layers.

In order to address the issue of label sparsity,~\cite{shibuya-hovy-2020-nested} proposed a hierarchical approach. If the classification of nested entities cannot be resolved in a single pass, the classification is continued iteratively until either the maximum number of iterations is reached or no new entities can be generated. Nevertheless, this approach is susceptible to error propagation, whereby an erroneous classification in a preceding iteration could impact subsequent iterations.

\noindent \textbf{B. Generation-based Method}

\citet{li-etal-2020-unified} proposed a unified framework to accomplish flat and nested NER tasks by formulating NER as a machine reading comprehension (MRC) task~\citep{liu2023semantic}. In this approach, the extraction of each entity type corresponds to specific questions. For instance, when the model is given the question ``which location is mentioned in the sentence?'' along with the original sentences, it generates an answer such as ``Washington''. This approach is similar to Prompt Tuning~\citep{liu2021gpt}, which avoids the labor-intensive process of constructing manual questions. However, in this method, the generated tokens must be mapped to pre-defined named entity types.

\citet{yan-etal-2021-unified-generative} proposed a novel pointer generation network. Given an input sentence, the model generates the entity indexes in this sentence that belong to entities. In such a way, flat, nested, and discontinuous entities can be recognized in a unified framework. \citet{skylaki2020named,fei2021rethinking,yang2021bottom,su2022global} are also following the idea of generating indexes of a sentence to recognize nested entities.

\noindent \textbf{C. Hypergraph-based Method}

\begin{figure}[ht]
\centering
\subfloat[Instance of a nested label result]{\includegraphics[scale=0.5]{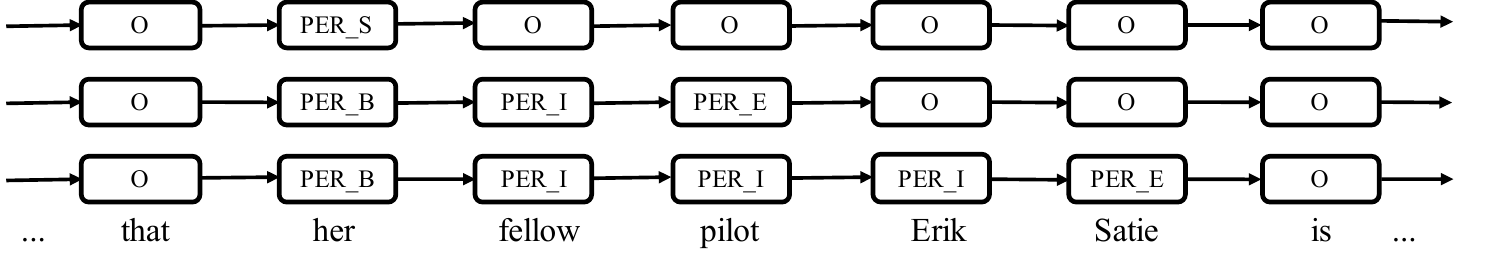}}
\hfill
\subfloat[Corresponding hypergraph structure]{\includegraphics[scale=0.5]{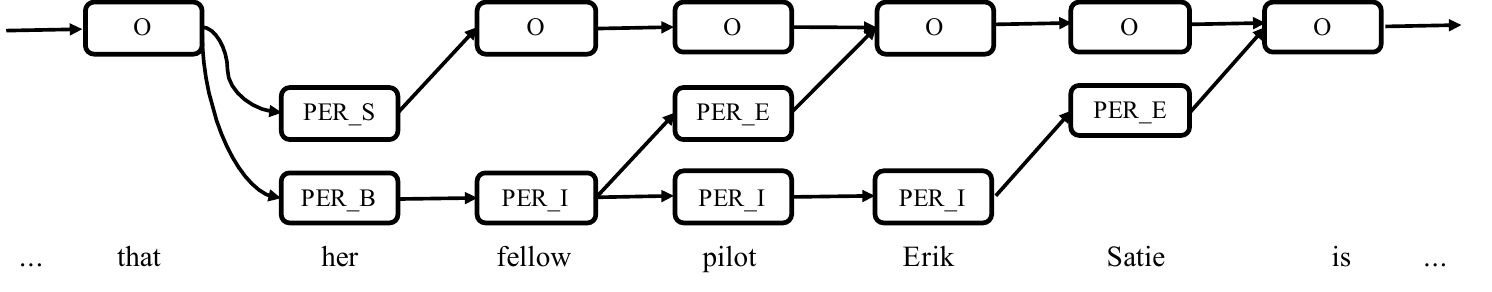}}
\caption{A typical example for nested NER with hypergraph solution}
\label{example_nest}
\end{figure}

A hypergraph is a generalized variant of a normal graph, which is characterized by an edge that can connect an arbitrary number of vertices~\citep{feng2019hypergraph}. It is widely used in the NLP community for the tasks of syntactic parsing, semantic parsing, and machine translation because it can accurately describe the relationship between objects with multiple associations. A set of objects with only binary relations can be described by a normal graph. However, when the objects are often related to each other in a more complex one-to-many or many-to-many, e.g., nested named entities, hypergraphs become a more appropriate data structure.
A typical example of nested NER with a hypergraph solution is shown in Figure~\ref{example_nest}.

\citet{finkel2009nested} firstly introduced hypergraphs into nested NER tasks, named Mention Hypergraph. In their model, Mention Hypergraph utilized nodes and directed hyper-edges to jointly represent named entities and their combinations. To compute the training loss, the proportion of accurate structures was calculated and divided by a normalized term. This term was obtained using a dynamic programming algorithm that aggregated feasible nested sub-graphs for NER. However, the normalized terms obtained from this algorithm included fractions of pseudo-structures, which led to errors. 

To deal with the problem of pseudo-structures,~\citet{muis2017labeling} proposed a gap-based marker model to identify nested entity structures by combining mention separators with features. In this method, the authors manually designed 8 types of mention separators for various scenarios. Based on the mention separators' states for any two consecutive tokens, they defined accurate and novel graph structures. However, since this approach only utilized local information to construct the graph structures, it may not be unambiguous for long-nested named entities. For instance, when presented with the nested entity ``a West African Crocodile'', which includes two separate entities, ``West African'' and ``a West African Crocodile'', their approach may also recognize ``a West African" as a named entity.

This ambiguous problem was solved by~\citet{wang-lu-2018-neural}, which proposes a segmental hypergraphs method. The method used an unambiguous ambiguity-free compact hypergraph representation to encode all possible combinations of nested named entities. Upon Mention Hypergraph~\citet{finkel2009nested}, segmental hypergraphs employed an inside-outside message-passing algorithm that can summarize the features of child nodes to the parent node and achieve efficient interference. 

Besides the above work,~\citet{wan2021region}~introduced the concept of regional hypernodes and a combination method of graph convolutional network (GCN) and BiLSTM to generate hypernodes for each region. \citet{yan2022local}~employed start token candidates and generated corresponding queries with related contexts, then used a query-based sequence labeling module to form a local hypergraph for each candidate.

\subsubsection{Few-shot NER}\label{sec:task2_ner}

\noindent \textbf{A. Metric Learning}

Metric Learning is a common technology in various few-shot tasks.
Prototypical Networks~\citep{snell2017prototypical} is a milestone in few-shot metric learning. Prototypical Networks compute the centroid of each category based on the support set. They determine the distance between the samples in the query set and the prototype center, followed by updating the model by optimizing this distance. Upon completion of the training phase, the embedding of each sample will be situated in closer proximity to the centroid of the corresponding category. Such an idea was largely inspired by Prototype Theory (see Section~\ref{sect:Prototype Theory}).

\citet{fritzler2019few} adopted the prototypical network into few-shot NER tasks. They argued that words in a sentence are interdependent and, therefore, the labeling of adjacent words should be taken into account. To address this issue, they substituted the conventional token input of Prototypical Networks with complete sentences. However, this method ignores the problem of the Outside (O) class in NER tasks, which actually represent different semantic meanings. This problem would significantly affect the model's performance under few-shot settings.

To avoid the above issues,~\citet{yang-katiyar-2020-simple} followed the nearest neighbor inference~\citep{wiseman-stratos-2019-label} to assign labels to tokens. In contrast to Prototypical Networks, which learn a prototype for each entity class, this study characterized each token by its labeled instances in the support set alongside its context. The approach determined the nearest labeled token in the support set, followed by assigning labels to the tokens in the query set that require prediction.

\citet{das-etal-2022-container} proposed CONTaiNER, which optimized the inter-token distribution distance. CONTaiNER employed generalized objectives to different token categories based on their Gaussian-distributed feature vectors.
Such a method has the potential to mitigate overfitting problems that arise from the training domains.

\noindent \textbf{B. Prompt Tuning}

\begin{figure}[!htbp]
\centering
\includegraphics[scale=0.7]{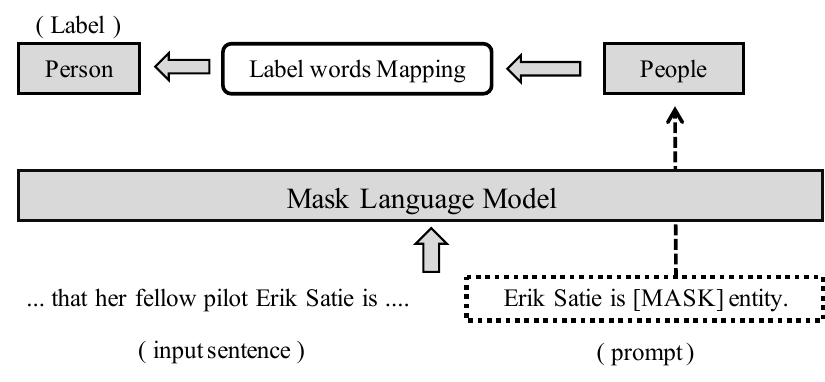}
\caption{A typical prompt tuning example for NER tasks.}
\label{prompt_tuning}
\end{figure}

Recently, prompt tuning has shown great potential on few-shot tasks by reformulating other tasks as mask language tasks~\citep{he2023virtual,mao2023biases,Timo2021It}.
Prompt tuning-based methods need construct prompts to obtain masked word predictions and then map predicted works into pre-defined labels, as shown in Figure~\ref{prompt_tuning}.

\citet{cui2021template} proposed a template-based method for NER, which first applied the prompt tuning to NER tasks. However, their method had to enumerate all possible spans of sentences combined with all entity types to predict labels, which suffered serious redundancy when entity types or sentence lengths increased. 

Manually defined prompts were labor-intensive and made the algorithm sensitive to these prompts. To avoid the manual prompt constructions,~\citet{ma2021template} tried to explore a prompt-free method for few-shot NER. The present study introduced an entity-oriented language model that decodes input tokens into their corresponding label words if they belong to entities. In cases where the tokens are not entities, the entity-oriented language model decodes the original tokens. Nevertheless, this approach encounters difficulties in labeling word engineering. While this study proposed an automated label selection technique, the associated experiments revealed some degree of instability.

COPNER~\citep{COPNER} introduced class-specific words to construct prompt tuning. By comparing each token with manually selected class-specific words, this method needed neither manual prompts nor label words engineering. The selected class-specific words (a representative word corresponding to a class) were directly concatenated with original sentences as prompts. However, the manual selection of class-specific words is subjective, and a single word may not entirely capture the semantics of an entity category.

\subsubsection{Joint NER and Relation Extraction}\label{sec:task43_ner}

\noindent \textbf{A. Parameter Sharing-based Multi-tasks Learning}

Considering that NER is usually combined with relation extraction tasks applied in various downstream tasks, jointly recognizing named entities and classifying relations is a hot topic in related fields. Multi-task learning is the most common solution in joint NER and relation extraction. \citet{Miwa2016} firstly employed a shared Bi-LSTM encoder to obtain token representations, and then fed encoded representations into NER and relation extraction classifiers, respectively. \citet{Sun2020} utilized a GCN as a shared encoder to enable joint inference of both entity and relation types. The core idea of the above study is that multi-task models can enhance the interactions between the learning of NER and relation extraction, and further alleviate the error propagation by sharing common parameters~\citep{he2021construction}. However, this work cannot ensure that the sharing of information is useful and proper. NER and relation extraction might need different features to result in precise predictions.

To deal with such a problem,~\citet{yan-etal-2021-partition} proposed an information filtering mechanism to provide valid features for NER and relation extraction. Their method used an entity and relation gate to divide cell neurons into different parts and established a two-way interaction between NER and relation extraction. In the final employed network, each neuron contained a shared partition and two task-specific partitions.

\noindent \textbf{B. Table Filling}

\begin{figure}[!htbp]
\centering
\includegraphics[scale=0.56]{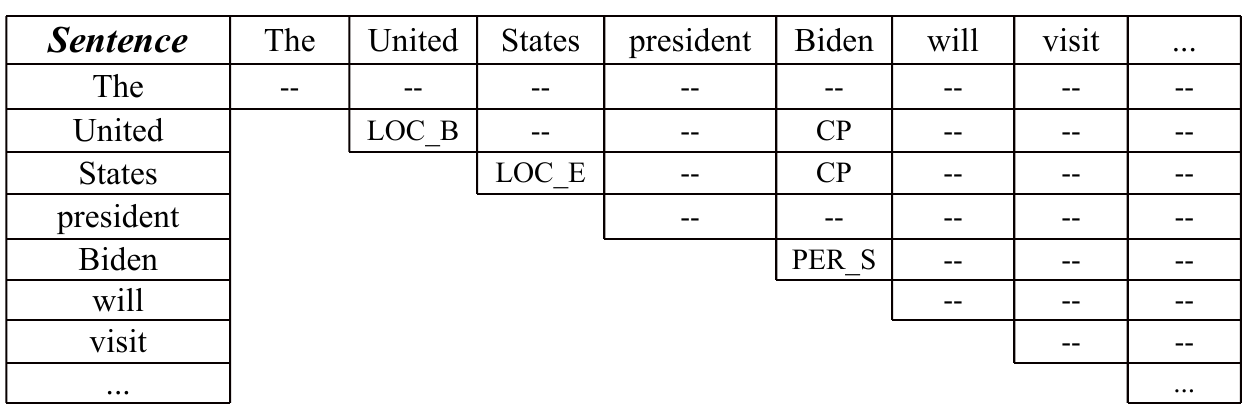}
\caption{The illustration of the table-filling strategy.}
\label{table_filling}
\end{figure}

While multi-task learning can improve the interdependence between NER and relation extraction, the relation extraction process still requires the pairing of all entities from the NER tasks to classify relations, making it impossible to completely eliminate error propagation. To solve the problem,~\citet{Miwa2014} proposed a table-filling strategy to achieve joint NER and relation extraction by labeling input tokens in a table. The method utilized token lists of sentences to form rows and columns. Then, they extracted entities using the diagonal elements and classified relations with a lower/upper triangular matrix of the table. This basic table-filling strategy can be seen in Figure~\ref{table_filling}. Nonetheless, this approach involved the explicit integration of entity-relation label interdependence, which necessitated the use of intricate features and search heuristics.

\citet{gupta2016table} incorporated neural networks with a table-filling strategy via a unified multi-task recurrent neural network. This method detected both entity pairs and the related relations with an entity-relation table, which alleviated the need for search heuristics and explicit entity-relation label dependencies. \citet{Zhang2017a} further integrated global optimization and syntax information into the table-filling strategy to combine NER and relation extraction tasks. \citet{ren-etal-2021-novel} argued that the above table-filling-based studies only focus on utilizing local features without the global associations between relations and pairs. \citet{ren-etal-2021-novel} first produced a table feature for every relation, followed by extracting two types of global associations from the generated table features. Finally, the table feature for each relation was integrated with the global associations. Such a process is performed iteratively to enhance the final features for joint learning of NER and relation extraction tasks.

\noindent \textbf{C. Tagging Scheme}

The table-filling approach can mitigate issues related to error propagation. However, these techniques require the pairing of all sentence elements to assign labels, resulting in significant redundancy. To address the redundancy and avoid error propagation,~\citet{zheng-etal-2017-joint}  proposed a novel tagging scheme that converted joint NER and relation extraction into a united task. The idea was similar to the solution for nested entities~\citep{strakova-etal-2019-neural}, which combined NER labels with relation extraction labels by modifying the annotation schema. For example, given the sentence ``The United States president Biden will visit ...'', by allocating the customized labels ``Country-President\_B\_1'', ``Country-President\_E\_1'' for tokens ``United", ``States'', and ``Country-President\_E\_2'' for token ``Biden'', the proposed method can directly obtain the triplet (United State, Country-President, Biden).

\citet{Yu2020,wei2019novel} proposed two similar methods. In contrast to conventional joint approaches for NER and relation extraction, which involve recognizing entities followed by relation classification, the two methods first identified all head entities. Next, for each identified head entity, they simultaneously predicted corresponding tail-entities and relations, achieving cascade frameworks combined with a customized tagging scheme. The typical joint NER and relation extraction tasks learn to model the conditional probability:
\begin{equation}
    P(h, r, t) = P(s)P(t \mid h)P(r \mid h,t),
\label{tagging_scheme_1}
\end{equation}
where $h$ represent head entity; $r$ represent relation; $h$ represent tail entity. The above methods combined the last two parts in Eq.~\ref{tagging_scheme_1}, yielding
\begin{equation}
    P(h, r, t) = P(s)P(t,r \mid s).
\label{tagging_scheme_2}
\end{equation}

\subsection{Downstream Applications}\label{sec:downstream_applications_ner}

\subsubsection{Knowledge Graph Construction}\label{sec:application3_ner}
Knowledge graphs are structured semantic knowledge bases for rapidly describing concepts and their interrelationships in the physical world, aggregating large amounts of knowledge by reducing the data granularity from the document level to the instance level~\citep{yao2022data}. Thus, knowledge graphs enable rapid response and reasoning about knowledge. At present, the application of knowledge graphs has become prevalent in industrial domains, such as Google search. Generally, the construction of Knowledge Graphs consists of three main parts: information extraction, information fusion, and information processing. The task of information extraction involves the identification of nodes through NER and the establishment of edges via relation extraction. The task of information fusion is utilized for normalizing nodes and edges. The normalized nodes and edges need to go through a quality assessment with the task of information processing to be added to knowledge graphs.

\citet{he2021construction} proposed a multi-task learning-based method for the construction of genealogical knowledge graphs. At first,~\citet{he2019extracting} collected unstructured online obituary data. Then, they extracted named entities as nodes and classified family relationships for these recognized people as edges to construct genealogical knowledge graphs. Similarly,~\citet{jiang2020biomedical} utilized NER and relation extraction for obtaining the nodes and edge in biomedical knowledge graphs. They proposed a customized tagging schema to convert the construction of biomedical knowledge graphs into a sequence labeling task with multiple inputs and multiple outputs. \citet{li2020real} proposed a systematic approach for constructing a medical knowledge graph, which involves extracting entities such as diseases and symptoms, as well as related relationships, from electronic medical records. \cite{silvestri2022iterative},~\cite{peng2019transfer}, and~\cite{shafqat2022standard} aimed to collect and utilize medical knowledge for NER. Further, constructing knowledge graphs requires the task of Entity Linking~\citep{tedeschi2021named} to normalize entities with different names. Entity Linking and NER are typically performed as pipeline tasks to yield more nodes and edges for the constructed graphs. Additionally, Entity Linking can be seen as a downstream task for NER, as it further refines the identified entities by linking them to a specific reference entity in a knowledge graph.

\subsubsection{Recommendation Systems}\label{sec:application2_ner}
 
Recommendation systems can be classified into two primary categories based on their solutions, namely content-based recommenders and collaborative filtering-based recommenders~\citep{batmaz2019review}. For both of these groups, gathering data on users and products is a crucial step in the entire process. In this regard, NER modules play a pivotal role. For example,~\citet{kim20125w1h} introduced the 5W1H model, which utilizes NER to extract contextual information, specifically Who, Why, Where, What, When, and How, to generate contextual recommendations.

\citet{zhou2020improving} argued that recommendation systems currently in use suffer from a deficiency of contextual information in conversational data, as well as a semantic gap between natural language expressions and the preferences of individual users for specific items. To overcome these challenges, word- and entity-oriented knowledge graphs were incorporated to enhance the data representations. Mutual Information Maximization was adopted to align the word-level and entity-level semantic spaces. The aligned semantic representations were used to develop a knowledge graph-enhanced recommender component to make accurate recommendations, and a knowledge graph-enhanced dialog component that can generate informative keywords or entities in the response text. A NER module is a crucial component in creating such a knowledge graph-enhanced system~\citep{wu2023megacare}.

\citet{iovine2020conversational} proposed a domain-independent, configurable recommendation system framework, named ConveRSE (Conversational Recommender System framEwork). ConveRSE utilized various interaction mechanisms, including natural language, buttons, and a combination of the two. The framework comprised a dialog manager, an intent recognizer, a sentiment analyzer, an entity recognizer, and a set of recommendation services. The entity recognizer component specifically focused on identifying relevant entities that were mentioned in the user's input, and linking them to an appropriate concept in the knowledge base. The ConveRSE framework's success is heavily reliant on the performance of the NER component, as it plays a crucial role in enhancing the system's overall performance.

\citet{wang2019exploring} proposed  RippleNet, an end-to-end framework that incorporates the knowledge graph into a recommender system. RippleNet overcame the limitations of previous embedding-based and path-based approaches to knowledge graph-aware recommendation by incorporating the knowledge graph as a form of supplementary information. RippleNet included both inward aggregation and outward propagation models. The inward aggregation version aggregated and incorporated neighborhood information when computing the representation of a given entity. By extending the neighborhood to multiple hops away, it was possible to model high-order proximity, thereby capturing users' long-distance interests. On the other hand, the outward propagation model propagated users' potential preferences and explored their hierarchical interests in knowledge graph entities.

\citet{upadhyay2021explainable} proposed an explainable job recommendation system by matching users with the most pertinent jobs, based on their profiles. The system also provided a human-readable explanation for each recommendation. The NER module was customized to extract pertinent details from both job postings and user profiles. These details were utilized to create comprehensible explanations for each recommendation. By identifying and categorizing entities, the NER module enhanced the accuracy and understandability of the textual explanations, providing a clear representation of the reasoning behind the recommendation system.

\subsubsection{Dialogue Systems}\label{sec:application1_ner}
Commonly, dialogue systems can be categorized into three main types, namely task-oriented, question-answering, and open-domain~\citep{ni2022recent}. NER plays a role in enhancing the natural language understanding of the three types of dialogue systems, organizing original user messages into semantic slots, and classifying data domain and user intention~\citep{li-etal-2017-end}. \citet{abro2022natural} proposed an argumentative dialogue system with NER and other natural language understanding tasks. The approach can enhance comprehension of user intent by comprehending injected entities and relationships. For the question-answering~\cite{dimitrakis2020survey} and open-domain dialogue systems, NER also plays a crucial role in the part of intent recognition and knowledge retrieval. For example,~\cite{zhang-etal-2021-kers-knowledge} developed a sequence of sub-goals with external knowledge to improve generation performance. External knowledge refers to a range of named entities and relationships that are associated with a conceptual entity. Leveraging external knowledge allows the dialogue system to deliver a more cohesive small talk from the open domain.

\subsection{Summary}\label{sec:summary_ner}

NER is a very important semantic processing technique for information retrieval. It is the manifest of cognitive semantics, because named entities are not simply categorized by their semantics. The classified named entities also reflect their inherent attributes in people's cognition. According to Prototype Theory (see Section~\ref{sect:Prototype Theory}), the inherent attributes of named entities can be represented by prototypes. It is gratifying to observe that a theory has had a significant influence on research related to few-shot NER. On the other hand, the ambiguity of named entity classification argued by Graded Membership (see Section~\ref{sect:Graded Membership}) and Grammatical Category (see Section~\ref{sect:Grammatical Category}) was rarely analyzed from computational linguistic aspects. We also do not see explainable NER studies that explain why an entity is classified into a particular category from the perspective of conceptual blending (see Section~\ref{sect:Conceptual Blending}). The NER research on these aspects is helpful for achieving human-like intelligence in categorizing named entities.

The availability of numerous named entity recognition (NER) datasets, both in general and medical domains, has significantly enhanced computational research in this area. This may be attributed to the great application value of NER, as well as a wide range of data annotation tools. Encyclopedias knowledge and domain-specific knowledge also provide external information to help NER models better understand the context and commonsense. Now, NER has developed many practical task setups to the need of technical applications, e.g., nested NER, few-shot NER, joint NER and relation extraction, and downstream tasks, e.g., knowledge graph construction, recommendation systems, and dialogue systems.

\subsubsection{Technical Trends}\label{sec:summary_technical_ner}

\begin{sidewaystable}[!htbp]
\scriptsize
\centering
\begin{tabular}{@{}llllllll@{}}
\toprule
Task & Reference  & Tech  & Feature and KB. & Framework & Dataset & Score & Metric \\ \midrule
\multirow{8}{*}{Nested NER}  &~\cite{katiyar2018nested}  & DL  & Emb. & Bi-LSTM & ACE-05  & 70.2\%  & F1 \\
 &~\cite{strakova-etal-2019-neural}  & DL  & Emb. & LSTM-CRF  & ACE-05  & 84.3\% & F1 \\
 &~\cite{shibuya-hovy-2020-nested} & DL  & Emb. & LSTM-CRF  & ACE-05  & 84.3\% & F1 \\
 &~\cite{li-etal-2020-unified} & DL  & BERT, Wikipedia  & Unified Framework & ACE-05  & 86.9\% & F1 \\
 &~\cite{yan-etal-2021-unified-generative} & DL  & BERT & Pointer Networks & ACE-05  & 84.7\% & F1 \\
 &~\cite{finkel2009nested} & Graph & Emb., Constituency Parsing  & Hypergraph  & GENIA & 72.0\% & F1 \\
 &~\cite{muis2017labeling} & Graph & Emb., Multigraph Representation & Hypergraph  & GENIA & 70.8\%  & F1 \\
 &~\cite{wang-lu-2018-neural}  & Graph & Emb., Segmental Hypergraphs & Hypergraph  & GENIA & 75.1\%  & F1 \\ 
 &~\cite{yang2021bottom} & DL  & BERT & Pointer Networks & ACE-05 & 85.0\%  & F1 \\ 
 &~\cite{su2022global}  & DL & BERT & Pointer Networks & CONLL04 & 88.6\%  & F1 \\ 
 \midrule
\multirow{6}{*}{\begin{tabular}[c]{@{}l@{}}Few-shot NER\\ 
(5 shot)\end{tabular}} &~\cite{fritzler2019few}$^\ast$  & DL  & Prototypical network  & RNN+ CRF  & Ontonotes & - & F1 \\
 &~\cite{yang-katiyar-2020-simple} & DL  & BERT & Nearest Neighbor  & I2B2  & 22.1\%  & F1 \\
 &~\cite{das-etal-2022-container}  & DL  & BERT & Contrastive  Learning  & I2B2  & 31.8\% & F1 \\
 &~\cite{cui2021template}  & DL  & BERT  & Prompt Tuning  & I2B2  & 36.7\% & F1 \\
 &~\cite{ huang-etal-2022-copner} & DL  & BERT & Prompt Tuning  & I2B2  & 43.7\% & F1 \\
\midrule
\multirow{6}{*}{\begin{tabular}[c]{@{}l@{}}Joint NER~ \\ 
~and RE \end{tabular}}  &~\cite{Miwa2016}  & DL  & Emb. & Bi-LSTM & ACE-05  & 55.6\%  & F1 \\
 &~\cite{Sun2020}  & Graph & Emb.,  & Bipartite Graph & ACE-05  & 59.1\%  & F1 \\
 &~\cite{yan-etal-2021-partition}  & DL  & BERT  & Partition Filter  & ACE-05  & 66.8\%  & F1 \\
 &~\cite{gupta2016table} & ML  & Emb., & Table filling  & CoNLL04 & 72.1\%  & F1 \\
 &~\cite{Zhang2017a} & DL  & Emb., & Table filling  & ACE-05  & 57.5\%  & F1 \\
 &~\cite{zheng-etal-2017-joint}  & DL  & Emb.,  & Tagging scheme  & NYT & 49.5\%  & F1 \\
 &~\cite{Yu2020} & DL  & Emb.,  & Tagging scheme  & NYT & 59.0\%  & F1 \\ 
 \midrule
\multirow{3}{*}{\begin{tabular}[c]{@{}l@{}} Task-driven NER  \end{tabular}}  &~\cite{shafqat2022standard}$^\ast$ & DL  & Emb., ICD-10 & Bi-LSTM  & no public & -  & F1 \\
 &~\cite{hirsch2016icd}$^\ast$  & DL  & Emb., UMLS & Bi-LSTM  & no public  & -  & F1 \\ 
 &~\cite{peng2019transfer}$^\ast$ & DL  & BERT, MIMIC-III & Fine Tuning  & no public  & -  & F1 \\ 
\bottomrule
\end{tabular}
\caption{A summary of representative NER techniques. The study with $^\ast$ means it cannot be compared with other studies since it did not report 5-shot results.}
\label{tech-trend}
\end{sidewaystable}

Due to the extensive research conducted on typical NER methods over the years, researchers are shifting their focus towards NER techniques that are more applicable to practical scenarios, for example, nested NER, few-shot NER, and joint NER and relation extraction. Recent technological trends for the aforementioned NER tasks are summarized in Table~\ref{tech-trend}.

Overall, nested NER can be addressed by multi-label, generation-based, and hypergraph-based methods. Among them, multi-label methods are straightforward and easy to implement. However, there are several limitations in the surveyed multi-label methods. For example, thresholds for multi-label selection are hard to decide empirically~\citep{katiyar2018nested}; multiple labels are suffering sparsity~\citep{strakova-etal-2019-neural} or error propagation~\citep{shibuya-hovy-2020-nested}, which can lower model performance. Generation-based methods are flexible. By reformulating NER tasks as question-answering, they can generate any results which satisfied the pre-defined requirements~\citep{shibuya-hovy-2020-nested,li-etal-2020-unified}. These methods are used for handling Flat NER~\cite{skylaki2020named}, nested NER~\citep{yan-etal-2021-unified-generative}, and discontinuous NER~\cite{fei2021rethinking}. However, a generation-based method is hard to control what is generated, even if some studies~\citep{skylaki2020named,fei2021rethinking,yang2021bottom,su2022global} have attempted to restrict the outputs of generation-based methods to a specific set of indexes (pointer network). The core point of the hypergraph-based method is about how to establish a hypergraph data structure to better represent interaction among all tokens in a sentence. These methods are good at modeling the interactions among all tokens in a sentence. It is important to note that the majority of hypergraph-based methods exhibit a task-specific nature, indicating a limited scope of applicability. These methods may not be universally applicable, and their effectiveness may be constrained by the specific task they are designed for.

Few-shot NER is usually achieved by metric learning and prompt tuning. Metric learning has demonstrated its effectiveness in various few-shot tasks~\citep{kaya2019deep, fritzler2019few}. For few-shot NER tasks, some works predict the final labels by comparing token-to-token distance~\citep{ yang-katiyar-2020-simple,das-etal-2022-container} or token-to-prototype distance~\citep{COPNER}. These methods have to decide different distance calculation functions according to different task~\citep{kulis2013metric} and suffer instability introduced by insufficient data. By exploiting the full potential of language models, prompt tuning is proposed and demonstrated as a promising technology for few-shot tasks~\citep{liu2021gpt,he2023virtual,liu2022p}. Prompt tuning reformulate NER as a mask language model task to reduce the gap between NER and employed pre-training LMs. The backward is that prompt tuning needs extra template construction and label word mappings and some studies have tried to deal with such problems~\citep{COPNER}.

For Joint NER and RE tasks, we summarize related studies into three groups, including parameter sharing-based multi-task learning, table-filling strategy, and customized tagging scheme. Parameter sharing is the basic idea in multi-task learning, which can be used to enhance the interaction between NER and RE~\citep{li2017neural,bekoulis2018joint}. This method can provide some relief from error propagation, but it cannot completely eliminate the issue. Also, this method has to pair every two entities for relation extraction, which introduces unnecessary redundancy. Table filling-based joint NER and relation extraction can completely eliminate error propagation by converting NER and relation extraction into a whole sequence-tagging task~\citep{gupta2016table,ren-etal-2021-novel,ma2022named}. However, these methods have to label every two token pairs in an input sentence in an enumerable fashion. If relation extraction is defined as an unidirectional task, the half of calculations are wasted. Following the idea of the table filling strategy, tagging scheme-based methods also model the NER and relation extraction as an integrated task. The fundamental concept of the tagging scheme is to merge the labels assigned for NER with those assigned for relation extraction into a unified label~\citep{zheng-etal-2017-joint, strakova-etal-2019-neural}. Such a method has the potential to circumvent issues related to both error propagation and redundancy; however, it may also lead to a sparsity of labels.

\subsubsection{Application Trends}\label{sec:summary_application_ner}

\begin{table}[!htbp]
\scriptsize
\centering
\begin{tabular}{llccc}
\toprule
Reference & Downstream Task & Feature & Parser & Explain. \\ \midrule
\cite{yao2022data} & Knowledge Graph Construction & & \checkmark & \\
\cite{he2021construction} & Knowledge Graph Construction & & \checkmark & \\
\cite{jiang2020biomedical} & Knowledge Graph Construction & & \checkmark & \\
\cite{li2020real} & Knowledge Graph Construction & & \checkmark & \\ 
\cite{kim20125w1h} & Recommendation Systems & & \checkmark & \checkmark \\
\cite{adomavicius2011context} & Recommendation Systems & \checkmark & & \\
\cite{zhou2020improving} & Recommendation Systems & \checkmark & & \\
\cite{iovine2020conversational} & Recommendation Systems & \checkmark & & \\
\cite{wang2019exploring} & Recommendation Systems & & \checkmark & \\
\cite{li-etal-2017-end} & Dialogue Systems & & \checkmark & \\
\cite{abro2022natural} & Dialogue Systems & & \checkmark & \\
\cite{dimitrakis2020survey} & Dialogue Systems & & \checkmark & \\
\cite{zhang-etal-2021-kers-knowledge} & Dialogue System & & \checkmark & \checkmark \\
\bottomrule
\end{tabular}
\caption{A summary of the representative applications of NER in downstream tasks. \checkmark denotes the role of NER in a downstream task.}\label{app-trend_ner}
\end{table}

We have discussed three main downstream applications of NER, including knowledge graph construction, dialogue systems, and recommendation systems. Table~\ref{app-trend_ner} illustrate related studies. Usually, NER is the basic module for providing recognized entities for further utilization. In this case, a NER model works as a parser to mine knowledge from unstructured text. The recognized entities and relations can be used as nodes and edges for knowledge graph construction. The entities can also serve as intent recognition methods in recommendation systems, and slot-filling methods in dialogue systems. For example,~\cite{wu2020tod} proposed a pre-trained task-oriented dialogue BERT, which significantly boosts the performance of a dialogue system by improving the intent detection sub-task. \cite{wang2020slot} proposed a method for recognizing related spans and value normalization with slot attention to improve the dialogue system. Besides, we also observe that using the identified named entities as features can also improve the performance of recommendation systems, because NER can help identify important entities that could be useful for making recommendations.

The most common problem is error propagation between NER and other components in a downstream system. \citet{kim2018two} employed a two-step neural dialog state tracker to alleviate the impact of the original error. With the development of PLMs and LLMs, many downstream tasks are organized as end-to-end processing tasks to achieve higher accuracy and mitigate error propagation issues. However, we can still observe that NER can improve the explainability in recommendation and dialogue systems~\citep{kim20125w1h,zhang-etal-2021-kers-knowledge}, which is also an important aspect of AI research. There is still a considerable untapped potential for integrating NER with other downstream tasks, e.g., explaining how concepts blend each other between different entities; what the inherent attribute of a group of entities the selected prototypes represent; how robust an identified named entity is.

\subsubsection{Future Works}\label{sec:summary_future_ner}

\noindent \textbf{Open-domain NER.} Compared with typical single-domain NER, open-domain NER has more categories. Besides, the entity classes are hardly defined in advance. For such reason, open-domain NER is more capable of handling rapidly expanding data, and mining more potential knowledge which is hidden in massive unstructured text data~\citep{hohenecker2020systematic,kolluru2020openie6}. Open-domain NER is significant because it discovers and connects world knowledge via automatic text mining. Many manually developed lexical resources, e.g., WordNet can only cover limited concepts. When the concepts come to multi-word expressions, manually mining, structuring and updating those concepts can result in the exponential growth of human efforts. Open-domain NER is helpful for mitigating human efforts and delivering a knowledge base that connects entities from different domains.

\noindent \textbf{Multi-lingual NER.} In light of the fact that a significant number of languages in existence lack sufficient annotated data, knowledge transfer from high-resource languages to low-resource languages can serve as a viable solution to compensate for the paucity of data~\citep{rahimi2019massively,tedeschi2021wikineural}. Developing robust multi-lingual NER systems that can perform across multiple languages will achieve more comprehensive knowledge graphs, linking entities from different languages. It is valuable because it may lead to a united concept representation system covering different languages. On the other hand, the task of developing multi-lingual NER systems is fraught with difficulties, primarily due to the inherent dissimilarities in entity types and language structures across different languages. As a result, aligning entities and transferring knowledge learned from one language to another can present significant challenges for multi-lingual NER systems.

\noindent \textbf{Unified framework for NER.} In the real-world scenario, there exist flat-named entities, nested entities, and discontinuous entities. Most NER-related studies only focus on the combination of flat with nested entities or flat discontinuous entities. Both of them cannot recognize all kinds of entities. Developing a unified framework to simultaneously handle such a problem becomes an urgent need for NER~\citep{fei2021rethinking}. Hierarchical concept representation knowledge bases may provide a preliminary ontology that can be used for organizing entities and their relationships. However, most of the ontology systems were manually developed by experts. This manually constructed knowledge may be invalid in specific application scenarios. A potential avenue for future research in NER is the development of a unified and robust framework for organizing entities. Such a framework could facilitate the creation of comprehensive knowledge graphs that capture the relationships between entities and can better support downstream tasks.

\noindent \textbf{Continual-learning for NER.} Humans exhibit a remarkable aptitude for transferring acquired knowledge from one task to another and retain their ability to perform the former task even after learning the latter. This ability is called continuous learning or life-long learning, which a regarded as an important characteristic of an intelligent system. Also, such ability can help us continue to use already deployed models when a new class of entity to be identified appears, rather than developed a new model from scratch~\citep{de2021continual}. There are some exploratory studies started to pay attention to such a problem. However, a satisfactory solution has not been found yet and existing methods still suffer the severe Catastrophic Forgetting~\citep{monaikul2021continual,xia2022learn,vijay2022nerda}. Continual learning is a critical skill for NER because NER is corpus-dependent. It is very important to update entity collections and the associated label sets, when a new corpus arrives~\citep{he2022jcbie}. In this case, detecting new entities and new labels with a former trained NER model represents a challenging yet highly promising research avenue.

\section{Concept Extraction}
\label{sect:Concept Extraction}

Concept extraction is a process to extract concepts of interest from the text. To our best knowledge, the task of computational concept extraction was first proposed by~\citet{DBLP:journals/coling/Montgomery82}, which analyzed the next 5 years of evolutionary progress in contemporary military message routing systems, with a focus on their transition towards becoming more advanced and knowledge-based systems. They argue that taxonomic hierarchies could be constructed to allow property inheritance of concepts, and therefore to perform rudimentary inference and analogic reasoning based on the taxonomies. \citet{DBLP:journals/coling/Montgomery82} also highlighted two important sub-tasks of concept extraction for the next-generation knowledge-based systems from the perspective of 1982, namely lexicon development and conceptual structure construction.

Recent research on concept extraction has been conducted in various fields of AI research, including natural language processing~(NLP) and data mining~\citep{miner2012practical}. Keyphrase generation~\citep{alami2020automatic} is one of the most common concept extraction tasks. It is a summarization task focusing on extracting keyphrases from a full passage to help readers quickly understand the passage, where keyphrases can be understood as the important concepts within a passage. Methods for keyphrase extraction can be both extractive~(copying from existing words) and abstractive~(not copying but summarizing and abstracting from existing texts). The process of generating keyphrases facilitates the creation of a lexicon that corresponds to a specific set of concepts. Another stream of concept extraction aims at the development of ontological knowledge bases to represent, e.g., commonsense knowledge~\citep{havasi2007conceptnet}, hypernym and synonym knowledge~\citep{snow2006semantic}, sentic knowledge~\citep{cambria2022senticnet}. These tasks tried to extract concepts to fit into pre-defined knowledge structures. Then, the structured knowledge can be directly used in downstream tasks.

Current concept extraction research is also grounded on related application scenarios, such as clinical concept extraction~\citep{fu2020clinical}, course concept extraction~\citep{pan-etal-2017-course}, and patent concept extraction~\citep{DBLP:conf/icdm/LiuWHWMLCTR20}. Clinical concept extraction is to transform massive unstructured electronic health records data into structured data; Course concept extraction is to extract important phrases in course captions to help to understand. Among them, clinical concept extraction is very similar to the information extraction task in NLP which aims at extracting most of the details in the unstructured text. Course and patent concept extraction are more similar to summarization tasks in NLP that target extracting important phrases. 

The main difference between concept extraction and NER tasks is that the extracted concepts or keyphrases are not identified by pre-defined entity classes. In contrast, they reflect the general idea of their contexts or target domain whose concepts are being discussed, while the goal of NER is to extract important factual information from the text. However, there are overlaps between NER and concept extraction when some concepts of interest, e.g., proper nouns can be also defined as named entities. Many domain-specific concept extraction tasks, e.g., clinical concept extraction, course concept extraction, and patent concept extraction can also be categorized as NER tasks because they aim at extracting concepts that are related to specific events. These events are also factual information. We review them in this section because they define themselves as concept extraction tasks in their original works. It also has become a trend of domain-specific concept extraction.

Another related field is relation extraction, which is a sub-field of information extraction. Relation extraction extracts information from raw text and represents it in the form of a semantic relation between entities~\citep{relation_extraction_survey}. The main difference is that, relation extraction targets at extracting relations between entities, while concept extraction targets at extracting noun entities. In knowledge graph development, relation extraction can help to connect nodes of concepts with purposeful relationships.

Concept extraction has also accelerated and contributed to multiple downstream applications, such as sentiment computing~\citep{cambria2022senticnet}, information retrieval~\citep{DBLP:conf/www/XiongPC17}, commonsense explanation generation~\citep{DBLP:conf/emnlp/FangZ22}. These applications mostly leverage explicitly extracted concepts.

Previous survey on concept extraction on focuses on clinical concept extraction~\citep{fu2020clinical}, which is a particular application field of concept extraction. In this section, we provide a more comprehensive review on concept extraction.

\subsection{Theoretical Research}\label{sec:theoretical_research_ce}

\subsubsection{Exemplar Theory}

\citet{medin1978context} argued that concepts are represented by a collection of particular exemplars or individual instances that are linked to the category. When we categorize an instance, we compare it with multiple specific exemplars of the category. This is different from Prototype Theory where a new instance is categorized by comparing the instance to the abstract prototype of the category (see Section~\ref{sect:Prototype Theory}). \citet{medin1978context} formed the task of concept categorization as a classification task, and conducted experiments with 32 participants. The experiments showed that the classification judgments made by participants were impacted by various factors. These factors included the extent of resemblance between the probe item and exemplars previously acquired, the number of prior exemplars that shared resemblances with the probe item, and the similarity present both within and between the categories of the previously learned exemplars. For concept extraction and categorization, Exemplar Theory may suggest that models may take categorized instances into account when categorizing a new instance.

\subsubsection{Semantic Primitives}

\citet{wierzbicka1972semantic} believed that it is possible to describe every human language by using a limited number of universal semantic primitives. These primitives are representative of fundamental concepts that form the basis of human communication and thinking. \citet{wierzbicka1972semantic} established 64 universal semantic primes, which consist of basic words or ideas that cannot be defined in relation to more elementary concepts. However, these primes can be utilized to describe all other concepts present within a language. Semantic Primitives suggest that concepts should be organized as multiple layers from the concrete to abstract ones. Decision-making that runs on concrete concepts can be completed through the upper-level abstract concepts that contain those concrete concepts. Thus, it is critical to represent the hierarchical and linking relationships between concepts. There are other theories mentioned before, e.g., Frame Semantics (see Section~\ref{sect:frame semantics}), that may guide concept structure development. Frame Semantics highlights the connection of related concepts, while Semantic Primitives suggest the hierarchical relationships between concepts and the distinction between primitive concepts and others.

\subsubsection{Conceptual Spaces}

\citet{gardenfors2004conceptual} defined concept as the ``theoretical entities that can be used to explain and predict various empirical phenomena concerning concept formation''. The author believed that concept representations are multi-dimensional, where each dimension is indicative of a different characteristic or property associated with the concept. For example, one could represent the concept of a car within a conceptual space that includes dimensions such as size, speed, color, and shape. This is very similar to current vectorial representations of words or entities in NLP, while the dimensionality of Conceptual Spaces is explainable by concept properties. \citet{gardenfors2004conceptual} also placed significant emphasis on the role of context in understanding and representing concepts. This is due to the fact that different contexts may emphasize different features or dimensions of concepts. Then, the connections between concepts are determined by the relationships between their property similarity in the conceptual space. For example, ``dog'' and ``cat'' are similar in the animal concept space, because their properties are similar; ``mammals'' can be separated from ``reptiles'' by a property difference boundary, although both are in the animal space. This may encourage concept extraction tasks to extract both concept entities and properties associated in contexts. This is because properties define how concepts are connected from the view of~\citet{gardenfors2004conceptual}.

\subsection{Annotation Schemes}\label{sec:annotation_schemes_ce}

From the goal of the keyphrase annotation aspect, there are in general two types of annotation schemes for keyphrase extraction-liked concept extraction. The first is to precisely select existing keyphrases from input text, but not to create semantically-equivalent phrases. The second is to both select existing keyphrases and create ``absent keyphrases'' that are necessary but do not exist in the input text~\citep{DBLP:conf/emnlp/Hulth03}.

From the format of assigned annotations aspect, there are in general two annotation schemes as well. The first scheme is to directly give the keyphrases existing in the source text. The second scheme treats the keyphrase extraction task as a sequence labeling task, and assigns a label to each of the tokens in source text~\citep{DBLP:conf/emnlp/Hulth03}. The assigned labels in the current dataset follow a BIO scheme defined in table~\ref{tag_scheme}. Specifically, three labels are used: B~(Beginning), I~(Inner), and O~(Other).

\subsection{Datasets}\label{sec:datasets_ce}

\begin{table*}[!htbp]
\centering
\scriptsize
\begin{tabular}{lllll}
\toprule
Dataset  & Task & Source  & \# Samples & Reference \\ \midrule
Inspec  & KE  & Inspec database  & 2,000 &~\citet{DBLP:conf/emnlp/Hulth03} \\
NUS & KE  & Google SOAP API  & 211 &~\citet{DBLP:conf/icadl/NguyenK07} \\
Krapivin  & KE  & ACM Digital Library  & 2,304 &~\citet{krapivin2009large} \\
SemEval2010 & KE  & ACM Digital Library  & 244 &~\citet{DBLP:conf/semeval/KimMKB10} \\
Twitter & KE  & Twitter  & 1,000 &~\citet{DBLP:conf/emnlp/ZhangWGH16} \\
KP-20K  & KE  & \begin{tabular}[c]{@{}l@{}}ACM Digital Library, \\ScienceDirect, \\and Web of Science\end{tabular} & 567,830 &~\citet{DBLP:conf/acl/MengZHHBC17} \\
CCF & KE  & China Computer Federation  & 13,449  &~\citet{DBLP:conf/icdm/WangLQXWCX18} \\
MLDBMD  & KE  & Academic Conferences & 128.1k  &~\citet{DBLP:conf/icdm/LiZ0Y18}  \\ \midrule
TempEval  & ClCE & Mayo Clinic  & 600 &~\citet{bethard2016semeval} \\
i2b2-2010 & ClCE & Clinical Records & 826 &~\citet{uzuner20112010} \\
n2c2-2018 & ClCE & Clinical Records & 505 &~\citet{henry20202018} \\
MIMIC & ClCE & MIMIC-III Database & 1,610 &~\citet{gehrmann2018comparing} \\ \midrule
MOOCs & CoCE & Coursera and XuetangX  & 4375 videos &~\citet{DBLP:conf/ijcnlp/PanWLLT17}  \\
EMRCM & CoCE & Chinese Textbooks  & 3,730 pages  &~\citet{DBLP:conf/icdm/HuangLWHMC0W19} \\ \midrule
USPTO & PCE & USPTO Database & 94,000  &~\citet{DBLP:conf/icdm/LiuWHWMLCTR20}  \\ 
\bottomrule
\end{tabular}
\caption{Concept extraction datasets and statistics. KE denotes Keyphrase Extraction. ClCE denotes Clinical Concept Extraction. CoCE denotes Course Concept Extraction. PCE denotes Patent Concept Extraction.}
\label{tab:ce_datasets}
\end{table*}

The surveyed popular concept extraction datasets and their statistics can be viewed in Table~\ref{tab:ce_datasets}. Overall the main thread of dataset development is (1) larger scale of datasets; (2) attending to both extractive keyphrases and abstractive keyphrases; (3) more fine-grained annotations for tags; (4) more application domains. \citet{DBLP:conf/emnlp/Hulth03} proposed one of the first keyphrase extraction datasets, termed Inspec. Their dataset is based on the scientific papers under {\it Computers and Control}, and {\it Information Technology} disciplines in the Inspec database. The keywords used in the scientific papers are selected as the keyphrases. Abstracts are used as the keyphrase extraction context. Keywords in scientific papers are used as keyphrases. Each abstract has two sets of keywords: a set of controlled terms, i.e., terms restricted to the Inspec thesaurus; and a set of uncontrolled terms that can be any suitable terms that may or may not be present in the abstracts. They collected 1000 abstracts as a train set, 500 as a validation set, and 500 as a test set.\\

\begin{mdframed}
\noindent\texttt{\scriptsize{{abstract: "[ `A', `scalable', `model', `of', `cerebellar', `adaptive', `timing', `and', `sequencing', `:', ...]"\\
doc bio tags: "[ `O', `B', `I', `O', `B', `I', `I', `O', `O', `O', ...]"\\
extractive keyphrases: "[ `scalable model', `cerebellar adaptive timing', ... ]"\\
abstractive keyphrase: "[ `cerebellar sequencing', ...]"
}}}
\end{mdframed}

\citet{DBLP:conf/icadl/NguyenK07} proposed the NUS dataset with the motivation that keyphrase extraction requires multiple judgments and cannot rely merely on the single set of author-provided keyphrases. They first used Google Search API to retrieve scientific publications, and then recruited student volunteers to participate in manual keyphrase assignments. They finally collect 211 documents, each with two sets of keyphrases: one is given by the original authors of the paper, and the other is given by student volunteers. The data format of NUS is the same as Inspec~\citep{DBLP:conf/emnlp/Hulth03}.

\citet{krapivin2009large} proposed the Krapivin dataset, consisting of around 2,000 scientific papers as well as their keywords assigned by the original authors. The scientific papers were published by ACM in the period from 2003 to 2005, and were written in English. One of the novelties of this dataset is that the text data in the scientific papers were collected with three distinct categories: title, abstract, and main body. They finally collect 460 test data and 1.84k validation data. The data format is similar to Inspec\citep{DBLP:conf/emnlp/Hulth03} but has a title and body in addition to the abstract.

SemEval-2010 Task 5~\citep{DBLP:conf/semeval/KimMKB10} is on automatic keyphrase extraction from scientific articles. Input for this task is a document from either of the four domains: distributed systems, information search, and retrieval, distributed artificial intelligence, and social and behavioral sciences. Outputs are manually annotated keyphrases for the document. This dataset contains 144 documents as a train set, and 100 documents as a test set. It also selects 40 documents from the train set to compose a trial set. For each set, documents are evenly distributed from the four topics. The annotation follows the first scheme in Section~\ref{sec:annotation_schemes_ce}. The data format is the same as Inspec~\citep{DBLP:conf/emnlp/Hulth03}.

\citet{DBLP:conf/emnlp/ZhangWGH16} constructed a keyphrase extract dataset from Twitter using an automatic text mining method. Their core assumption is that hashtags in a tweet can be used as keyphrases for the tweet. To construct the dataset, they first collected 41 million tweets, and then filtered them which contain non-Latin tokens. URL links, and reply tweets were removed. Thus, the remaining text only contains tweets and a hashtag. They finally kept 110K tweets. To evaluate the quality of the collected tweets, they sampled 1000 tweets and chose three volunteers to score them. As a result, 90.2\% tweets are suitable, and 66.1\% are perfectly suitable. The annotation follows the first scheme in Section~\ref{sec:annotation_schemes_ce}. \\

\begin{mdframed}
\noindent\texttt{\scriptsize{{tweet: "Hard to believe it but these are REAL state alternatives to taking  Obamacare \\ funds from the gov't (via @Upworthy)"\\
keyphrase: "obamacare"}}}
\end{mdframed}

\citet{DBLP:conf/ijcnlp/PanWLLT17} proposed a keyphrase extraction dataset, where data were sourced from online course captions. Labels are existing phrases in the captions. The courses are computer science and economics courses, selected from two famous MOOC platforms --- Coursera and XuetangX. Labels were first filtered from captions using automatic methods and then annotated by two human annotators. A candidate concept was only labeled as a course concept if the two annotators were in agreement. As a result, they collected captions from 4375 videos, and 16720 labeled concepts.\\

\begin{mdframed}
\noindent\texttt{\scriptsize{{course caption: "You might learn how to write a bubble sort and learn why a bubble sort is not as good as a heapsort."\\
keyphrase: "[ `bubble sort', `heapsort' ]"
}}}
\end{mdframed}

KP-20K~\citep{DBLP:conf/acl/MengZHHBC17} is a testing dataset, where the input texts are titles and abstracts of computer science research papers collected from ACM Digital Library. The labeled keyphrases are the keyphrases shown in the research papers. The annotation follows the second scheme in Section~\ref{sec:annotation_schemes_ce}, since the keyphrases given by authors were not necessarily existing keyphrases in the papers. KP-20K has the same data format as Inspec.

\citet{DBLP:conf/icdm/HuangLWHMC0W19} were motivated to automatically construct an educational concept map. The educational concept map shows concepts that will be learned in courses, as well as the temporal relation between the concepts~(e.g., to learn concept A, it is a prerequisite to learn concept B; Concept A and concept B can help with the understanding of each other). To construct the dataset written in Chinese, they first used OCR to obtain the text from textbooks, then manually labeled key concepts for each textbook~(as ``key concept'' or ``not key concept'') and finally manually annotated the relationships among the labeled key concepts~(as ``$w_i$ is $w_j$'s prerequisite'', ``$w_i$ and $w_j$ has collaboration relationship'', or ``no relationship''). As a result, they collected 3730 pages in textbooks, 1092 key concepts, 818 prerequisite relations, and 916 collaboration relations. However, in their GitHub repo, only keyphrases and relations between keyphrases can be found, while the text cannot be found.\\ 

\begin{mdframed}
\noindent\texttt{\scriptsize{{keyword: "[ `average', `weighted average', ... ]"\\
relation: "[ `average : weighted average', ... ]"
}}}
\end{mdframed}

There are concept extraction datasets focused on a specific domain, e.g., clinical concepts (TempEval~\citep{bethard2016semeval}, i2b2-2010~\citep{uzuner20112010}, n2c2-2018~\citep{henry20202018}, and MIMIC~\citep{gehrmann2018comparing}), course concepts (MOOCs~\citep{DBLP:conf/ijcnlp/PanWLLT17}, and EMRCM~\citep{DBLP:conf/icdm/HuangLWHMC0W19}), and patent concepts (USPTO~\citep{DBLP:conf/icdm/LiuWHWMLCTR20}). They also followed keyphrase extraction setups, whereas the targets are to extract concepts of interest.

\subsection{Knowledge Bases}\label{sec:knowledgebases_ce}

Besides classical lexicon resources such as WordNet, encyclopedias~(including Baidu Encyclopedias and Wikipedia) can also be used to provide external knowledge for concepts~\citep{DBLP:conf/ijcnlp/PanWLLT17}. Methods for extracting concepts based on embedding techniques may encounter issues with low frequency, where some of the concepts have infrequent occurrences. \citet{DBLP:conf/ijcnlp/PanWLLT17} utilize word embeddings~\citep{mikolov2013distributed}, which is trained on encyclopedias, to obtain the semantic embedding for each concept. Inspec database is a scientific and technical database storing scientific papers. The papers of this database have been used to construct a keyphrase extraction dataset.

\begin{table}[!htbp]
\centering
\scriptsize
\begin{tabular}{llll}
\toprule
Name               & Knowledge                    & \#Entities & Structure    \\ \midrule 
WordNet & Lexical & 155,327 & Tree \\
Baidu Encyclopedia & World                         & 6,223,649  & Unstructured \\
Wikipedia          & World & 9,834,664  & Unstructured \\
Inspec             & Science & 20,000,000 & Unstructured \\ 
\bottomrule
\end{tabular}
\caption{Useful knowledge bases for concept extraction.}
\label{tab:ce_kb}
\end{table}

\subsection{Evaluation Metrics}\label{sec:evaluation_metrics_ce}

The field of concept extraction also uses Precision, Recall, and F1-score as evaluation metrics. Some keyphrase extraction research considered the task as an information retrieval task. Then, the information retrieval metric, e.g., mean average precision (MAP) was also used for keyphrase extraction as the main measure. It is calculated by taking the average of the average precision scores for each query in a dataset.
\begin{equation}
    MAP = \frac{1}{n}\sum_{i=1}^{n}Avg\_Precision_i,
\end{equation}
where n is the total number of queries. $Avg\_Precision_i$ denotes the averaged precision of query $i$. In the context of keyphrase extraction, the MAP score is determined by comparing the generated list of keyphrases with a predefined gold standard set, and evaluating the average precision of the top $n$ keyphrases, where $n$ corresponds to the total number of keyphrases in the gold standard set. Each generated keyphrase is considered as a query; The gold standard set serves as the relevant document.

\subsection{Annotation Tools}\label{sec:annotation_tools_ce}

Since the annotation schemes of concept extraction are similar to that of NER. The aforementioned NER annotation tools can also be used for annotating concept extraction data. Numerous studies have investigated the utilization of pre-existing keywords in scientific publications~\citep{DBLP:conf/emnlp/Hulth03,DBLP:conf/icadl/NguyenK07,krapivin2009large,DBLP:conf/semeval/KimMKB10,DBLP:conf/emnlp/ChenZ0YL18} or hashtags in tweets~\citep{DBLP:conf/emnlp/ZhangWGH16}, whereby such in-context information can serve as labels without requiring additional annotation efforts, provided that the labels align with the research objectives.

\subsection{Methods}\label{sec:methods_ce}

\subsubsection{Keyphrase Extraction}\label{sec:key_phase}

The task of keyphrase extraction is to obtain keyphrases from a document to represent and summarize the document with the keyphrases. There are generally two trends of methods, namely extractive keyphrase extraction and generative keyphrase extraction.

Extractive methods appear first but have a systematic disadvantage in that they can only extract existing phrases in the documents. For example,~\citet{DBLP:conf/acl/MengZHHBC17} argued that in addition to present keyphrases, there are also absent keyphrases, which can better summarize a document but do not explicitly present in the document. Generative methods, however, can generate every possible word. Therefore generative methods can alleviate the disadvantage of extractive methods, but might be more difficult because it requires a model to accurately catch the ``semantic meaning'' of a document to precisely generate a keyphrase.

\noindent \textbf{A. Extractive Keyphrase Extraction}

\citet{DBLP:conf/emnlp/ZhangWGH16} focused on the task of keyphrase generation on Twitter data, and framed this task as a sequence labeling task. They proposed a joint-layer RNN model. For each input token, the joint-layer RNN model outputs two indicators~($\hat{y}_{1}$ and $\hat{y}_{2}$), where $\hat{y}_{1}$ has two values $True$ and $False$, indicating whether the current word is a keyword. $\hat{y}_{2}$ has 5 values $Single$, $Begin$, $Middle$, $End$ and $Not$ indicating the current word is a single keyword, the beginning of a keyphrase, the middle of a keyphrase, the ending of a keyphrase, or not a part of a keyphrase, respectively. Their experiments show that the joint-layer RNN model outperforms both the vanilla RNN model and the LSTM model. However, when $\hat{y}_{1}$ and $\hat{y}_{2}$ have contradictions, it might be hard to find an optimal strategy to determine which indicator to refer to. In addition, joint-layer RNN can only extract an existing sequence as a keyphrase, but cannot abstractively obtain a (non-existing but better) keyphrase.

\citet{DBLP:conf/icdm/WangLQXWCX18} hypothesized that the performance of keyphrase extraction could be improved in the unlabeled or insufficiently labeled target domain by transferring knowledge from a resource-rich domain. They accordingly proposed a topic-based adversarial neural network~(called TANN) that can learn transferable knowledge across domains efficiently by performing adversarial training. The experiment section shows that TANN largely outperforms joint-layer RNN~\citep{DBLP:conf/emnlp/ZhangWGH16}.

\citet{DBLP:conf/icdm/LiZ0Y18} proposed an unsupervised method for concept mining, which was motivated by the fact that supervised methods might be hard to generalize to unseen domains. They assumed that the quality of an extracted concept can be measured by its occurrence contexts and proposed a pipeline method for concept mining. The method first populates many raw concepts extracted from text, and then evaluates the concepts by comparing the embedding of concepts against the current local context.

\citet{DBLP:conf/www/AlzaidyCG19} identified two limitations of previous supervised approaches: 1) They classify the labels of each candidate phrase independently without considering potential dependencies between candidate phrases. 2) They do not incorporate hidden semantics in the input text. Correspondingly,~\citet{DBLP:conf/www/AlzaidyCG19} addressed keyphrase extraction as a sequence labeling task, and proposed a model named Bi-LSTM-CRF that unite both the advantages of LSTM~(capturing semantics) and CRF~(Conditional Random Field, capturing dependencies,\citep{DBLP:conf/icml/LaffertyMP01}). Their results show that Bi-LSTM-CRF outperforms CopyRNN~\citep{DBLP:conf/acl/MengZHHBC17} by a large margin.

\citet{DBLP:conf/ijcai/FangHHTHLH021} hypothesized that previous extractive methods ignore structured information in the raw textual data~(title, topic, and clue words), which might lead to worse performance. They accordingly proposed a model named GACEN that can utilize the title, topic, and clue words as additional supervision to provide guidance. GACEN also utilized CRF to model dependencies in the output. The experiment section shows that GACEN outperforms Joint-layer-RNN~\citep{DBLP:conf/emnlp/ZhangWGH16} and CopyRNN~\citep{DBLP:conf/acl/MengZHHBC17}.

\noindent \textbf{B. Generative Keyphrase Extraction}

\citet{DBLP:conf/acl/MengZHHBC17} were motivated that classic keyphrase generation methods can only extract the keyphrases that appear in the source text. Those methods are unable to reveal and leverage the full semantics for keyphrase ranking. Consequently, they proposed an RNN-based generative model incorporating a copying mechanism~\citep{DBLP:conf/acl/GuLLL16}~(named with CopyRNN), which can generate absent keyphrases. Their method uses an encoder-decoder architecture to catch the semantics of the input text.

Previous methods such as~\citet{DBLP:conf/acl/MengZHHBC17} suffered from both coverage~(not all keyphrases are extracted) and repetition~(similar keyphrases are extracted) problems. For the coverage issue,~\citet{DBLP:conf/emnlp/ChenZ0YL18} integrated a coverage mechanism~\citep{DBLP:conf/acl/TuLLLL16} into their approach, which enhances the attention distributions of multiple keyphrases in order to cover a wider range of information within the source document and effectively summarize it into keyphrases. For the repetition issue, they constructed a target side review context set that contains contextual information of generated phrases.

\citet{DBLP:conf/emnlp/YeW18} believed that although sequence-to-sequence~(seq2seq) models have achieved good performance, model training still relies on large amounts of labeled data. Correspondingly, they leveraged unsupervised learning methods such as TF-IDF and self-learning algorithms to create keyphrase labels for large amounts of unlabeled data. Then, they train their model with a mixture of self-labeled and labeled data together for training. They also used multi-task learning to train their model. Experiments show that their performance outperforms previous works.

\citet{DBLP:conf/aaai/ChenGZKL19} argued that prior research on keyphrase generation has treated the document title and main body in the same manner, overlooking the significant role that the title plays in shaping the overall document. They accordingly proposed a Title-Guided Network~(TG-Net) where the title is additionally employed as a query-like input to particularly assign attention to the title. The performance of TG-Net outperforms CopyRNN~\citep{DBLP:conf/acl/MengZHHBC17}. 
Their ablation study also shows the importance of additional attention to the title.

\subsubsection{Structured Concept Extraction}\label{sec:kbd_ce}

Compared with keyphrase extraction-liked concept extraction, structured concept extraction aimed to develop an ontology where concepts are connected with each other by certain relationships. Here, we introduce three knowledge bases resulting from concept extraction: WordNet, ConceptNet, and SenticNet. Out of them, WordNet focuses more on a word-level ontology, ConceptNet focuses more on a concept-level ontology~(e.g., also including phrases for concepts), and SenticNet is a concept-level ontology focusing on contributing to sentiment analysis tasks.

WordNet is a manually developed knowledge base, where words and concepts are hierarchically organized. \citet{snow2006semantic} proposed a taxonomy induction method to expand WordNet 2.1 concepts by automatic noun hyponym acquisition, achieving 10,000 novel synsets with 84\% precision. Compared to previous methods that relied on individual classifiers to uncover new relationships based on pre-designed or automatically extracted textual patterns, the proposed approach considers input from multiple classifiers to enhance the overall structure of the taxonomy and prioritizes the optimization of the entire taxonomy structure with a probabilistic architecture. \citet{snow2006semantic} also proposed an (m,n)-cousin classification-based model to learn coordinate terms, which allows it to integrate heterogeneous evidence from different classifiers and choose the correct word sense to which to attach a new hypernym. The evaluation of the inferred taxonomies produced by the algorithm was conducted by directly comparing them with the WordNet 2.1 taxonomy. This was achieved by testing each taxonomy using a set of human judgments of noun pairs sampled from newswire text, to determine the hypernym and non-hypernym relationships.

ConceptNet~\citep{havasi2007conceptnet} grew out of Open Mind Common Sense project that aimed at commonsense acquisition. Contributors delivered knowledge by fulfilling blanks within a sentence, For example, given ``[~~] can be used to [~~]'', the concepts, e.g., ``ink'' and ``print'' and the associated relationship ``UsedFor'' can be obtained. ConceptNet aimed to obtain and structure concepts automatically from natural language. It obtained concepts (the nodes) in the form of noun phrases, verb phrases, adjective phrases, prepositional phrases, or complete verb phrases~\citep{havasi2007conceptnet}. The edges of ConcepNet are predicates that represent the relationships between two concept nodes, such as ``IsA'', ``PartOf'', UsedFor, and more. \citet{havasi2007conceptnet} defined 21 basic relation types. In the latest ConceptNet 5.5~\citep{speer2017conceptnet}, the relations are increased to 36. Concepts and predicates were obtained via pattern matching. Each collected sentence is compared with pre-defined regular expressions, e.g., ``NP is used for VP''(UsedFor), ``NP is a kind of NP''(IsA), ``NP can VP'' (CapableOf). NP (noun phrases) and VP (verb phrases) are concepts, while ``UsedFor'', ``IsA'', and ``CapableOf'' are predicates. In the case of a complex sentence that contains several clauses, the patterns are employed to extract a simpler sentence from it, which can then be subjected to the concept and predicate extraction process. To evaluate ConceptNet, its assertions were compared with those in similar lexical resources to determine their alignment.

SenticNet is a commonsense knowledge base that is used for affective computing. The concepts were extracted by a graph-based semantic parsing method~\citep{cambria2014senticnet} and assigned with sentiment polarity labels. Sentences are divided into chunks, e.g., ``go walk'', first. Then, verb-noun chunks are normalized by stemming, and included in the concept set. The PoS-based bigram algorithm is used to extract object concepts. To capture event concepts, the approach explores matches between object concepts and normalized verb-noun chunks. Finally, single-word concepts, e.g., ``house'' that have appeared in the clause as multi-word concepts ``beautiful house'' are deemed redundant and are therefore excluded. In the following version of SenticNet~\citep{cambria2016senticnet}, the authors proposed an automatic method to discover primitives from the SenticNet concepts, based on hierarchical clustering and dimensionality reduction. Thus, the ``animal'' concept can be identified as the primitive of ``cat'', ``dog'', or ``pet''. Later,~\citet{DBLP:conf/icdm/CambriaMHL22} proposed a pipeline method for concept extraction, which is used for expanding SenticNet with multi-word expressions. They first deconstructed text using sentence chunking, semantic parser, and PoS tagging. Then, verb and noun chunks are extracted and normalized as concepts. The proposed method offers novel contributions in utilizing morphology for syntactic normalization and employing primitives for semantic normalization. The method was evaluated on a sentiment analysis task, achieving explainable and primitive- and concept-level sentiment analysis via algebra operations. The latest version of SenticNet~\citep{cambria2022senticnet} offers the function that sentiment predictions can be effectively conducted on the primitive level, mitigating symbol grounding problems.

An important task of concept extraction is to abstract concept representations from entities. Unlike SenticNet which obtains abstract concepts (primitives) by selecting the most typical entities from a group of extracted similar entities~\citep{cambria2016senticnet},~\citet{ge2022explainable} proposed a conceptualization method that can directly abstract concepts from input text. The task is realized in the metaphor identification and interpretation domain. The authors aimed to generate concept mappings from metaphorical word pairs to explain the metaphoricity of the word pairs. For example, given ``\textit{blind} alley'', ``\textsc{street} is \textsc{adult}'' can be automatically generated. This work is the realization of conceptual metaphor theory~\citep{lakoff1980metaphors} (see Section~\ref{sect: conceptual metaphor}) that the generated concept mapping explains the mapping of source (e.g., \textsc{adult}) and target (e.g., \textsc{street}) concepts of a metaphor. The conceptualization (e.g., from ``alley'' to ``\textsc{street}'') was achieved by selecting the most appropriate hypernym on the chain from the leaf node of ``alley'' to the root node ``entity'' in WordNet. The most appropriate hypernym is defined as the node that can cover the major senses of the leaf, meantime, keeping it as concrete as possible\footnote{Intuitively, ``entity'' can cover all possible senses of the ``alley'' in WordNet, while it is not the ideal concept representation of ``alley'', because it is too abstract. Thus, the authors aimed at a concrete concept representation that can cover the majority senses of a word.}. The conceptualization and concept mapping method was evaluated on a metaphor identification task, yielding better performance and explainability on the task. Subsequently, within MetaPro Online~\citep{mao2023metaproonline}, the conceptualization algorithm is synergistically integrated with sequential metaphor identification and interpretation techniques, culminating in the attainment of end-to-end concept mapping generation from full sentences.

\subsubsection{Domain-specific Concept Extraction}\label{sec:domain-specific_ce}

\noindent \textbf{A. Clinical Concept Extraction}

The task of clinical concept extraction is to extract structural information from unstructured clinical narratives~\citep{fu2020clinical}. \citet{DBLP:conf/semeval/LiH16a} constructed a dataset for a seminal task called ``UTA-DLNLP at SemEval-2016 Task 12'' for clinical concept extraction. A system developed for this task should task raw clinical notes or pathology reports as input, and identify event expressions consisting of the ``the spans of the expression in the raw text'', ``contextual modality'', ``degree'', ``polarity'', and ``type''. As a baseline for this task, they propose a convolutional neural network to learn hidden feature representations for predictions, taking text and part-of-speech tags as input.

\citet{DBLP:journals/midm/LiuYWCTWX17} adopted BiLSTM to recognize the entity in clinical text. They found that BiLSTM outperforms the CRF baselines. \citet{gehrmann2018comparing} compared CNN with classic rule-based methods, bag of words, n-grams, and embedding-based logistic regression. They found that CNN is a valid alternative to rule-based and classic NLP methods, and should be further investigated. \citet{DBLP:journals/jamia/YangBHW20} comprehensively explored 4 widely used transformer-based architectures, including BERT~\citep{devlin2018bert}, RoBERTa~\citep{liu2019roberta}, ALBERT~\citep{DBLP:conf/iclr/LanCGGSS20}, and ELECTRA~\citep{DBLP:conf/iclr/ClarkLLM20}.
They compared the 4 models to long short-term memory conditional random fields~(LSTM-CRFs)~\citep{DBLP:journals/corr/HuangXY15} baselines and found that transformer-based models are effective for clinical concept extraction tasks. \citet{DBLP:conf/acl/LangeAS20} proposed a joint model for both clinical concept extraction and de-identification tasks. De-identification is important since in some clinical concept extraction scenarios, the privacy of patients should be protected. They hypothesized that jointly modeling the two tasks can be beneficial, and proposed two end-to-end models. One is a multitask model where the tasks share the input representation across tasks; the other is a stacked model, which used the privacy token predictions to mask the corresponding embeddings in the input layer and only use the masked embeddings for concept extraction. They found that the performance of the concept extraction model can be improved by training and evaluating it on anonymized data, thereby confirming their initial hypothesis.

\noindent \textbf{B. Course Concept Extraction}

In tasks involving the extraction of course concepts, the concepts are typically defined as the knowledge concepts that are taught in the course videos, as well as the related topics that aid in the students' comprehension of the course videos~\citep{DBLP:conf/ijcnlp/PanWLLT17}.
Identifying course concepts at a fine level is very important, as students with different backgrounds need different concepts to quickly understand the main content of a course~\citep{DBLP:conf/ijcnlp/PanWLLT17}.

\citet{DBLP:conf/ijcnlp/PanWLLT17} contributed the first attempt to systematically investigate the problem of course concept extraction in MOOCs. in the past, course concepts were presented by instructors at a general level, with only a few concepts being covered in an entire course video. However, they emphasized the significance of identifying course concepts at a granular level, i.e., automatically identifying all course concepts from each video clip, to facilitate easier comprehension. They identified a challenge for the task that the course concept appears at a low frequency mainly because the different courses have different concepts. They accordingly proposed to utilize word embedding to catch the semantic relations between words and incorporate online encyclopedias to learn the latent representations for candidate course concepts. They also proposed a graph-based propagation algorithm to rank the candidates based on learned representations.

\citet{DBLP:conf/dsc/WangF0Z18} argued that external knowledge must be involved to solve the concept extraction problem and proposed to utilize both the structured and unstructured data in Wikipedia to provide external knowledge to concept extraction. Their results show that their method outperforms prior works~\citep{DBLP:conf/ijcnlp/PanWLLT17}.

\noindent \textbf{C. Patent Concept Extraction}

\citet{DBLP:conf/icdm/LiuWHWMLCTR20} developed a framework to extract technical concepts from patents. Patent documents have different structures than other documents. For instance, they have ``title'', ``abstract'', and ``claim'', which exhibit a multi-level of information. Motivated by this, the authors proposed a framework named UMTPE, which can effectively leverage multi-level information to extract concepts.

\subsection{Downstream Applications}\label{sec:downstream_applications_ce}

\subsubsection{Sentiment Computing}

SenticNet 7~\citep{cambria2022senticnet} is a neuro-symbolic sentiment analysis system, based on SenticNet knowledge base. It assumes that concepts that share the same primitive would have similar sentiments. One can use algebra operations to achieve sentiment analysis with the symbolic and structural knowledge base. Incorporating a symbolic knowledge base and a transparent algorithm provides SenticNet's reasoning process with the benefit of interpretability and accuracy.

\citet{li2023skier} proposed a neuro-symbolic system for conversational emotion recognition. ConceptNet was used as a knowledge base to acquire commonsense knowledge out of context. For example, if a person mentions that he will ``chop all onions we have and cry'', another conversation participant expresses ``disgust'' emotion. This is because ``onion IsA lacrimator'' is a commonsense in ConceptNet. Such a commonsense cannot be obtained from the dictionary meanings of ``onion'' and the context, while ConceptNet commonsense knowledge provides the evidence and explainability to infer such an emotional status from the context. The authors used an utterance dependency parser and a neural network to learn symbolic knowledge to enhance the explainability and accuracy of their method.

By using the concept mapping method from the work of~\citet{ge2022explainable},~\citet{han2022hierarchical} used concept mappings to support depression detection and explanation. The hypothesis is that depression patients may have similar cognition patterns that are reflected in their metaphorical expressions. Thus, they used concept mappings as additional features besides tweets. The concept mappings were generated from tweets that contained metaphors. They also proposed an explainable encoder that can identify significant concept mappings that contribute to depression detection. The concept mappings also improve the accuracy of depression detection, besides explaining the common concept mapping patterns.

\subsubsection{Information Retrieval}\label{sec:application1_ce}

\citet{DBLP:conf/www/XiongPC17} manually analyzed the potential problems of a literature search website SemanticScholar.org, and found that the issue of ``Concept Not Understood'' represents one of the most significant challenges. The reason is that previous methods measure similarity based on text, but not on their semantic embeddings. As a result, they proposed an embedding-based similarity matching method, which extracts the concepts in both query and documents and measures the similarity between these concepts to obtain the similarity between a query and a document. \citet{DBLP:conf/kdd/LiuHHLCSH18} used extracted knowledge concepts as one of the inputs to obtain a unified semantic representation for educational excises. The representation is further used to retrieve similar excises based on similarity with other representations.

\subsubsection{Dialogue Systems}\label{sec:application3_ce}

\citet{young2018augmenting} integrated commonsense knowledge from ConceptNet in their dialogue system. They believed that in human dialogues, individuals responding to each other is not dependent on the most recent utterance only, but also on recollecting pertinent information related to the concepts addressed within the dialogue, e.g., commonsense. Thus, in retrieval-based dialogue generation, the model considers both the message content and relevant commonsense knowledge to effectively choose a suitable response.

\citet{huang2020grade} proposed a new dialogue coherence evaluation matric, termed Graph-enhanced Representations for Automatic Dialogue Evaluation (GRADE). \citet{liu2016not} argued that traditional BLEU-liked statistic-based metrics are biased in response coherence. Thus,~\citet{huang2020grade} were motivated to propose a metric that measures the coherence by the topics of utterances. They believe that a cohesive exchange of dialogues is characterized by a seamless transition between topics. Thus, they used a ConceptNet-based method to construct topic-level dialogue graphs. The topic-level dialogue graphs were constructed by connecting the concepts that are extracted from utterances. The edge was weighted and undirected, which was derived from the shortest path between two nodes in the ConceptNet. Such an evaluation metric can better represent the coherence of topics between utterances because it measures the relatedness of concepts from different utterances.

\subsubsection{Commonsense Explanation Generation}\label{sec:application4_ce}

\citet{DBLP:conf/emnlp/FangZ22} grounded concept extraction in the context of commonsense explanation generation. Commonsense explanation generation aims to generate an explanation in natural language to explain the reason why a statement is anti-commonsense. For example, given ``he took a nap in the sink'', the model aimed to generate ``a sink is too small and dirty to take a nap in''. The concepts, ``small'' and ``dirty'' (bridge concepts), are obtained via a prompt-tuning method. The authors developed a masked word prediction template to query the bridge concepts that are most likely to appear in the ``mask'' position. Then, they use a generator to generate the explanation with the concatenation of the original statement and the discrete bridge concepts. This method improves the explainability in explaining why a statement is anti-commonsense. 

\subsection{Summary}\label{sec:summary_ce}

A concept is an abstract idea that is reflected in the mind. Concept extraction is the foundation of detecting the main idea of a context and developing conceptual knowledge bases. Related theoretical research showed that concepts may be abstracted from multiple specific exemplars~\citep{medin1978context} or prototypes~\citep{rosch1973natural}. There are limited primitives that construct human cognition and reasoning, which are the foundation of complex concepts~\citep{wierzbicka1972semantic}. According to~\citet{gardenfors2004conceptual}, conceptual space is multi-dimensional. The similarity between concepts can be measured by the similarity between concept properties. These theoretical research works frame the tasks of concept extraction from the perspectives of lexicon development and conceptual structure construction. On the other hand, current computational concept extraction methods divide this task into three categories, namely keyphrase extraction, structured concept extraction, and domain-specific concept extraction. We found that the existing computational approaches inadequately address the tasks that have been put forth by the academic community's theoretical research. Although current concept extraction methods are limited, this task has greatly improved the explainability of downstream tasks such as sentiment computing, information extraction, and counter-commonsense recognition.

\subsubsection{Technical Trends}\label{sec:summary_technical_ce}

\begin{sidewaystable}[!htbp]
\centering
\scriptsize
\begin{tabular}{llllllll}
\toprule
Task & Reference & Techniques & Feature and KB & Framework & Dataset & Score & Metric \\ \midrule
\multirow{5}{*}{Extractive KE} &~\citet{DBLP:conf/emnlp/ZhangWGH16} & DL & word2vec & Joint-layer RNN & Twitter & 86.40\% & F1 \\
 &~\citet{DBLP:conf/icdm/WangLQXWCX18} & DL & word2vec & BiLSTM, adversarial loss & CCF & 29.60\% & F1 \\
 &~\citet{DBLP:conf/icdm/LiZ0Y18} & Pipeline & word2vec & Similarity matching & MLDBMD & 97.00\% & MAP \\
 &~\citet{DBLP:conf/www/AlzaidyCG19} & DL & word2vec & BiLSTM-CRF & KP-20K & 35.63\% & F1 \\
 &~\citet{DBLP:conf/ijcai/FangHHTHLH021} & DL & word2vec & Attention; CRF & KP-20K & 45.69\% & F1 \\ \midrule
\multirow{5}{*}{Generative KE} &~\citet{DBLP:conf/acl/MengZHHBC17} & DL & word2vec & RNN & KP-20K & 32.80\% & F1@5 \\
 &~\citet{DBLP:conf/emnlp/ChenZ0YL18} & DL & word2vec & Seq2seq & Krapivin & 31.80\% & F1@5 \\
 &~\citet{DBLP:conf/emnlp/YeW18} & DL & word2vec & Seq2seq, semi-supervised & KP-20K & 30.80\% & F1@5 \\
 &~\citet{DBLP:conf/aaai/ChenGZKL19} & DL & word2vec & \begin{tabular}[c]{@{}l@{}}Seq2seq, additional\\Title input\end{tabular} & KP-20K & 37.20\% & F1@5 \\ \midrule
\multirow{6}{*}{Structured CE} &~\citet{havasi2007conceptnet} & Knwl. eng. & textual patterns & Pattern matching & - & - & - \\
 &~\citet{snow2006semantic} & SL & feature vectors, WN & Probabilistic & - & - & - \\ 
 & SenticNet & \begin{tabular}[c]{@{}l@{}}chunking, \\sem. pars., \\PoS tag.\end{tabular} & syntactic patterns & Syntactic parsing & - & - & - \\
 &~\citet{ge2022explainable} & SL & statistics, WN & Elbow algorithm & - & - & - \\
 \midrule
\multirow{6}{*}{Clinical CE} &~\citet{DBLP:conf/semeval/LiH16a} & DL & \begin{tabular}[c]{@{}l@{}}token mention, pos\\ tag, word shape\end{tabular} & CNN & TempEval & 78.80\% & F1 \\
 &~\citet{DBLP:journals/midm/LiuYWCTWX17} & DL & word2vec, character2vec & BiLSTM & i2b2-2010 & 85.78\% & F1 \\
 &~\citet{gehrmann2018comparing} & DL & word2vec & CNN & MIMIC & 76.00\% & F1 \\
 &~\citet{DBLP:journals/jamia/YangBHW20} & DL & word2vec & Transformer & n2c2-2018 & 88.36\% & F1 \\
 &~\citet{DBLP:conf/acl/LangeAS20} & DL & word2vec & Multitask-biLSTM & i2b2-2010 & 88.90\% & F1 \\ \midrule
\multirow{2}{*}{Course CE} &~\citet{DBLP:conf/ijcnlp/PanWLLT17} & Graph & word2vec; Encyclopedia & Graph-based propagation & MOOCs & 41.60\% & MAP \\
 &~\citet{DBLP:conf/dsc/WangF0Z18} & Graph & word2vec, Wikipedia & Graph-based propagation & MOOCs & 47.50\% & MAP \\ \midrule
Patent CE &~\citet{DBLP:conf/icdm/LiuWHWMLCTR20} & ML & \begin{tabular}[c]{@{}l@{}}self pretrained \\word2vec, DBpedia\end{tabular} & Clustering & USPTO & 43.37\% & F1 \\ 
\bottomrule
\end{tabular}
\caption{A summary of representative concept extraction techniques. KE denotes keyphrase extraction. CE denotes concept extraction. SL denotes statistical learning. Knwl. eng. denotes knowledge engineering. SenticNet denotes the works of~\citet{cambria2014senticnet,cambria2016senticnet,cambria2022senticnet,DBLP:conf/icdm/CambriaMHL22}. We do not show the evaluation results for structured concept extraction methods, because they all used very task-specific evaluation methods and datasets, where the results are not comparable.}
\label{tac:ce_technical_trends}
\end{sidewaystable}

Within the domain of keyphrase extraction, generative keyphrase extraction takes advantage of generating ``absent keyphrases'', compared to extractive keyphrase extraction. Both tasks followed the general development of the NLP fields. They likely considered the task as a sequence labeling task (extractive keyphrase extraction) or a generation task (generative keyphrase extraction), and used typical NLP frameworks, e.g., sequence labeling and sequence to sequence frameworks. However, it is unclear if these general NLP frameworks have really learned how to summarize the main idea of context or just have learned by label distributions. There were no task-specific mechanisms proposed to explicitly learn the keyphrase extraction task on the concept level, with an explainable decision-making process. On the other hand, keyphrase extraction-based concept extraction is helpful for obtaining concept lexicons. However, compared to structured concept extraction, keyphrase extraction cannot learn the relationships between concepts. The theoretical research of Conceptual Spaces from~\citet{gardenfors2004conceptual} suggested that the similarity between concepts can be measured by their properties. It suggests that keyphrase extraction-based concept extraction should consider extracting properties together with keyphrases. Thus, the later works can use keyphrases and the associated properties to structure concepts by similarities.

In contrast, structured concept extraction research likely utilized statistical learning and syntactic parsing methods. This is because the aim of structured concept extraction is to develop a large knowledge base or detect structured relationships between concepts. Labeled data are insufficient in these areas. Thus, unsupervised methods are preferred. However, the concept knowledge base development is task-specific. As a result, the concepts in different knowledge bases share different relationships. For example, ConceptNet aimed to parse concepts sharing 36 commonsense relationships; Stanford WordNet~\citep{snow2006semantic} was expended in synonyms and hypernyms relationships; SenticNet grouped concepts and extract primitives for sentiment computing;~\citet{ge2022explainable} abstracted concepts for concept mappings. Then, the evaluation of different concept extraction methods is different. Most of the evaluation was implemented on different downstream tasks. It shows

Domain-specific concept extraction is very similar to NER tasks. They used graph, machine learning methods, and external knowledge, e.g., encyclopedias and Wikipedia to discover concepts in a domain, e.g., clinical, course, or patent concepts. Similar to keyphrase-based concept extraction, these domain-specific concept extraction methods did not try to structure concepts after extraction. This is important because it distinguishes concept extraction from current NER tasks in specific domains.

\subsubsection{Application Trends}\label{sec:summary_application_ce}

\begin{table}[!htbp]
\centering
\scriptsize
\begin{tabular}{llccc}
\toprule
Reference & Downstream Task & Feature & Parser & Explain. \\ \midrule
\citet{cambria2022senticnet} & Sentiment Computing & \checkmark & \checkmark & \checkmark \\
\citet{li2023skier} & Sentiment Computing & \checkmark & & \checkmark \\
\citet{han2022hierarchical} & Sentiment Computing & \checkmark & \checkmark & \checkmark \\
\citet{DBLP:conf/www/XiongPC17} & Information Retrieval & \checkmark & & \checkmark \\
\citet{DBLP:conf/kdd/LiuHHLCSH18} & Information Retrieval & \checkmark & & \checkmark \\
\citet{young2018augmenting} & Dialogue Systems & \checkmark & & \\
\citet{huang2020grade} & Dialogue Systems & \checkmark & \checkmark & \\
\citet{DBLP:conf/emnlp/FangZ22} & Commonsense Explanation Generation & \checkmark & & \checkmark\\ \bottomrule
\end{tabular}
\caption{A summary of the representative applications of concept extraction in downstream tasks.}
\label{tab:ce_downstream_apps}
\end{table}

Concept extraction methods and their product, e.g., knowledge bases have been widely used in downstream tasks, e.g., sentiment computing, information retrieval, dialogue systems, and commonsense explanation generation. Compared to other low-level semantic processing techniques, the roles of concept extraction are more diverse in downstream applications. For all the surveyed downstream tasks, the products of concept extraction can be used as additional features to improve model performance on downstream tasks. On the other hand, concept extraction techniques can be used as a parser to obtain knowledge from unstructured text. The structured concepts with certain relationships can also improve the explainability of a downstream task model, e.g., explaining anti-commonsense~\citep{DBLP:conf/emnlp/FangZ22} and concept mapping patterns of depressive patients~\citep{han2022hierarchical}.

In the era of PLM and LLM, it seems many complex tasks can be achieved from end-to-end with deep neural networks. However, black box-liked neural networks prevent humans from understanding their decision-making mechanisms. This may be contrary to the original intention of human beings to build AI, e.g., giving machines the ability to think like humans. Neuro-symbolic AI which combines the knowledge of symbolic representations with neural networks, seems to be able to compensate for the lack of model interpretability of pure neural networks because symbolic representations in natural language, e.g., words and concepts are human-readable. We can explain a prediction by viewing what symbolic knowledge is activated. Meantime, symbolic knowledge can represent commonsense knowledge, which is difficult for neural networks to learn from corpora. As the fundamental technique of knowledge base development, concept extraction has a huge potential in downstream applications.

\subsubsection{Future Works}\label{sec:summary_future_ce}

\noindent \textbf{Open domain concept extraction.} Prior research on concept extraction has primarily concentrated on extracting concepts within a particular domain, while other concept extraction efforts aimed at developing knowledge bases have focused on extracting concepts with predefined relations. These approaches severely limit the application scope of knowledge bases. It would be more practical to extract concepts and relations in an open domain, where both the concepts and relations are not focused on specific types. This requires an ontology study to guide the concept extraction, e.g., what can be defined as concepts and relations. It is a more challenging task than the joint NER and relation extraction task, because relationships and concepts are self-aware within a learning model, rather than pre-defined by humans.

\noindent \textbf{Multi-modal concept extraction.} ``Concept'' is also very relevant to human visual recognition. It is argued that for humans, the ability of visual classification is obtained from concept learning, which learned the generalized concept description from sample observations such that a given observation can be identified as a learned concept~\citep{seel2011encyclopedia,DBLP:conf/icml/XiongTW21}. On the other hand, the abstractness of concepts is strongly related to imagery~\citep{paivio1965abstractness}, because abstract concepts are those that are not applicable to tangible, perceptible objects that can be observed through touch, sight, hearing, or other sensory experiences~\citep{lohr2022abstract}. Thus, learning the relationships between concepts and imagery can help concept extraction research hierarchically organized concepts, e.g., primitives, concepts, and entities. However, till now, there is a lack of research papers working on multi-modal concept extraction to our best knowledge. It could be also interesting to investigate possible synergies in concept extraction between different modalities.

\noindent \textbf{Concept extraction evaluation.} Current concept extraction methods were evaluated on an application task, e.g., sentiment analysis to SenticNet or testing specific relationships, e.g., hypernym and hyponym relationship to ConceptNet and WordNet extension. The issue with such an evaluation method is that it can only reflect the effectiveness of a developed knowledge base or concept extraction method on a specific domain. Since different knowledge bases have different application targets, it's hard to evaluate and compare them with unified criteria. It would be valuable to propose a framework for knowledge base evaluation that is independent of specific tasks. It would be helpful to understand the quality of included concepts, relationships, and their representations.

\noindent \textbf{More concept extraction applications.} Despite the attention some scholars have given to neuro-symbolic AI, the body of related works remains relatively scant in comparison to end-to-end neural network models. One possible explanation for this disparity is that, at present, there is greater emphasis placed on the accuracy of the model rather than the transparency of its decision-making process. Thus, there is a need for more concept extraction applications, which can aid in enhancing the explainability of neural network-based models. It offers insights for the development of knowledge bases, prompting researchers to reassess how they extract and organize concepts in order to more effectively support subsequent applications.

\section{Subjectivity Detection}
\label{sect:Subjectivity Detection}

Conventionally, subjectivity detection is defined as a task to determine whether a text is subjective or not, where a subjective text expresses personal feelings, evaluations, and speculations~\citep{wiebe1994tracking}, whereas an objective one merely delivers factual information. Generally, subjectivity can manifest in different forms, e.g., opinions, allegations, desires, beliefs, and suspicions~\citep{liu2010sentiment} to express private states. It is not an easy task to identify the use of subjective language, as a subjective sentence does not always contain an opinion~\citep{liu2010sentiment}. Therefore, it is important for the subjectivity detection task to find reliable clues. Aside from opinion-bearing words, syntax also provides essential clues in reporting private states, because grammaticalization involves the recruitment of items to mark the speaker’s point of view~\citep{traugott2010revisiting}.

Early works often equated the presence of subjectivity to the presence of subjectivity-bearing words in a sentence~\citep{riloff2003learning, kim2005automatic, he2022meta, bao2021bert}. However, subjectivity is context- and domain-dependent. Some words are only subjective in certain contexts or domains. Therefore, many researchers incorporated syntactic dependencies~\citep{wilson2004just, xuan2012linguistic}, interactions between neighboring sentences~\citep{wiebe1994tracking, pang2004sentimental} or in discourse~\citep{biyani2014using} to extract different levels of contextual information. An alternative to this subjective-lexicon-based approach is the word-frequency-based approach~\citep{rustamov2013sentence, kamil2018adaptive}, which is completely domain-independent by learning from document-level information. However, this approach has difficulties capturing syntactic dependencies. By now, subjectivity detection research has been divided into several distinct tasks, each with its unique objectives. One such task is individual subjectivity detection, which focuses on detecting subjectivity at the sentence level. In contrast, context-dependent subjectivity detection aims to incorporate discourse information and a broader context in detecting subjectivity. Cross-lingual subjectivity detection, on the other hand, strives to identify subjectivity in various languages. Moreover, multi-modal subjectivity detection is concerned with identifying subjective expressions in different modalities such as audio and video. Finally, the bias detection task is centered on identifying biased statements in ostensibly impartial articles.

Subjectivity detection is commonly considered as a sub-task of sentiment analysis since it serves as a filtering step for polarity detection~\citet{liu2010sentiment}. It can also be helpful for downstream tasks that require a distinction between opinionated and non-opinionated sentences, such as opinion and information retrieval~\citep{zhang2007opinion, wiebe2011finding}, analyses in financial and political domains~\citep{wang2021signaling, tang2014learning, al2022sentence}, question answering systems~\citep{yu2003towards, li2008cocqa}, etc.

\subsection{Theoretical Research}\label{sec:theoretical_research_sd}

Given the broadness of subjective expressions, e.g., expressing personal feelings, evaluations, and speculations, the related theoretical research in this domain is also rich. 

\subsubsection{Subjective Elements}

Early linguistic works studied subjective language extensively in third-person narrative text. \citet{banfield2014unspeakable} defined the \textit{SELF} of a sentence as the speaker in conversation or the narrating character in third-person fictional text. She identified a variety of morphological, lexical, and syntactic elements, termed subjective elements, that always express the private states, i.e., emotions and opinions, of the sentence's \textit{SELF}. However, many linguistic elements are subjective only in certain conditions. Therefore, following~\citet{banfield2014unspeakable},~\citet{wiebe1990recognizing} further defined a category termed potential subjective elements, which expanded the subjective elements with some linguistic elements that can, but not always, report the private state of a character. \citet{wiebe2000learning} applied these findings to identify the subjective language in the non-fictional text, suggesting that potential subjective elements are also valid subjectivity clues for texts other than third-person narrative fiction.

\subsubsection{Speech Acts}

Speech acts have a strong connection with subjective expressions because speech acts perform actions, such as making a promise, giving an order, or expressing a belief. \citet{austin1975things} argued that language is not just a tool for describing the world but also a means of accomplishing things in the world. Through speech acts, individuals can influence the world around them and the actions of others. In this sense, many seemingly objective expressions with speech acts can become subjective. For example, if someone says,
\ex. I promise to do it.

The utterance is not just conveying information but also performing the act of making a promise. A more subjective case is 
\ex. I believe that it will rain tomorrow.

When individuals express belief in such a manner, they are essentially asserting their mental disposition or perspective towards a specific statement. This entails making a claim about their inner state or outlook toward a proposition. \citet{austin1975things} argues that a considerable number of utterances possess illocutionary force, which signifies that their purpose is not merely to communicate information but also to accomplish something beyond that. Thus, subjective expressions may be more than we think in our everyday language.

\subsubsection{Conceptual Metaphor}\label{sect: conceptual metaphor}

\citet{lakoff1980metaphors} argued that metaphors are not solely a linguistic phenomenon, but also mirror human cognition via concept mappings. When an individual uses a metaphorical expression, they employ a source concept to represent a target concept in a particular context, thereby conveying their cognitive attitude toward the target concept. This process, known as concept mappings, facilitates such representation. In instances such as the statement 
\ex. Our love is a journey. \label{eg: love is journey}

The individual utilizes the concept of a ``journey'' as the source to represent the target concept of ``love'', expressing their subjective feeling that their love is characterized by both ups (joy) and downs (sadness). ``Our love is a journey'' cannot be an objective statement, because the two concepts are from different domains, i.e., literally, love is not a journal. Thus, there is a semantic contrast between the literal and contextual meanings of a metaphor~\citep{mao2019end}. The semantic disparities inherent in metaphors suggest that relying on the literal meanings of a statement alone is insufficient in substantiating its subjectivity. Even though the statement of Example~\ref{eg: love is journey} does not use any obvious opinionated words, e.g., ``happy'' and ``sad'', it also expresses a personal feeling. Thus, the pragmatics of statements must also be taken into account in subjective detection.

\subsection{Annotation Schemes}\label{sec:annotation_schemes_sd}

For general subjectivity detection, it is sufficient for a dataset to annotate a sentence, snippet, or document as subjective (positive/negative) or objective (neutral). Nevertheless,~\citet{wiebe2005annotating} proposed the MPQA scheme, which annotates text at the word and phrase levels. The MPQA scheme is suitable for fine-grained subjectivity detection that aims to identify the source, target, and properties of each expression of the private state.

\citet{wilson2008annotating} proposed the AMIDA Scheme for annotating subjectivity in speech. This scheme marks word spans that are in the following three main categories: subjective utterances, objective polar utterances, and subjective questions. A subjective utterance is a word span that expresses a private state. An objective polar utterance delivers positive or negative factual information without expressing a private state. A subjective question is a question in which the speaker is eliciting the private state of someone else. Each category is divided into finer classes that indicate the polarity and certainty of an utterance.

\subsection{Datasets}\label{sec:datasets_sd}

\begin{table}[!htbp]
\centering
\scriptsize
\begin{tabular}{lllrl}
\toprule
Dataset & Task & Source  & \# Samples & Reference \\
\hline
MPQA &  ISD, CDSD & English news articles  & 9,700 &~\citet{wiebe2005annotating} \\
MPQA Gold  &  ISD, CLSD & Spanish sentences  & 504 &~\citet{mihalcea2007learning}  \\
Multi-MPQA &  ISD, CLSD & Machine-translated MPQA  & 9,700 &~\citet{banea2010multilingual}  \\
Movie  &  ISD, CDSD & Rotten Tomatoes, IMDB & 10,000  &~\citet{pang2004sentimental} \\
WebDoc  & CDSD  & English web documents & 1,076  &~\citet{chesley2006using} \\
TREC  & CDSD  & WSJ & 2,000  &~\citet{yu2003towards}  \\
Debate  & CDSD  & \begin{tabular}[c]{@{}l@{}}Political and ideological\\dataset\end{tabular} & 53,453  &~\citet{al2022sentence} \\
Twitter1  &  ISD & English tweets & 200,000 &~\citet{barbosa2010robust}  \\
Twitter2  &  ISD & English tweets & 498 &~\citet{serrano2015sentiment}  \\
Forum  &  CDSD & online forums & 700 &~\citet{biyani2014using}  \\
SemEval 2013  &  ISD & English tweets & 12,002 &~\citet{nakov2013semeval} \\
NET &  ISD & \begin{tabular}[c]{@{}l@{}}English nuclear energy\\tweets\end{tabular} & 2,308  &~\citet{khatua2020deciphering} \\
MLT &  ISD, CLSD & \begin{tabular}[c]{@{}l@{}}Multi-lingual nuclear\\energy tweets\end{tabular} & 7,700  &~\citet{satapathy2017subjectivity} \\
TASS &  ISD, CLSD & Spanish tweets & 10,000  &~\citet{villena2015overview} \\
Email & CDSD  & BC3 corpus & 1,800  &~\citet{murray2011subjectivity} \\
AMIDA &  MMSD &  AMI Meeting Corpus  & 13  &~\citet{wilson2008annotating}  \\
ICT-MMMO & MMSD  & Youtube review videos  & 370  &~\citet{martin2013youtube}  \\
MOUD  &  MMSD & Youtube review videos  & 498  &~\citet{morency2011towards} \\
Conservapedia &  BD & Conservapedia statements & 1,000  &~\citet{hube2018detecting} \\
WNC &  BD & Wikipedia sentence pairs & 180,000  &~\citet{pryzant2020automatically} \\
\bottomrule
\end{tabular}
\caption{Subjectivity detection datasets and statistics. ISD denotes individual subjectivity detection. CDSD denotes context-dependent subjectivity detection. CLSD denotes cross-lingual subjectivity detection. MMSD denotes multi-modal subjectivity detection. BD denotes bias detection.}\label{tab:sd-datasets}
\end{table}

A summary of all the introduced datasets can be found in Table \ref{tab:sd-datasets}. Generally, subjectivity detection data are organized in the following forms. A text is typically labeled as either subjective or objective, with the former category often further classified as positive, negative, or neutral. The following examples are from SemEval-2013 Task 2B: Sentiment Analysis on Twitter~\citep{nakov2013semeval}. 
\begin{mdframed}
\noindent\texttt{\scriptsize{{id1: "264215390773727232"\\
id2: "276151090"\\
text: "Alex Poythress had 11 points and 7 rebounds in his debut with Kentucky during an exhibition game on Thursday. He played 28 minutes."\\
label: objective}}}
\\

\noindent\texttt{\scriptsize{{id1: "263732569508552704"\\
id2: "369152026"\\
text: "Kick-off your weekend with service! EV!'s Get on the Bus trip to the Boys \&amp; Girls Club is Friday from 3-6! Hope to see you there :)"\\
label: "positive"}}}
\\

\noindent\texttt{\scriptsize{{id1: "213342054351257601"\\
id2: "189656827"\\
text: "Desperation Day (February 13th) the most well known day in all mens life."\\
label: negative}}}
\\

\noindent\texttt{\scriptsize{{id1: "263803288074477568"\\
id2: "396953010"\\
text: "It seem like Austin Rivers is tryin to had to get a bucket. I feel em tho my 1st game in the league I was trying hard too"\\
label: neutral}}}
\end{mdframed}

For fine-grained subjective annotation, the labels are annotated at the span level. The following examples are from SemEval-2013 Task 2A: Sentiment Analysis on Twitter~\citep{nakov2013semeval}. 
\begin{mdframed}
\noindent\texttt{\scriptsize{{id1: "255732290246815744"\\
id2: "315400337"\\
text: "Billy Cundiff may be leaving Washington.  Hopefully he won't miss the door on the way out."\\
start id: "7"\\
end id: "7"\\
label: "positive"}}}
\\

\noindent\texttt{\scriptsize{{id1: "255732290246815744"\\
id2: "315400337"\\
text: "Billy Cundiff may be leaving Washington.  Hopefully he won't miss the door on the way out."\\
start id: "9"\\
end id: "10"\\
label: "positive"}}}
\end{mdframed}

MultiParty Question Answering (MPQA)~\citep{wiebe2005annotating} is derived from 535 English news articles from a wide variety of news sources, manually annotated for subjectivity. The corpus contains 9,700 sentences, 55\% of which are labeled as subjective and 45\% as objective. The MPQA Gold~\citep{mihalcea2007learning} contains 504 Spanish sentences manually annotated for subjectivity, where 273 sentences are subjective and 231 are objective. The Multi-MPQA~\citep{banea2010multilingual} contains parallel corpora to the MPQA dataset in five languages other than English, namely, Arabic, French, German, Romanian, and Spanish.

The Movie Review dataset (Movie)~\citep{pang2004sentimental} contains 5,000 movie review snippets collected from Rotten Tomatoes\footnote{\url{https://www.rottentomatoes.com}}, considered as subjective. Furthermore, 5,000 sentences are collected from plot summaries from the Internet Movie Database (IMDB)\footnote{\url{https://www.imdb.com}}, considered as objective. All reviews and plot summaries are sourced from movies released post-2001, preventing overlap with the polarity benchmark dataset~\citep{pang2004sentimental}. A data sample, either sentence or snippet, is at least 10 words long. The Debate dataset~\citep{al2022sentence} is derived from the political and ideological dataset~\citep{somasundaran2010recognizing}, containing 53,453 sentences from political and ideological posts and comments. The instances are automatically labeled for subjectivity by using lexicon-based and syntactic-pattern-based classifiers~\citep{riloff2003learning}.

Numerous microtext corpora exist that can serve as benchmark datasets for subjectivity detection. \citet{barbosa2010robust} presented a dataset containing 200,000 English tweets, where roughly 100,000 are subjective and the rest are objective. \citet{serrano2015sentiment} manually annotated 498 English tweets as positive, negative, or neural. SemEval 2013~\citep{nakov2013semeval} is a collection of 12,002 English tweets labeled as objective, positive, negative, or neutral. Nuclear Energy Tweets (NET)~\citep{khatua2020deciphering} contains 2,308 English tweets about nuclear energy, manually annotated for subjectivity. The Multilingual Tweets (MLT) dataset~\citep{satapathy2017subjectivity} is a collection of 12,719 tweets about nuclear energy in English, French, Spanish German, Malay, and Indonesian, 7,700 out of which are manually labeled for subjectivity. The Taller de Analisis de Sentimientos en la SEPLN (TASS) corpus~\citet{villena2015overview} contains 10,000 tweets in Spanish, collected from posts by 150 public figures in fields of sports, politics, and communication during the period from 2011 to 2012. Each tweet is labeled as positive, neutral, negative, or without opinion.

The Web Document dataset~\citep{chesley2006using} contains 1,076 English web documents, sourced from traditional news websites and blog posts on diverse topics. Each document is manually annotated as objective, positive, or negative. The Text REtrieval Conference (TREC) dataset~\citep{yu2003towards} is a collection of 8,000 WSJ articles evenly distributed in the categories of editorial, letter to editor, business, and news. The articles and sentences from the former two categories are mapped as opinions (subjective), while the ones from the latter two are facts (objective). The Forum dataset~\citep{biyani2014using} contains 700 threads from online forums Trip Advisor–New York\footnote{\url{http://www.tripadvisor.com/ShowForum-g60763-i5- New_York_City_New_York.html}} and Ubuntu Forums\footnote{\url{http://ubuntuforums.org}}, manually annotated for subjectivity. Email~\citep{murray2011subjectivity} contains 1,800 sentences derived from BC3 corpus~\citep{ulrich2008publicly}, 172 out of which are labeled as subjective. 

For multi-modal subjectivity detection, the AMIDA dataset~\citep{wilson2008annotating} consists of 19,071 dialogue act segments from 20 conversations from the AMI Meeting Corpus~\citep{mccowan2005ami}, manually annotated with the AMIDA scheme. 42\% of the dialogue act segments are tagged with at least one subjective annotation. The Institute for Creative Technologies Multi-Modal Movie Opinion (ICT-MMMO) dataset~\citep{martin2013youtube} contains 370 Youtube review videos labeled as strongly negative, weakly negative, neutral, weakly positive, and strongly positive. Multimodal Opinion Utterances Dataset (MOUD)~\citep{morency2011towards} is a collection of 80 Youtube review videos annotated as positive, negative, and neutral.

For the bias detection task, which aims to identify subjective bias in Wikipedia, the following datasets are widely used. Conservapedia~\citep{hube2018detecting} is a collection of 1,000 single-sentence statements from Conservapedia\footnote{\url{http://www.conservapedia.com}}, manually annotated as biased or unbiased. Wiki Neutrality Corpus (WNC)~\citet{pryzant2020automatically} contains 180,000 aligned Wikipedia sentence pairs. Each pair consists of a sentence before and after bias neutralization by English Wikipedia editors.

\subsection{Knowledge Bases}\label{sec:knowledgebases_sd}

Lexicons of subjectivity clues and patterns are commonly used for subjectivity detection, as summarized in Table \ref{tab:sd-knowledge}. The General Inquirer~\citep{stone1966general} is a lexicon consisting of 10,000 words sorted into 180 categories for content analysis. The Subjectivity Clues lexicon~\citep{riloff2003learning} is a list of words that are subjective in most cases (strongly subjective) and words that may have subjective use in certain contexts (weakly subjective). MPQA Subjectivity Lexicon~\citep{wilson2005recognizing} expanded the Subjectivity Clues using additional dictionaries and lexicons, containing over 8,000 subjectivity clues. 

Knowledge bases that provide sentiment information are also widely used for subjectivity detection. WordNet-Affect~\citep{strapparava2004wordnet} is a set of synsets derived from WordNet that effectively represents affective concepts. SentiWordNet, as introduced in the previous section, is based on WordNet. Each word in SentiWordNet is given three scores indicating its positivity, negativity, and objectivity. SenticNet~\citep{cambria2022senticnet} is a concept-level knowledge base that includes semantic, sentic, and polarity associations.

\begin{table}[!htbp]
\centering
\scriptsize
\begin{tabular}{llrl}
\toprule
Name & Knowledge  & \# Entities & Structure  \\
\hline
The General Inquirer & Sentiment labels & 4,000 & List  \\
MPQA Subjectivity Lexicon & Subjectivity clues & 8,000 & List  \\ 
SentiWordNet & \begin{tabular}[c]{@{}l@{}}Structured lexical\\knowledge by concept\end{tabular} &  100,000 & Graph  \\
WordNet-Affect & lexical knowledge &  4,787 & Graph \\
SenticNet & Sentiment scores  & 200,000 & Graph \\
\bottomrule
\end{tabular}
\caption{Useful knowledge bases for subjectivity detection.}\label{tab:sd-knowledge}
\end{table}

\subsection{Evaluation Metrics}\label{sec:evaluation_metrics_sd}

The performance of subjectivity detection is commonly evaluated via accuracy and F-measure.

\subsection{Annotation Tools}\label{sec:tools_sd}

The aforementioned NER annotation tools (see Section~\ref{sec:annotation_tools_ner}) can be used for subjectivity detection because these tools can annotate labels for spans (fine-grained subjectivity detection) and sentences (coarse-grained subjectivity detection).

\subsection{Methods}\label{sec:methods_sd}

\subsubsection{Individual Subjectivity Detection}\label{sec:task1_sd}

In individual subjectivity detection, the subjectivity of a sentence is evaluated in isolation and irrespective of any contextual factors. The primary methods used for addressing this task include lexicon-based, word frequency, and deep learning approaches.

\noindent \textbf{A. Lexicon-based}

Drawing on the premise that sentences that contain commonly-subjective expressions are more likely to be subjective, lexicon-based methods utilize a manually-constructed lexicon of subjective words, clues, or patterns to determine the subjectivity of a given sentence.

\citet{riloff2003learning} introduced an unsupervised rule-based classifier that leverages the identification of subjective clues and patterns to detect subjective sentences, while also employing bootstrapping to recognize objective sentences based on the absence of such indicators. The clues were manually collected and annotated. The patterns were generated by the AutoSlog-TS algorithm~\citep{riloff1996automatically}, based on pre-defined syntactic templates. \citet{wiebe2005creating} further improved this bootstrapping system by using the labeled sentence produced by the rule-based method as initial training data for a Na\"ive Bayes classifier. The major weakness of these methods is the unreliable assumption that the absence of subjective clues and patterns indicates objectivity, resulting in false-positive errors.

\citet{kim2005automatic} first compiled lists of words that convey opinions and those that do not, which were manually annotated with corresponding classes and levels of strength. They expanded the lists with a common English word list by measuring the WordNet distance between a common word and the compiled seed lists. They further identified additional opinion words and non-opinion words from editorial and non-editorial WSJ documents by computing their relative frequencies. By detecting the subjectivity of a given sentence based on the presence of a single strong valence word, their method achieved 65\% accuracy on MPQA.

\citet{benamara2011towards} argued that sentence-level subjectivity detection cannot fully leverage context, because a sentence may contain several opinion clauses, and opinion expressions may be discursively related. As such, they proposed a segment-level annotation based on the Segmented Discourse Representation Theory~\citep{asher2003logics}, where segments are labeled as explicitly subjective, implicitly subjective, subjective non-evaluative, and objective. This fine-grained annotation can better enhance polarity detection, as segments in the latter two categories do not covey positive, negative, or opinion. However, the limitation of this method is that the four label classes are unbalanced in the corpus. Additionally, implicitly subjective segments are often nuanced and hard to identify. Thus, it would be challenging to design an appropriate classifier. The paper circumvented this problem by reframing the task as two parallel binary classification tasks and obtained 82.31\% accuracy with a manually compiled French lexicon and SVMs as classifiers.

Merely detecting the existence of subjective keywords is often an insufficient indication of a sentence's subjectivity. Other works attempted to enrich the feature set by incorporating more sentence-level information. Relying on expert knowledge of parse tree,~\citet{xuan2012linguistic} manually constructed a set of syntax-based patterns from unigrams and bigrams to extract features. A MaxEnt model was employed as the classifier, obtaining 92.1\% accuracy on the Movie dataset. \citet{remus2011improving} hypothesized that the readability of a sentence was related to its subjectivity. Hence, readability formulae such as Devereux Readability Index~\citep{smith1961devereux} and Easy Listening~\citep{fang1966easy} were incorporated as features in addition to the MPQA Subjectivity Lexicon, obtaining 84.5\% F-measure on Moive.

Compared to standard text, microtext such as tweets contains informal and irregular expressions, making it more difficult for machines to process. Many works proposed subjectivity detection systems that specifically targeted Twitter text. Given the word constraint imposed by Twitter, a tweet is generally regarded as a sentence. \citet{barbosa2010robust} believed that using subjectivity detection as an upstream task would improve the performance of polarity detection on Twitter text. Aside from conventional features such as subjective clues and PoS tags, they leveraged Tweet-specific syntax features, e.g., links and upper case. An SVM classifier was employed, which achieved 81.9\% accuracy on the Twitter dataset, and improved the accuracy of polarity detection by 5.6\%. Following their footsteps,~\citet{sixto2016approach} incorporated more Tweet-specific features that leveraged the structure of Twitter, e.g., the relationship between tweets, users, hashtags, and links. Using the stacking classifier proposed by~\citet{cotelo2015explorando}, their method obtained 89.8\% on TASS. To reduce human effort,~\citet{keshavarz2018mhsublex} created a Twitter subjectivity lexicon automatically through a meta-heuristic approach, i.e., a genetic algorithm, which produced separate lists of subjective and objective words. A Bayse network was employed to classify a given tweet based on its subjective and objective word counts, achieving 60.9\% on SemEval 2013. Alternatively,~\citet{khatua2020deciphering} leveraged the concept-level knowledge base SenticNet as their lexicon, which is able to provide implicit meaning associated with commonsense concepts. Their method obtained 80.7\% accuracy on the NET dataset.

The methods introduced above have a common limitation, i.e., the lexicons are lists of keywords, instead of word meanings. Some subjective clues in fact have both subjective and objective word senses, which are not distinguishable in keyword lexicons, leading to false-positive errors. This problem can be mitigated by incorporating a Subjectivity WSD (SWSD) system to build a sense-aware lexicon. \citet{akkaya2009subjectivity} trained a supervised targeted SWSD system using SVM. The training data was compiled using words that are both in the MPQA Subjectivity Lexicon and the sense-tagged SENSEVAL corpora~\citep{kilgarriff2000introduction, preiss2001proceedings, litkowski2004senseval}. Alternatively,~\citet{ortega2013improving} applied an unsupervised, clustering-based SWSD system~\citep{anaya2006word} on SentiWordNet to label each subjective word with fine-grained sense. Both SWSD systems were applied to a rule-based classifier similar to the one proposed by~\citet{riloff2003learning}. The supervised one improved accuracy by 1.3\% on MPQA, while the unsupervised one improved F-measure by 6.5\% on Movie. A prominent limitation of lexicon-based methods is that they require external resources such as sentiment lexicon and knowledge base.

\noindent \textbf{B. Word Frequency}

Word-frequency-based methods detect subjectivity by modeling word presence or occurrence within a corpus. Therefore, compared to lexicon-based methods, they are language-independent and require neither manual annotation nor linguistic knowledge. They are also less computationally expensive due to the reduction of feature sets.

\citet{rustamov2013sentence, kamil2018adaptive} proposed a language-independent feature extraction algorithm with a novel statistical measure of word occurrence called Pruned ICF (Inverse-Class Frequency), which is proven to be more effective than the standard IDF (Inverse-Document Frequency). Additionally, they applied two widely-used methods for pattern recognition to detect subjectivity, namely Fuzzy Control System (FCS)~\citep{helmi2009human} and Adaptive Nero-Fuzzy Inference System (ANFIS)~\citep{fuller1995neural}, achieving the accuracy of 91.3\% and 91.66\% on the Movie dataset, respectively. The latter obtained better performance due to the addition of a neural network layer. Inspired by empirical evidence that hybrid systems improve the performance of NLP classifiers,~\citet{rustamov2018hybrid} further integrated FCS, ANFIS, and HMM into a sequential hybrid system, where input sentences that are wrongly labeled by the prior classifier are passed onto the subsequent one. Using the same feature extraction method as the previous paper, this system increased the accuracy to 92.24\% on Movie.

\citet{wang2013fast} proposed a novel dropout algorithm to optimize the feature learning process. Conventional dropout training in neural network~\citep{hinton2012improving} prevents feature co-adaptation by randomly sampling neurons and input features and setting them to zeros, which leads to slow training. The authors suggested fast dropout training as a more efficient alternative, using a Gaussian approximation to draw samples. They applied this dropout method to Na\"ive Bayes Support Vector Machine (NBSVM)~\citep{wang2012baselines}, which extracts features based on word presence. Their method not only achieved the accuracy of 93.6\% on Movie and 86.3\% on MPQA, but also greatly decreased the training time. Experiments also showed that fast dropout training could be applied to other loss functions and neural networks.

Latent Dirichlet Allocation (LDA)~\citet{blei2003latent} is a weakly-supervised generative model that assumes every document is a distribution of latent topics, which is determined by word frequencies. \citet{he2010bayesian, maas2011learning} suggested that subjectivity detection can be solved by LDA, based on the intuition that subjective sentences likely contain opinionated words. Hence, the paper modified conventional LDA so that the latent topics are word-level sentiment labels. An additional layer is inserted between word and document levels to model sentence-level subjectivity labels. Sentiment lexicons are incorporated to establish an informed prior distribution for word-level sentiment labels, achieving 71.2\% accuracy on MPQA. On the other hand,~\citet{lin2011sentence} argued that LDA likely discovers topics based on semantic similarities, instead of sentiment. Therefore, they modified LDA, so that it directly models word probabilities conditioned on topic distributions to capture semantic information. To explicitly extract sentiment information, they incorporated supervised sentiment analysis as an auxiliary task. Their method achieved 88.58\% on the Movie dataset.

The drawback of the word frequency approach is that the order of the words is not considered. Thus, syntactic information cannot be effectively learned using this approach.

\noindent \textbf{C. Deep Learning}

The acquisition of precise sentence representations is crucial for subjectivity detection, and as such, numerous studies have examined neural sentence modeling as a language-independent alternative to parse trees. \citet{kalchbrenner2014convolutional} presented a Dynamic CNN (DCNN) that is able to capture short- and long-range relations. The core component of DCNN is dynamic pooling, which outputs the sub-sequence of $k$ maximum values in the input sequence, where $k$ can be dynamically chosen. Hence, DCNN produces a hierarchical feature graph that contains syntactic, semantic, and structural patterns of the input sentence. However, their sentence representations do not retain any intermediate information, e.g., word-level and phrase-level features. To address this,~\citet{zhao2015self} described a self-adaptive hierarchical sentence model named AdaSent. Inspired by gated recursive CNN~\citep{cho2014properties}, AdaSent forms a pyramid-shape directed acyclic graph, where the bottom level is word representations and the top level is sentence representations. In this process, the gating network receives information from each level and selects the most appropriate representations for the given task. Their method obtained an accuracy of 95.5\% on Movie and 93.3\% MPQA.

With similar motivation for higher-order dependencies,~\citet{chaturvedi2018bayesian} proposed a Bayesian Network-based Extreme Learning Machine (BNELM) framework for subjectivity detection. Single-layer feedforward neural networks, known as Extreme Learning Machines (ELMs), excel at inductive learning. However, the excessive number of hidden neurons in ELMs often leads to overfitting and slow performance. To address these weaknesses, Bayesian networks were introduced to model connections among the hidden neurons of ELM, as they can prune redundant and irrelevant hidden neurons and capture high-dimensional features. Furthermore, ELM cannot handle non-linear data such as sequences of sentences. Thus, an RNN layer was used to extract temporal features. Upon it, a fuzzy classifier was applied to achieve stability in case of noisy data, producing the output labels. Additionally, a deep CNN was employed prior to BNELM to provide low-dimensional features. The framework achieved the accuracy of 75\% on MPQA Gold and 89\% on TASS, outperforming previous ELM-centric models, namely, standard ELM and Sparse Bayesian ELM~\citep{soria2011belm}.

Likewise,~\citet{satapathy2017subjectivity} employed CNN and RNN to extract spatial and temporal information respectively. To make the model more robust, they incorporated reinforcement learning, namely Point-wise Probability Reinforcement (PPR)~\citep{frenay2015reinforced}, to regularize the learning process of CNN and reduce the influence of outliers. Specifically, convolutional layers in the CNN component were added iteratively, where the weight of each neuron was fine-tuned by the reinforced maximum likelihood of PPR. Their method did not perform very well on MPQA, obtaining 50\% F-measure. However, it achieved a good performance of 76\% F-measure on the multi-lingual Twitter dataset MLT.

In the same vein, PLMs can also provide beneficial universal representations learned from a plethora of unlabeled text. For instance,~\citet{al2022sentence} fed GloVe embeddings to different types of RNN variants, among which LSTM with attention mechanism achieved the best accuracy of 89.53\% on MPQA and 83.83\% on their proposed political and ideological dataset, whereas Bi-LSTM with attention achieved the best accuracy of 92.8\% on Movie. \citet{kim2014convolutional} fine-tuned pre-trained Word2Vec with a simple CNN, obtaining accuracy of 93.4\% on Movie and 89.5\% on MPQA. 

Furthermore, many works observed that it is complementary to combine PLM and MTL for more effective learning of text representations~\citep{liu2019multi, sun2019fine, mao2021bridging}. Motivated by this,~\citet{huo2020utilizing} fine-tuned BERT using MTL, where the BERT layers are shared among subjectivity detection and three other text classification tasks. Similarly,~\citet{satapathy2022polarity} proposed an MTL framework for subjectivity and polarity detection. The framework leverages BERT as embedding, which is fed into two separate self-attention Bi-LSTM layers. A neural Tensor Network (NTN)~\citep{socher2013reasoning} was used as the information-sharing layer. Both methods employed a simple softmax classifier for each task. The former achieved 95.23\% accuracy on Movie, while the latter obtained 95.1\%. However, a shared limitation is that, despite their overall good performance, some of the tasks did not exceed single-task learning baselines. This is likely because both methods adopted hard parameter sharing MTL~\citep{crawshaw2020multi}, which emphasizes more on generalization rather than optimization.

\citet{sagnika2021attention} presented an attention-based CNN-LSTM model for subjectivity detection, which served as a pre-processing step for sentiment analysis. The combination of CNN and LSTM enabled the model to capture both spatial and temporal information. Additionally, it utilized word embeddings enhanced by sentiment-related information~\citep{sagnika2020improved}. Initially, the training of the model was carried out with the Movie dataset, after which it was utilized to analyze the sentiment of the IMDb dataset. The objective sentences were eliminated from the dataset to form a modified set of reviews. Various models were tested as sentiment classifiers. The subjectivity detection model not only obtained 97.1\% accuracy on the Movie dataset, but also consistently improved the performance of sentiment analysis.

\subsubsection{Context-Dependent Subjectivity Detection}\label{sec:task2_sd}

The method of individual detection categorizes each sentence without considering its context. However, subjectivity detection and sentiment classification are contextual problems since lexical items can affect each other in a discourse setting~\citep{aue2005customizing, polanyi2006contextual}. \citet{pang2004sentimental} was the first to leverage inter-sentence context information to filter out objective sentences, in order to better serve document-level polarity detection. Based on the hypothesis that adjacent text spans might have the same subjectivity label~\citep{wiebe1994tracking}, suggested an algorithm known as the ``minimum cuts algorithm'' that aims to optimize the subjectivity status score for every sentence separately, while also punishing the assignment of different labels to two closely related sentences. These two sub-objectives are independent of each other, making the model more flexible for the addition of features.

Context-dependent methods can be divided into two categories, namely, the feature engineering approach and the statistical approach.

\noindent \textbf{A. Feature Engineering}

A common way to incorporate document-level information is by designing relevant features.
\citet{das2009theme} proposed a domain-independent rule-based algorithm, named theme detection. The model utilized document-level features, e.g., positional aspects (document title, first paragraph, last two sentences), the positions of subjectivity clues, and the distance between any two thematic words. As with many techniques at the sentence level, this approach also integrated syntactic characteristics and resources such as SentiWordNet and MPQA Subjectivity Lexicon. It achieved precision and recall of 76.08\% and 83.33\% on MPQA.

To automatically select an appropriate feature set,~\citet{das2010subjectivity} employed the genetic algorithm (GA)~\citep{holland1992adaptation, sastry2005genetic}, which is a probabilistic search method, to find the optimal range of values of every feature. To capture context information, positional aspects, word distribution, and document theme~\citep{wiebe2000learning} were incorporated as discourse-level features, aside from the commonly-used lexical and syntactic features. The GA then identified the globally optimal feature set by natural selection and computed the corresponding accuracy of the classifier through the fitness function. An advantage of the proposed method over other statistical classifiers is that the entire input sentence is encoded by GA and used as features, instead of using $n$-gram. Their method obtained the F-measure of 93.02\% on MPQA and 95.69\% on Movie.

\citet{biyani2014using} noticed a gap in subjectivity detection targeting online forums. Moreover, they argued that lexical features are highly dimensional, leading to the risks of overfitting and slow training. Thus, they presented a Forum dataset, and designed a set of non-lexical thread-specific features. Specifically, they leveraged thread structure and dialog acts and utilized lexicons and tools such as MPQA Subjectivity Lexicon and SentiStrength~\citep{thelwall2012sentiment} to extract sentiment features. With the addition of conventional lexical features, the logistic regression classifier obtained 77.01\% accuracy on Forum.

\noindent \textbf{B. Statistical Approach}

To minimize human effort in designing features, a statistical approach automatically learns features from a given corpus using statistical models. \citet{yu2003towards} simply implemented a Na\"ive Bayes classifier for document-level subjectivity detection, which achieved the F-measure of 97\% on the TREC dataset proposed by them. 

Motivated by the observation that language models are adept at representing knowledge of the text they were trained on,~\citet{karimi2017language} proposed a language-model-based document-level subjectivity detection method. During training, a subjective reference language model and an objective were built using labeled documents. During inference, a language model was constructed for each input document, which was compared with the reference language models using KL-divergence~\citep{lafferty2001document}, producing two similarity scores. The difference between these two scores was regarded as the subjectivity score of the document. The final output of the model was a sorted list of input documents, based on their subjectivity scores. To achieve language non-specificity, the paper also proposed a semi-supervised method where the reference language models were built on a lexicon divided into subjective and objective parts, based on polarity scores. The supervised method obtained 94.63\% MAP on the Movie dataset, whereas the unsupervised obtained 53.61\% MAP.

Word embeddings can only provide limited syntactic and semantic information~\citep{belinkov2018evaluating}. Therefore, to better initialize their model,~\citet{chaturvedi2016bayesian} employed a Gaussian Bayesian Network (GBN)~\citet{friedman1998learning} layer to capture long-range features among successive sentences, which were used to pre-train the CNN classifier. The GBN layer converted the sentence sequence from the MPQA dataset into a time series of word frequency, captured second-order word dependencies with a time lag of 2, and generated a subset of sentences that contained the most significant words from the MPQA Subjectivity Lexicon. The model adopted a CNN sentence model with convolution kernels of increasing size, which combined the local word dependencies within the kernel size to model long-range syntactic relations. It was pre-trained with the sub-set of sentences produced by GBN before being trained on the full dataset, obtaining the accuracy of 93.2\% on MPQA and 96.4\% on Movie.

\subsubsection{Cross-Lingual Subjectivity Detection}\label{sec:task3_sd}

\noindent \textbf{A. Language-Independent Approach}

For feature-engineering-based subjectivity detection, lexical resources and tools are often not readily available for non-English languages. A common approach to circumvent this problem is to use non-language-specific features that are based on the presence or occurrence statistics of a corpus, e.g., word frequency~\citep{rustamov2013sentence, wang2013fast, kamil2018adaptive, blei2003latent, lin2011sentence, belinkov2018evaluating} and language modeling~\citep{karimi2017language}. \citet{mogadala2012language} further introduced language-independent feature weighing, leveraging unigram and bigram frequencies, and unigram word length. Entropy-based category coverage difference~\citep{largeron2011entropy} was employed as the feature selection method.

\noindent \textbf{B. Translation Approach}

Another solution is the translation approach, where lexical resources for the target language are automatically generated by translating the resources and tools available for English, usually with the help of statistical machine translation (SMT)~\citep{kim2006identifying, mihalcea2007learning, banea2008multilingual, wan2009co, banea2011multilingual, amini2019cross}. \citet{banea2010multilingual} conducted a study on English and five other highly lexicalized languages, proving that a multi-lingual feature space constructed through SMT improved the accuracy of subjectivity detection on all languages involved. However, the sentence translation process can lead to the loss of essential lexical information such as inflection and formality, which often served as an indicator of subjectivity~\citep{banea2008multilingual}.

\citet{chaturvedi2016lyapunov} mitigated this information loss during translation by using a neural network to transfer resources from English to Spanish. They first translated the MPQA Subjectivity Lexicon into Spanish using an SMT system~\citep{lopez2008statistical}. A MaxEnt-based PoS tagger~\citep{toutanvoa2000enriching} and a multi-lingual WSD system~\citep{moro2014multilingual} were incorporated in the preprocessing stage to minimize the loss of lexical information during translation. Their proposed model, named Lyapunov Deep Neural Network (LDNN), extracted spatial features from the input Spanish sentence and its translated English form using CNN, which were then combined with an RNN to capture the bilingual temporal features. To mitigate the vanishing gradient problem with RNN, a Lyapunov function was used as the error function of RNN for stable convergence. Utilizing the high-level features produced by Lyapunov-guided RNN, a multiple kernel learning~\citep{subrahmanya2009sparse, zhang2010adamkl} classifier yielded the prediction. Their model obtained 84.0\% F-measure on MPQA Gold, and 88.4\% accuracy on TASS.

\subsubsection{Multi-Modal Subjectivity Detection}\label{sec:task4_sd}

While most studies on detecting subjectivity have concentrated on text-based data, the identification of subjective expressions in other modalities, such as audio and video, presents an important area for research. For instance,~\citet{murray2009predicting, murray2011subjectivity} proposed an automatic pattern extraction method for subjective expression in spoken conversation, which is able to extract Varying Instantiation N-Grams (VIN) from labeled and unlabeled data. Unlike convention $n$-gram, a VIN is a trigram where each unit can be either a word or a PoS label, which is a more robust alternative to syntactic parsers for fragmented and disfluent text, such as meeting transcripts. Combined with a large raw feature set, a MaxEnt classifier scored the F-measure of 52\% on the AMIDA dataset.

The method above, however, did not leverage any information from other modalities. \citet{raaijmakers2008multimodal} explored the effectiveness of lexical and acoustic features in speech subjectivity detection. Specifically, they investigated word, character, prosody, and phoneme $n$-grams. Following~\citet{wrede2003spotting, banse1996acoustic}, the prosodic features were extracted based on pitch, energy, and the distribution of energy in the long-term averaged spectrum. The word-, character-, and phoneme-level features were extracted from manual speech transcripts. A separate BoosTexter classifier~\citep{schapire2000boostexter} was employed for each feature set, whose predictions were combined using a simple linear interpolation strategy~\citep{raaijmakers2007sentiment} to obtain the final output. The combination of the four types of feature sets achieved 75.4\% accuracy and 67.1\% F-measure on AMIDA. Furthermore, experiments showed that word- and character-level features contributed the most to higher results, whereas prosodic features yielded marginal improvements.

\subsubsection{Bias Detection}\label{sec:task5_sd}

Bias detection refers to the task of identifying biased statements from supposedly impartial articles. Specifically, in Wikipedia, the Neutral Point of View (NPOV) is a core principle that ensures neutrality for controversial topics. Thus, the goal of this task is to detect sentences that violate NPOV policy on a Wikipedia page. Bias detection is closely related to subjectivity detection. Its development mirrors the technical trends of the latter. However, it is considered to be more complex, because the linguistic cues of biased language are often nuanced, and depend heavily on the context.

For lexicon-based approaches,~\citet{recasens2013linguistic} manually compiled a biased word lexicon and feature set that covered framing bias (use of subjective words or phrases that links to a particular point of view), and epistemological bias (linguistic cues that modify the credibility of a statement). However, their method focused only on detecting a single bias-inducing word in a known biased statement. Furthering their work,~\citet{hube2018detecting} constructed a more comprehensive biased word lexicon for sentence-level bias detection. To minimize human efforts, they leveraged Word2Vec to expand a seed word list by measuring the distance between word vectors. Aside from the lexicon, other syntactic and semantic features were incorporated, e.g., tri-gram, PoS tags, Linguistic Inquiry Word Count (LIWC)~\citep{pennebaker2001linguistic}, framing bias features, and epistemological bias features. By using a Random Forest classifier, their method obtained 74\% precision on their proposed Conservapedia dataset.

\citet{aleksandrova2019multilingual} proposed a semi-automatic method to construct a multi-lingual bias detection corpus, consisting of Bulgarian, French, and English sentences from Wikipedia. Their method was applicable for building a corpus from a Wikipedia archive in any language, as it does not rely on language-specific features. Additionally, they provided the performance of three baseline models, namely BoW, fastText~\citep{joulin2017bag}, and logistic regression~\citep{hosmer2000applied}, among which BoW achieved the best overall average F-measure of 59.57\% across the three languages.

For neural network approaches,~\citet{hube2019neural} employed RNN to capture the inter-dependency of words and their context. To address the weakness of RNN in modeling long-range information, a hierarchical attention mechanism~\citep{yang2016hierarchical} was adopted, which applied word-level attention on each sentence to compute sentence representations, upon which sentence-level attention was applied to learn biased cues from different samples. Following previous feature-based works, they concatenated GloVe embedding, PoS tags, and LIWC features as word representations.

PLMs were also widely used in bias detection. 
\citet{pryzant2020automatically} extended the work of~\citet{recasens2013linguistic} by using a pre-trained BERT-based detector to identify bias-inducing words and neutralizing them via an LSTM-based editor. A join embedding mechanism was employed to allow the detector control over the editor. They also introduced the WNC dataset for detecting and editing biased language, on which their model obtained 93.52\% BLEU and 45.80\% accuracy for the produced edits. However, a limitation is that they primarily targeted single-biased words. To mitigate this,~\citet{pant2020towards} enabled multi-word detection by identifying bias at the sentence level. They employed the weighted-average ensemble method on several BERT-based models to detect biased language, which obtained 71.61\% accuracy and 70.40\% F-measure on WNC.

\subsection{Downstream Applications}\label{sec:downstream_applications_sd}

\subsubsection{Sentiment Computing}\label{sec:application1_sd}

The presence of objective texts can dilute the task of sentiment computing. Therefore, the machine can better classify the remaining non-objective opinions by using subjectivity detection as an upstream task\citep{satapathy2017subjectivity, das2020subjectivity}. For document-level sentiment analysis specifically,~\citet{bonzanini2012opinion} showed that subjectivity detection reduced the amount of data to 60\% while still producing the same polarity classification results as full-text classiﬁcation. The analysis reveals that a considerable portion of real-world textual data is objective in nature, and this may cause an imbalance in sentiment analysis and opinion-mining tasks without subjectivity detection.

\citet{pang2004sentimental, das2020subjectivity} applied subjectivity detection to filter out objective sentences in reviews prior to classifying their polarity. Similarly,~\citet{kamal2013subjectivity} first extracted subjective sentences from customer reviews and then employed a rule-based system to mine feature-opinion pairs from the subjective sentences. \citet{barbosa2010robust, soong2019essential} used subjectivity detection in sentiment analysis for Twitter microtext. These works proved that removing objective content from the dataset indeed makes the learning of sentiment more effective.

\subsubsection{Information Retrieval}\label{sec:application2_sd}

Subjectivity detection can serve as a subsystem in an information retrieval system to determine whether a document is subjective or objective~\citep{soong2019essential}, because information retrieval systems normally aim to retrieve either opinionated or factual topic-relevant text from web sources, e.g., tweets, blog posts, reviews webpages, etc. \citep{paltoglou2014opinion}.

For opinion retrieval, it helps to select candidate opinionated documents. For instance,~\citet{zhang2007opinion} first employed an SVM classifier that used unigram and bigram features to identify subjective documents. Then, they separated relevant documents from irrelevant ones. For factual information retrieval, on the other hand, subjectivity detection helps to filter out opinionated text such as allegations and speculations to prevent false hits. \citet{wiebe2011finding} implemented a Na\"ive Bayes subjectivity classifier and a domain-relevant indicator for selective subjective sentence filtering. If a sentence was classified as subjective, it would be discarded unless it was also labeled as relevant by the indicator.

\subsubsection{Hate Speech Detection}\label{sec:application3_sd}

Hate speech detection is a task that identifies abusive speech targeting a person or a group based on stereotypical group characteristics, e.g., ethnicity, religion, or gender, on social media~\citep{warner2012detecting}. Since hate speech is often marked by its content, tone, and target~\citep{cohen2011fighting}, its detection is similar to that of polarity. Additionally, subjectivity clues tend to be surrounding the polarizing and arguing topics, which aligns well with hate speech detection. As such, subjectivity detection can be used as a filtering subsystem in hate speech detection.

For instance,~\citet{gitari2015lexicon} employed a rule-based subjectivity classifier that leveraged lexicons including MPQA Subjectivity Lexicon and SentiWordNet to identify subjective sentences. From the extracted sentences, they built a hate speech lexicon using bootstrapping and WordNet. Experiments showed that the addition of subjectivity detection significantly improved the performance of the hate speech classifier.

\subsubsection{Question Answering System}
QA systems generally encounter two types of questions - the ones that expect truth as answers, and the ones that expect opinions. Therefore, it is crucial for a QA system to distinguish opinions from facts, and provide the appropriate type depending on the question~\citep{yu2003towards}.

To achieve this goal, a QA system should operate in two stages. First, it must determine whether a question calls for a subjective or objective answer, which is its subjectivity orientation~\citep{li2008cocqa, li2008exploring, aikawa2011community}. Then, the system needs to consider subjectivity as a relevant factor in the information retrieval process.

Subjectivity detection can be incorporated as a filter or feature set in a QA system. For instance,~\citet{stoyanov2005multi} modified the conventional QA system by applying a subjectivity filter and an opinion source filter on the initial IR results, which improved the system significantly. On the other hand,~\citet{wan2016modeling} leveraged subjective features from reviews to provide users with a list of relevance-ranked reviews, which improved the performance of answering binary questions from categories with abundant data.

\subsection{Summary}\label{sec:summary_sd}

Subjectivity detection is a cognitive semantic processing task. It categorizes statements by subjective and objective classes. Theoretical research indicates that subjectivity can be detected by certain subjective elements, e.g., morphological, lexical, and syntactic elements~\citep{banfield2014unspeakable}. Thus, computational subjectivity research has developed lexical resources, e.g., Subjectivity Clues~\citep{riloff2003learning}, and MPQA Subjectivity Lexicon~\citep{wilson2005recognizing}. On the other hand, subjectivity can be also explained from the perspectives of pragmatics, e.g., speech acts~\citep{austin1975things} and conceptual metaphors~\citep{lakoff1980metaphors}. Related subjectivity detection works defined the task as classification tasks. Although those classification tasks can be further divided into course-grained and fine-grained classifications, e.g., document-level, sentence-level, and span-level subjectivity detection, there have not been studies aimed at explaining the subjectivity from pragmatic perspectives, e.g., speech acts and metaphors.

The application of subjectivity detection has proven to be supportive in downstream tasks, such as sentiment computing, information retrieval, hate speech detection, and QA systems. This is because these downstream tasks normally aim at mining opinions from subjective expressions. Subjectivity detection can filter out the objective ones, thus yielding the desired input for downstream tasks.

\subsubsection{Technical Trends}\label{sec:summary_technical_sd}

\begin{sidewaystable}[!htbp]
\centering
\scriptsize
\begin{tabular}{llllllll}
\toprule
Task & Reference & Techniques  & Feature and KB & Framework  & Dataset  & Score & Metric  \\ \midrule
\multirow{31}{*}{ISD} &~\citet{riloff2003learning} & Rule & SCSL & Logic Rules  & -  & - & - \\
 &~\citet{kim2005automatic} & Statistics  & MPQA, WN & WN distance & MPQA & 65.00\% & Acc \\
 &~\citet{benamara2011towards} & Statistics  & \begin{tabular}[c]{@{}l@{}}Lexical, stylistic,\\syntactic, discursive \\features\end{tabular}  & SVM & Self-collected  & 82.31\% & Acc \\
 &~\citet{xuan2012linguistic} & Statistics & \begin{tabular}[c]{@{}l@{}}MPQA, syntax-\\based patterns\end{tabular} &  MaxEnt  & Movie & 92.10\% & Acc \\
 &~\citet{remus2011improving} & Statistics & MPQA, readability  & SVM  & Movie  & 84.50\% & F1 \\
 &~\citet{barbosa2010robust} & Statistics & \begin{tabular}[c]{@{}l@{}}MPQA, POS,tweet-\\specific features\end{tabular}  & SVM  & Twitter1  & 81.59\% & Acc \\
 &~\citet{sixto2016approach} & Statistics & \begin{tabular}[c]{@{}l@{}}MPQA, tweet-specific \\features\end{tabular} & Stacking classifier  & TASS  & 89.80\% & Acc \\
 &~\citet{keshavarz2018mhsublex} & Statistics & SCSL & Genetic algorithm  & SemEval2013  & 60.90\% & Acc \\
 &~\citet{khatua2020deciphering} & DL & SenticNet  & CNN  & NET  & 80.70\% & Acc \\
 &~\citet{akkaya2009subjectivity} & Statistics & MPQA, SWSD  & SVM & MPQA  & 81.30\% & Acc \\
 &~\citet{ortega2013improving} & Rule & MPQA, SWSD  & Clustering, logic rules & Movie  &55.68\% & F1 \\
 &~\citet{kamil2018adaptive} & Statistics & Pruned ICF  & ANFIS & Movie & 91.66\% & Acc \\
 &~\citet{rustamov2018hybrid} & Statistics & Pruned ICF  & FCS, ANFIS, HMM & Movie & 92.24\% & Acc \\
 &~\citet{wang2013fast} & Statistics & Word presence  & NBSVM & Movie & 93.60\% & Acc \\
 &~\citet{maas2011learning} & Statistics & \begin{tabular}[c]{@{}l@{}}Semantic and sentiment\\embeddings\end{tabular} & \begin{tabular}[c]{@{}l@{}}Probabilistic model, \\LDA\end{tabular}. & MPQA & 71.20\% & Acc \\
 &~\citet{lin2011sentence} & Statistics & Sentiment  & LDA & Movie & 88.58\% & Acc \\
 &~\citet{zhao2015self} & DL & Word2Vec  & CNN & Movie & 95.50\% & Acc \\
 &~\citet{chaturvedi2018bayesian} & DL & MPQA, POS & \begin{tabular}[c]{@{}l@{}}ELM, RNN, CNN,\\fuzzy classifier\end{tabular} & MPQA Gold & 75.00\% & Acc\\
 &~\citet{satapathy2017subjectivity} & DL & GloVe, MPQA & CNN, PPR & MPQA  & 50.00\% & F1\\
 &~\citet{al2022sentence} & DL & GloVe  &  RNN, Att & Movie & 92.80\% & Acc \\
 &~\citet{kim2014convolutional} & DL & Word2Vec &  CNN & Movie & 93.40\% & Acc \\ 
 &~\citet{huo2020utilizing} & DL & BERT  &  MTL & Movie & 95.23\% & Acc \\
 &~\citet{satapathy2022polarity} & DL & BERT  &  MTL, RNN,NTN & Movie & 95.10\% & Acc \\
 &~\citet{sagnika2020improved} & DL & \begin{tabular}[c]{@{}l@{}}Sentiment-enhanced\\word embedding\end{tabular} & CNN, LSTM & Movie & 97.10\% & Acc \\ \bottomrule
\end{tabular}
\caption{A summary of representative subjectivity detection techniques (Part 1). ISD denotes individual subjectivity detection. SCSL denotes self-collected subjectivity lexicon. SWSD denotes subjectivity WSD.} \label{tab:sd-summary1}
\end{sidewaystable}

\begin{sidewaystable}[!htbp]
\centering
\scriptsize
\begin{tabular}{llllllll}
\toprule
Task & Reference & Techniques  & Feature and KB & Framework  & Dataset  & Score & Metric  \\ \midrule
\multirow{11}{*}{CDSD}  &~\citet{pang2004sentimental} & Statistics  & SCSL & \begin{tabular}[c]{@{}l@{}}Minimum cuts, \\Na\"ive Bayes\end{tabular} & Movie & 86.40\% & Acc \\
 &~\citet{das2009theme} & Rule  & \begin{tabular}[c]{@{}l@{}}MPQA, doc-level \\features, SWN\end{tabular} & Logic rules  & MPQA  & 79.54\% & F1\\
 &~\citet{das2010subjectivity} & Statistics & \begin{tabular}[c]{@{}l@{}}MPQA, POS, doc-\\level features\end{tabular} &  Genetic algorithm & Movie & 95.69\% & F1 \\
&~\citet{biyani2014using} & Statistics  & \begin{tabular}[c]{@{}l@{}}MPQA, SentiStrength,\\thread-specific features\end{tabular}  & Logistic regression & Forum & 77.01\% & Acc \\
&~\citet{yu2003towards} & Statistics  & MPQA, POS  & Na\"ive Bayes  & TREC  & 97.00\% & F1  \\
&~\citet{karimi2017language} & Statistics  & Language model  & Rank by similarity  & Movie  & 94.63\% & MAP  \\  
  &~\citet{chaturvedi2016bayesian} & DL & MPQA & GBN, CNN & Movie & 96.40\% & Acc \\ \hline
\multirow{4}{*}{CLSD} &\citet{banea2010multilingual} & ML  & MPQA & SMT, Na\"ive Bayes  & Multi-MPQA EN  & 74.72\% & Acc  \\ 
&~\citet{mogadala2012language} & Statistics  & \begin{tabular}[c]{@{}l@{}}Unigram and bigram \\freq., word length\end{tabular} & Na\"ive Bayes & Multi-MPQA EN & 92.50\% & F1 \\  
&~\citet{chaturvedi2016lyapunov} & DL & MPQA, WSD & SMT, CNN, RNN & MPQA Gold & 84.00\% & F1 \\ \hline
\multirow{3}{*}{MMSD}  &~\citet{murray2011subjectivity} & Statistics  & VIN, raw features & MaxEnt  & AMIDA & 52.00\% & F1  \\  & 
~\citet{raaijmakers2008multimodal} & Statistics  & \begin{tabular}[c]{@{}l@{}}Lexical, prosodic, and\\phonemic features\end{tabular}  & BoosTexter  & AMIDA  & 75.40\% & Acc \\ \hline
\multirow{8}{*}{BD} &~\citet{recasens2013linguistic} & Statistics  & \begin{tabular}[c]{@{}l@{}}Biased lexicon,\\POS\end{tabular} & Logistic regression  & Self-collected & 34.35\% & Acc  \\  
 &~\citet{hube2018detecting} & Statistics  & \begin{tabular}[c]{@{}l@{}}Word2Vec, POS, LIWC,\\biased lexicon\end{tabular} & Random Forest  & Conservapedia  & 74.00\% & Prec. \\
 &~\citet{aleksandrova2019multilingual} & Statistics & Word frequency  & BoW  & Self-collected  & 59.57\% & F1 \\
 &~\citet{hube2019neural} & DL  & GloVe, POS, LIWC  & RNN  & Self-collected  & 77.10\% & F1 \\
 &~\citet{pryzant2020automatically} & DL  & BERT  & LSTM  & WNC  & 45.80\% & Acc \\
 &~\citet{pant2020towards} & DL  & BERT  & Ensemble, BERT  & WNC  & 71.61\% & Acc \\ \bottomrule
\end{tabular}
\caption{A summary of representative subjectivity detection techniques (Part 2). CDSD denotes concept-dependent subjectivity detection. CLSD denotes cross-lingual subjectivity detection. MMSD denotes multi-modal subjectivity detection. BD denotes bias detection. SWN denotes SentiWordNet.} \label{tab:sd-summary2}
\end{sidewaystable}

Subjectivity detection is a well-studied sub-problem in affective computing and opinion mining. There are five technical trends in this area, namely individual, context-dependent, cross-lingual, multi-modal subjectivity detection, and bias detection. A summary of the trends can be found in Tables~\ref{tab:sd-summary1} and \ref{tab:sd-summary2}.

For individual subjectivity detection (Table~\ref{tab:sd-summary1}), the subjectivity of each sentence or snippet is determined only by the lexical, syntactic, and semantic information of the sentence itself. There are mainly three types of methods for individual subjectivity detection. First, the lexicon-based approaches rely on external lexicons that contain subjective and sentiment clues to predict the subjectivity of a sentence. The weakness of such an approach is that subjective clues are often not extensive and reliable enough to determine the subjectivity of a sentence. Some works attempted to address this issue by utilizing sentence-level features to extract syntactic information~\citep{wilson2004just,xuan2012linguistic, barbosa2010robust}, or incorporating WSD to identify subjective clues according to context~\citep{akkaya2009subjectivity, ortega2013improving}. Nonetheless, these methods cannot fully extract the underlying sentence structure and contextual information. Word-frequency-based approaches, on the other hand, predict sentence subjectivity according to the word presence or occurrence in a given corpus, thus being able to adapt to new domains and languages. Additionally, this approach requires little external resources or human effort. However, similar to the lexicon-based approach, word frequency methods lack the ability to capture syntactic information. To address this limitation, deep-learning-based methods utilize neural networks to learn spatial and temporal dependencies. Specifically, PLMs are widely used for their ability to provide universal representations~\citep{kim2014convolutional, liu2019multi, sun2019fine}.

For context-dependent subjectivity detection (Table~\ref{tab:sd-summary2}), the subjectivity of a sentence is determined with regards to its surrounding context, e.g., inter-sentence-level~\citep{pang2004sentimental, belinkov2018evaluating}, document-level~\citep{yu2003towards, das2009theme, karimi2017language}, or discourse-level~\citep{biyani2014using} information. In the existing works, such information is typically captured through feature engineering or statistical means.

As a large part of subjectivity detection works to some extent relies on external subjective clues, cross-lingual subjectivity detection aims specifically to solve the lack of lexical resources for non-English languages. There are mainly two branches of thought to address this problem (Table~\ref{tab:sd-summary2}). One is to make use of language-independent methods such as word frequency~\citep{rustamov2013sentence, lin2011sentence, kamil2018adaptive, belinkov2018evaluating} and language modeling~\citep{karimi2017language}. The other is to generate resources for the target language from English lexicons with the help of SMT systems~\citep{banea2010multilingual, chaturvedi2016lyapunov}.

Multi-modal subjectivity detection is a rising field of interest in accordance with the rising need for sentiment analysis in various media (Table~\ref{tab:sd-summary2}). Existing works utilized lexical, prosodic, and phonemic features for subjectivity detection in spoken conversations~\citep{murray2011subjectivity, raaijmakers2008multimodal}. Subjectivity detection in other modalities such as video remains mostly unexplored.

Bias detection is a task that is closely related to subjectivity detection (Table~\ref{tab:sd-summary2}). It aims to identify biased statements from supposedly impartial articles such as Wikipedia. Despite its greater complexity, the identification of bias exhibits technical patterns that are akin to those found in subjectivity detection, e.g., lexicon-based~\citep{recasens2013linguistic, hube2018detecting}, deep learning~\citep{hube2019neural, pryzant2020automatically}, and cross-lingual~\citep{aleksandrova2019multilingual} methods.

\subsubsection{Application Trends}\label{sec:summary_application_sd}

\begin{table}[!htbp]
\centering
\scriptsize
\begin{tabular}{llcc}
\toprule
Reference & Downstream Tasks  & Feature & Parser \\
\hline
\citet{bonzanini2012opinion}  & Sentiment Computing & & \checkmark \\
\citet{pang2004sentimental} & Sentiment Computing & & \checkmark \\
\citet{das2020subjectivity} & Sentiment Computing & & \checkmark \\
\citet{kamal2013subjectivity} & Sentiment Computing & & \checkmark \\
\citet{barbosa2010robust} & Sentiment Computing & & \checkmark \\
\citet{soong2019essential}  & Sentiment Computing & & \checkmark \\
\citet{zhang2007opinion}  & Information Retrieval & & \checkmark \\
\citet{wiebe2011finding}  & Information Retrieval & & \checkmark \\
\citet{cohen2011fighting} & Hate Speech Detection & & \checkmark \\
\citet{gitari2015lexicon} & Hate Speech Detection & \checkmark & \checkmark \\
\citet{li2008cocqa} & Question Answering  & \checkmark & \checkmark \\
\citet{li2008exploring} & Question Answering  & \checkmark & \checkmark  \\
\citet{aikawa2011community} & Question Answering  & \checkmark & \checkmark \\
\citet{stoyanov2005multi} & Question Answering  & & \checkmark \\
\citet{wan2016modeling} & Question Answering  & \checkmark & \\
\bottomrule
\end{tabular}
\caption{A summary of the representative applications of subjectivity detection in downstream tasks.}
\end{table}

Due to its filtering nature, subjectivity detection is widely used as a parser for many downstream tasks, e.g., sentiment analysis~\citep{pang2004sentimental, barbosa2010robust, kamal2013subjectivity, soong2019essential,das2020subjectivity}, information retrieval~\citep{zhang2007opinion, wiebe2011finding}, hate speech detection~\citep{gitari2015lexicon}, and QA systems~\citep{stoyanov2005multi, wan2016modeling}. Most existing works take the pipeline approach, using the filtered results from subjectivity detection as the input of the target application. On the other hand, we also observe that subjectivity lexicons can also be useful features to support hate speech detection and QA systems.

A survey of literature pertaining to subjectivity detection reveals that the progress made in this research area has not kept pace with the advancements made in its downstream sentiment computing tasks, e.g., sentiment analysis~\citep{gandhi2022multimodal}. This is likely because sentiment analysis may deliver more fine-grained classification outputs, which helps to gain business insights, e.g., sentiment polarities on product or service reviews. However, it should be noted that while positive, negative, and neutral sentiment polarities represent subsets of subjective texts, there exists a substantial portion of texts that are objective in nature, presenting factual information. Objective texts are likely to be infrequent in reviews of products or services, as customers often use such platforms to express their opinions. However, in the context of opinion mining on social media, it is crucial to differentiate between subjective and objective statements, given that even statements with neutral sentiment polarities can be indicative of an individual's opinion. Thus, it is still necessary to conduct subjectivity detection before sentiment analysis.

\subsubsection{Future Works}\label{sec:summary_future_sd}

\noindent\textbf{Fine-grained subjectivity detection.} A sentence may contain several clauses with differing subjectivity. For instance, a sentence may present two or more opinions, or contain both opinions and factual information. Therefore, to better assist downstream applications, fine-grained subjectivity detection that identifies the particular opinion-bearing clauses is worthy of investigation. However, there is limited research on this issue. \citet{benamara2011towards} proposed segment-level subjectivity detection. \citet{wilson2004just} proposed a method specifically for classifying the subjectivity of deeply nested clauses. There is scope for additional research to exploit the full potential of the fine-grained subjectivity annotation offered by the MPQA scheme~\citep{wiebe2005annotating}.

\noindent\textbf{Multi-modal subjectivity detection.} Subjectivity detection using information from multiple modalities remains largely unexplored. There is related multi-modal research that might provide inspiration for future works. \citet{wrede2003spotting} aimed to identify hot spots, which are regions in a meeting where participants are highly involved in the discussion, using solely a set of prosodic features. \citet{hillard2003detection, galley2004identifying} both targeted the detection of agreements and disagreements in meetings. The former explored the combination of lexical and prosodic features, whereas the latter incorporated pragmatic features that captured the interactions between speakers. \citet{neiberg2006emotion} recognized positive, negative, and neutral emotions in meetings using lexical and acoustic-prosodic features. \citet{somasundaran2007detecting} detected sentences and turns in meetings that express sentiment and arguing opinions using lexical and discourse features. \citet{morency2011towards, martin2013youtube, tsai2019multimodal} conducted sentiment analysis on review videos using linguistic features, acoustic features, and visual features (face tracking).

\noindent\textbf{Explainable subjectivity detection.} While much of the subjectivity detection research has utilized lexical resources such as subjectivity and affective lexicons to explain the subjective nature of text based on individual words, these resources do not capture the pragmatic nuances of words within their contextual environment. This is because the utilized lexical knowledge is context-independent. Theoretical research has explained subjectivity from the perspective of pragmatics~\citep{austin1975things,lakoff1980metaphors}. It would be valuable to study subjectivity detection that detects and explains subjectivity. Explainable subjectivity detection could push the development of more linguistics-inspired models that can account for the complexities of subjectivity and its expression in natural language. Additionally, there is potential for cross-disciplinary collaboration between linguistics, cognitive science, and computer science to further advance our understanding of subjectivity and its detection in various domains. 

\section{Discussion}\label{sect:discussion}

\subsection{Interactions between the Surveyed Tasks}

In preceding sections, we have provided an introduction to the relationships between our surveyed tasks and downstream applications. Nevertheless, it is important to recognize that these tasks are intrinsically interconnected. For example, WSD and anaphora resolution are mutually supportive for each other. Consider the following sentence:

\ex. I observed a colossal mammoth statue on the summit. It's really cool.

In this case, an anaphora resolution model should be capable of linking ``it'' to the ``mammoth statue'', assuming the significance of ``cool'' is interpreted as ``a form of approval due to the appealing attributes of the mammoth statue'', rather than the low temperature associated with the ``summit''. Conversely, if the antecedent of ``it'', denoting the ``mammoth statue'', is established, the intended meaning of ``cool'' can be easily discerned. This symbiotic enhancement is also observable in the context of WSD and NER. Within the context of the following sentence, disambiguating the sense of ``hit'' aids NER in recognizing ``King's Arm'' as a location rather than a person.

\ex. I hit the King's Arm yesterday. It's my preferred pub in London.

Likewise, accurately identifying ``King's Arm'' as a location bolsters WSD models in determining that ``hit'' should be interpreted as ``visited''.

In the domain of concept extraction, WSD for multi-word expressions assumes heightened significance. For example, ``go bananas'', ``cloud computing'', and ``pain killer'' are best captured as concepts with multi-word expressions, rather than independent words, since their meanings manifest coherently only when interpreted as integrated wholes. Absent WSD for multi-word expressions, the task of concept extraction struggles to delineate conceptual boundaries within a given sentence. Furthermore, the application of WSD techniques extends to textual subjectivity detection. Taking the adjective ``fine'' for example, it ordinarily corresponds with the subjective text due to its meaning referring to the subjective feeling of being satisfactory, as seen in ``Tesla Model X is a fine car''. However, ``fine'' can also appear in objective contexts, if construed as a monetary penalty, as demonstrated in ``I received a fine yesterday for speeding''. Another instance is the term ``long'', which can be employed in an objectively spatial context as well as a negatively subjective sense akin to ``tenacious''. Integrating a sense-sensitive approach into subjectivity detection can enhance its performance.

The aforementioned interconnectedness and instances highlight the intricate interplay of language. The explication of linguistic interpretations can encompass various dimensions, even though the surveyed tasks pertain to fundamental semantic endeavors. These tasks exhibit interdependencies and mutual dependencies. Consequently, diverse learning methods may be requisite for addressing these multidimensional linguistic interpretation tasks.

\subsection{The Impacts of Deep Learning on Semantic Processing}

In the current neural network models with end-to-end task-processing purposes, the aforementioned linguistic interpretation facets might be encapsulated within a black box, lacking explicit representation. The limitation of these approaches lies in their inability to elucidate how language is employed and construed across divergent semantic facets. While the pursuit of human-like accuracy in deep learning-based systems is prominent, it is essential to acknowledge that the simulation of human cognitive and interpretive mechanisms, akin to human-like intelligence, may just gain a secondary focus in contrast to the endeavor for heightened task accuracy in the NLP domain.

Prior to the era of deep learning, semantic processing tasks often relied on rule-based or symbolic methods~\citep{zhang2021neural}. These approaches aimed to distill the linguistic intuitions and insights associated with a given semantic processing task by leveraging a variety of linguistic features. Algorithms were devised to capture the specific linguistic nuances of each task, and substantial endeavors were undertaken to unveil the overarching principles governing semantic interpretation~\citep{akkaya2009subjectivity,cambria2014senticnet}. The process of cross-validating different linguistic features played a prominent role in the pursuit of enhanced predictive accuracy.

Nonetheless, the emergence of neural networks has brought about a convergence in the landscape of semantic learning and representations. Within the domain of semantic learning, a prevalent strategy involves utilizing contextual cues to predict a target word. This is achieved through various learning paradigms such as continuous bag of words~\citep[word2vec,][]{mikolov2013distributed}, masked word prediction~\citep[as seen in models like BERT and RoBERTa,][]{devlin2018bert,liu2019roberta}, or the prediction of the subsequent word~\citep[exemplified by the GPT families,][]{radford2018improving,radford2019language,brown2020language}. This approach has undoubtedly yielded remarkable accomplishments across a wide range of NLP tasks. Neural networks excel at capturing the fundamental meanings of words and sentences within vectorized representations, and their ability to encode contextualized meanings as the network architecture becomes deeper.

In light of the demonstrated efficacy of the aforementioned neural semantic learning paradigms, the emphasis on tailoring models to capture task-specific linguistic intuitions has diminished somewhat, compared to rule-based or symbolic methods. Nevertheless, a pertinent query arises: Is the general unified target word prediction approach of pre-training the optimal strategy for achieving multi-dimensional semantic understanding? Semantic representations in vector form possess the capacity to apprehend spatial correlations among meanings, manifesting through distinct proximities of similar and dissimilar meanings. These spatial relationships are forged through the learning of word associations. Nonetheless, substantial knowledge, such as commonsense, causality, and occurrences that are either unprecedented or infrequent, e.g., novel metaphors~\citep{ge2023survey}, remain beyond the direct purview of contextual understanding. Consequently, the comprehension of intricate constructs like frame semantics~\citep{fillmore2006frame}, narratives, and cognitive mechanisms – which intricately hinge on facets like knowledge representation, commonsense reasoning, social cognition, and learning – presents challenges when solely relying on vector representations for their explication~\citep{cambria2014jumping}.

Considering the strong connections between semantic processing tasks and linguistics, it is advisable to direct heightened attention toward the incorporation of linguistic and cognitive intuitions and the exploration of semantic interpretative dimensions via neural networks and neuro-symbolic methods that marry the advantages of neural nets and symbolic knowledge representations. This constitutes a salient characteristic demarcating computational semantics-focused research from pure machine learning-oriented deep learning studies, inspiring a broader exploration of semantic processing.

\subsection{Semantic Processing and Large Language Models}

ChatGPT and GPT-4 have expanded the reach of LLMs across diverse domains. Their remarkable proficiency in text generation, multitask execution, and complex task handling has garnered significant attention within the NLP community. Meantime, it is evident that there are noteworthy challenges associated with these expansive models, including issues such as hallucinations and complex task reasoning~\citep{mao2023gpteval}.

In the context of dialogues, \citet{cabrera2023zeno} have introduced an innovative evaluation framework tailored for LLMs. Their study has compared multiple such models and identified a recurring challenge of hallucination – a scenario where the generated content appears plausible but is, in fact, entirely fictional. Embedding semantic knowledge into these models presents an avenue to offer them with a more percise comprehension of real-world information, thereby diminishing the likelihood of generating content that lacks substantiated foundation. For instance, consider the sentence ``the horse flew over the barn''. A model enriched with semantic acumen would promptly discern the implausibility of such an event, thereby reducing the susceptibility to produce hallucinatory output. The scope of semantic acumen encompasses not solely the literal meaning of ``barn'', but also encompasses a more widespread understanding of its prevalent dimensions of size and height. Such a semantically informed model can manifest as a system adept at recognizing inconsistencies or deviations from anticipated semantic patterns. Alternatively, it could be a model proficient in grasping ordinary concept associations grounded in the frame semantics. Moreover, semantic processing can assist in reformulating user queries to render them more machine-friendly, thereby mitigating the potential for hallucinations stemming from vague queries.

On the other hand,~\citet{mao2023gpteval} have reported that ChatGPT demonstrates satisfactory performance in general scientific knowledge and can effectively address questions necessitating open-ended responses. Nonetheless, it is not without errors, particularly in cases requiring multi-step reasoning. This shortcoming may be attributed to the current practice of employing solely feedforward propagation and fast inference in LLMs~\citep{bubeck2023sparks}. The absence of human-like deliberation for complex inquiries impedes the model's capacity for intricate multi-step reasoning tasks. Recent strides in Chain-of-Thought Prompting~\citep{wei2022chain} underscore the potential of decomposing complex problems into intermediate steps to enhance the complex reasoning capabilities of LLMs. In this context, semantic processing emerges as a valuable asset for task decomposition. It can assist in identifying pivotal concepts and entities, and delineating the principal topic into coherent logical subtopics or sequential steps. Semantic comprehension ensures a seamless and coherent progression of steps, yielding prompts that are not only more efficient but also conducive to the adept reasoning of intricate challenges by LLMs.

\section{Conclusion}
\label{sect:Conclusion}

In this survey, we have reviewed recent semantic processing techniques, e.g., WSD, anaphora resolution, concept extraction, NER, and subjectivity detection. We summarized useful datasets, annotation tools and knowledge bases that can facilitate the research in these domains. We also summarized the technical trends of these techniques, related theoretical research, and their downstream applications. We found that the breadth and depth of semantic processing can be greatly extended, both from the perspective of the needs of theoretical research and downstream applications. This is because current computational semantic processing techniques are limited in their reliance on specific task settings and available datasets. The review of the downstream applications of semantic processing techniques could potentially stimulate further research into fusion methodologies, which seek to enhance the performance of downstream tasks. The semantic processing methods can not only deliver effective features for downstream tasks, but also gain insights into analyzing model behaviors and studying linguistic and cognitive patterns. 

As we continue to advance in the field of NLP, using powerful PLMs and LLMs has become increasingly common to tackle more complex NLP tasks. However, it is important to note that there is still great academic value in studying the low-level semantic tasks that these models are built upon. These tasks help us understand how language is presented and received, how semantics relates to human cognition, and how semantic processing tasks are interrelated. We observe that numerous contemporary semantic processing tasks have been translated into machine learning problems, which have somehow diminished linguistic motivations and intuitions from these computational studies. Shaping semantic processing tasks into tasks that are more conducive to machine learning can indeed improve the accuracy of specific tasks. However, improving accuracy in a single-task setting is not the only pursuit of semantic processing. We should pay more attention to how semantic processing techniques can better serve humans and machines to explain language phenomena. We hope that this paper can stimulate more research directions in the field of semantic processing and inspire researchers to place greater emphasis on the nature and cognition of semantics. With the development of more powerful tools such as PLMs and LLMs, it is perhaps valuable for our research to use these tools to address those fundamental linguistic challenges that were previously considered daunting. Regardless of the sophistication of tasks that can be performed by LLMs, basic semantic processing tasks remain crucial for comprehending and utilizing language effectively. These tasks serve as the foundation upon which our understanding of language is built.

\section*{CRediT Authorship Contribution Statement}
\textbf{Rui Mao}: Conceptualization, Methodology, Investigation, Writing-original draft (Introduction, Discussion, and Conclusion) \& harmonization. \textbf{Kai He}: Investigation, Writing-original draft (Named Entity Recognition). \textbf{Xulang Zhang}: Investigation, Writing-original draft (Subjectivity Detection). \textbf{Guanyi Chen}: Investigation, Writing-original draft (Anaphora Resolution). \textbf{Jinjie Ni}: Investigation, Writing-original draft (Word Sense Disambiguation). \textbf{Zonglin Yang}: Investigation, Writing-original draft (Concept Extraction). \textbf{Erik Cambria}: Conceptualization, Writing - review \& editing, Project administration, Supervision, Funding acquisition.

\section*{Declaration of Competing Interest}
Rui Mao, Zonglin Yang and Erik Cambria report financial support was provided by Continental Automotive Singapore Pte. Ltd.

\section*{Acknowledgment}

This study is supported under the RIE2020 Industry Alignment Fund – Industry Collaboration Projects (IAF-ICP) Funding Initiative, as well as cash and in-kind contribution from the industry partner(s).

\bibliographystyle{elsarticle-harv} 
\bibliography{references.bib}

\end{document}